\documentclass[preprint]{elsarticle}

\usepackage{times}
\usepackage{epsfig}
\usepackage{amsmath}
\usepackage{amssymb}
\usepackage{booktabs}
\usepackage[accsupp]{axessibility}
\usepackage{subcaption}
\usepackage{adjustbox}

\usepackage{natbib}
\usepackage{graphicx}
\usepackage{txfonts}
\usepackage{hyperref}

\usepackage{color}
\usepackage{float}
\usepackage[rflt]{floatflt} 
\usepackage{mathtools}
\usepackage[normalem]{ulem}
\usepackage{float} 
\usepackage[table]{xcolor} 

\newcommand{\off}[2]{{}{}}

\usepackage[ruled,vlined,lined,linesnumbered]{algorithm2e}
\usepackage{dsfont}

\usepackage{lineno}

\usepackage[capitalize]{cleveref}
\crefname{section}{Sec.}{Secs.}
\Crefname{section}{Section}{Sections}
\Crefname{table}{Table}{Tables}
\crefname{table}{Tab.}{Tabs.}

\begin{document}

\title{UruDendro, a public dataset of cross-section images of \textit{Pinus taeda}.}

\author[1]{Henry Marichal}
\author[2]{Diego Passarella}
\author[3]{Christine Lucas}
\author[4]{Ludmila Profumo}
\author[4]{Verónica Casaravilla}
\author[4]{María Noel Rocha Galli}
\author[3]{Serrana Ambite}
\author[1]{Gregory Randall}

\affiliation[1]{organization={Instituto de Ingeniería Eléctrica, Facultad de Ingeniería, Universidad de la República},
city={Montevideo},
country={Uruguay}}
\affiliation[2]{organization={Centro Universitario Regional Noreste, Universidad de la República},
city={Tacuarembó},
country={Uruguay}}
\affiliation[3]{organization={Centro Universitario Regional Litoral Norte, Universidad de la República},
city={Paysandú},
country={Uruguay}}
\affiliation[3]{organization={Centro Universitario Regional Noreste, Universidad de la República},
city={Rivera},
country={Uruguay}}

\maketitle
\thispagestyle{empty}

\begin{abstract}
The automatic detection of tree-ring boundaries and other anatomical features using image analysis has progressed substantially over the past decade with advances in machine learning and imagery technology, as well as increasing demands from the dendrochronology community.  This paper presents a publicly available database of 64 scanned images  of transverse sections of commercially grown \textit{Pinus taeda} trees from northern Uruguay, ranging from 17 to 24 years old.  The collection contains several challenging features for automatic ring detection, including illumination and surface preparation variation, fungal infection (blue stains), knot formation, missing cortex or interruptions in outer rings, and radial cracking. This dataset can be used to develop and test automatic tree ring detection algorithms. This paper presents to the dendrochronology community one such method, Cross-Section Tree-Ring Detection (CS-TRD), which identifies and marks complete annual rings in cross-sections for tree species presenting a clear definition between early and latewood.  
We compare the CS-TRD performance against the ground truth manual delineation of all rings over the UruDendro dataset.  The CS-TRD software identified rings with an average F-score of 89\% and RMSE error of 5.27px for the entire database in less than 20 seconds per image.  Finally, we propose a robust measure of the ring growth using the \emph{equivalent radius} of a circle having the same area enclosed by the detected tree ring. Overall, this study contributes to the dendrochronologist's toolbox of fast and low-cost methods to automatically detect rings in conifer species, particularly for measuring diameter growth rates and  stem transverse  area using entire cross-sections.  
\end{abstract}

\section{Introduction}
\label{sec:intro}

There is a global effort to digitize tree rings and automatize ring-width measurements. A growing number of open tree-ring data sets with site-scale metadata and ring measurements are available, but many without the accompanying images that are useful for further analysis of tree-ring attributes or for testing new tree-ring detection algorithms. The International Tree Ring Database is the most extensive collection of tree-ring data, with publicly available time series data for ring width, stable isotopes, wood density, and other ring parameters from over 5000 sites on six continents \cite{ITRDB}. For growth data on commercially valuable species, the GenTree Dendroecological Collection \cite{GenTreeDendroecologicalCollection} is a dataset with tree-ring width measurements from 3600 trees, including seven tree species from Europe. Initiatives to share not only ring data but also imagery for tree-ring detection, visualization, and analyses include DendroElevator \cite{griffin2021gigapixel}, as well as do-it-yourself lower-cost systems such as CaptuRING, with images of core samples and code available online \cite{garcia2022capturing}. 

Tree-ring detection is a long-term challenge that has made advances in automatic detection over the last decade via the application of image technology \cite{CerdaHM07, Norell-2009,GillertCVPR}. As early as 322 BC in Greece, Theophrastus noted that trees lay a new growth ring each year. Despite the antiquity of tree-ring observations, the science of tree-ring measurement and dating within the field of dendrochronology is a young and thriving discipline with many frontiers for advancement, among them the application of image processing approaches or technologies such as Image Analysis of Reflected Light (IARL) \cite{IARL1995} to automate the identification and  measurement of tree rings. Several software packages dedicated to tree-ring measurement exist at present. Some consist of tools that help practitioners manually mark ring boundaries; others include image-processing tools that allow for automated ring boundary detection but that are generally limited to species with nonporous wood, such as pines, with a distinct color transition between rings. In general, performance is widely variable depending upon the wood anatomy of each species, climate, and so on. Among such instruments are commercially available products such as WinDENDRO~\cite{windendro} and LignoVision\footnote{\url{https://software.rinntech.com/lignovision/}}. COORecorder  \cite{bualuț2016powerful} and the CDendro \cite{maxwell2021measuring} package are also commonly used and affordable tools to facilitate ring detection, early to latewood transition,  and ring measurement using digital imagery. These software (SW) tools are mainly devised for interactive use, with tools that help the practitioner mark rings; in general, they are devised for the analysis of imaged cores (wood bores) and coupled with a microscope or optical magnification device. The R package MtreeRing~\cite{MtreeRing} uses mathematical morphology for noise reduction, Canny edge detection, and watershed segmentation to detect rings automatically. As with many algorithms, MtreeRing proposes an interactive tool for manual marking\footnote{\url{https://ropensci.github.io/MtreeRing/}}. These packages are largely designed for use with core samples (often 5,15 mm in width x length of the stem radii) as opposed to entire cross-sections. Several studies compare the performance of manual and automated tracing approaches using these software tools. For example, \cite{ArenasCastro2015}  compare the results of tree-ring detections using WinDENDRO \cite{windendro}, statistical tools such as LINTAB-TSAPWin\footnote{\url{https://rinntech.info/products/tsap-win/}}  and tools from a GIS package. 

Due to challenges in ring visibility, programming learning curves, and costs, many practitioners still use a manual approach, measuring ring width with a tree-ring measuring system. For broadleaf species with false rings and micro-ring structures, especially from tropical and subtropical climates, manually marking tree rings through a microscope is still the preferred option nowadays \cite{Speer2010}. Nonetheless, in the case of nonporous wood, IARL can quickly provide data such as ring width, early- and latewood widths, as well as whole ring surface area and early and latewood surface area in cross-sections. This is a novel strategy to provide valuable data for dendroecology and silvicultural research, particularly considering that the rate of carbon accumulation increases continuously with tree size and age, despite the decline in ring width with age \cite{sillett2010increasing, Stephenson2014}. 

Automatic ring detection in cross-sectioned wood samples is also becoming more available, particularly for species of commercial interest in the forestry sector. Recently, an open database containing measurements and images of 100 Norway spruce logs (\textit{Picea abies} (L.) H.Karst.) from Northeastern France was published \cite{longuetaud}. The database includes several image modalities, including RGB images of both log ends and hyperspectral and computed tomography images of wood cross-sections sampled at both ends. The measurements performed on cross-sections include wood density, growth ring widths, and pith location but not traced rings throughout the entire section. To the best of our knowledge, very few data sets include both images and ground-truth tree-ring tracings. An example of such  a publicly available dataset  is the one presented by Kennel et al., which comprises seven images of \textit{Abies alba} \cite{KennelBS15}. 
Another example, Gillert et al.,\cite{GillertCVPR} presented a dataset of microscopy images composed of Dryas octopetala, Empetrum hermaphorditum and Vaccinium myrtillus species with 213 delineated samples (1987 rings). 
Nonetheless, these image databases' low sample size and species diversity limit the potential for testing new tree-ring detection algorithms. Within the context of a collaborative effort within the tree-ring community and parts of the forestry sector to develop dendro-metric algorithms, public data sets are fundamental for the fair and transparent comparison of the performance of different detection algorithms. One of the aims of this work is to contribute to this effort. 

A solution for detecting complete growth rings on cross-sections was proposed by Cerda et al. \cite{CerdaHM07}, based on the Generalized Hough Transform and tested on 10 images. Norell et al. proposed a method to automatically compute the number of annual rings in end faces acquired from sawmills \cite{Norell-2009}. It applies the grey-weighted polar distance transform~\cite{Norell2007GreyWP} to cores that include the pith and avoids knots or other irregularities. They used 24 images for training and 20 for testing the method. Zhou et al. proposed a method based on the traditionally manual approach, i.e., tracing two perpendicular lines across the cross-section and counting the peaks using a watershed-based method and showed results on five cross-sections of \textit{Pinus massoniana} \cite{Zhou2012}. A more recent study presents a machine learning-based approach for the automatic detection of tree rings from core samples in \textit{Picea abies} \cite{Polek2022AutomationOT}. This is the most extensive approach, and most algorithms and manual protocols use core instead of cross-sections as image input. Henke et al. proposed a semiautomatic method for the detection of tree rings on entire cross-sections using an Active Contours approach which shows positive results using several samples ~\cite{Henke2014SemiautomaticTR}.  An automatic method for pith localization and ring detection on complete cross-sections of trees was developed by Makela et al.~\cite{JacobiSets}. Neither the algorithm nor the data are available for all these algorithms, making it impossible to test these approaches against images of other species in other regions.  The Dual-Tree Complex Wavelet Transform \cite{Kingsbury2001ComplexWF} was used by Kennel et al. \cite{KennelBS15} as part of an active contour approach. This method uses the cross-section and gives very good results on seven publicly available images (including the ground truth rings). To the best of our knowledge, the code is unavailable, thus rendering difficulties for transferring this method to other species.
Gillert et al, \cite{GillertCVPR} proposed a method for tree-ring detection over the whole cross-section but applied it to microscopy images. They apply a deep learning approach using an Iterative Next Boundary Detection Network, trained and tested with microscopy images.

In southeastern South America, including Uruguay, southern Brazil, and northeastern Argentina, forestry has grown exponentially over the past five decades \cite{Bernardi}. Solely within Uruguay, the surface area of commercial plantations has increased from less than 180,000 ha to 1,103,686 ha, or 6\% of the country’s terrestrial surface area between 1975 and 2021  \cite{DGF}.  Plantations in Uruguay are composed of 80\% Eucalyptus (mainly \textit{E. grandis }W.Hill ex Maiden, including hybrids, clones, and \textit{E. salign} a Sm.) and 18\% \textit{Pinus taeda} L. and \textit{Pinus elliottii} Engelm. \cite{MGAP2022}.  Plantations of \textit{Pinus taeda }are grown for solid wood products, including timber, plywood, and Cross Laminated Timber. With the growing experience in silvicultural practices (plantation density management, thinning, and pruning techniques) for these exotic species, some problems have emerged in commercially grown wood. One of these problems is the presence of reaction - or compression - wood (CW) in \textit{Pinus taeda}. CW has remarkably different properties than regular wood, particularly higher wood density, greater contraction coefficient, reddish color, and slower drying rates. These features produce different problems in the industrialization phase, such as excessive warping during sawing, higher moisture content, splitting and cracking, and excessive deformation after drying. These problems are generated at different phases of industrialization, in sawmills, uncoiled wood, and during the kiln-drying processes, generating losses of nearly 10 percent throughout the production chain\footnote{D. Malates and L. Ingaramo, personal communication.}.  One of the goals of automated ring detection in \textit{Pinus taeda }cross-sections is to develop a method that could also measure the prevalence of compression wood abnormalities in \textit{Pinus }spp and study the factors that promote their formation. 

This study aims to develop automated ring detection on \textit{Pinus taeda} cross-sections from  Uruguay.  In order to develop a more efficient method for measuring growth parameters that could be performed on a massive scale, we aimed to develop software capable of automatic tree-ring boundary detection.  We first developed a public database for \textit{Pinus taeda} from Uruguay, including 64 cross-section images and expert-traced tree rings (ground truth) on those images.  The use of entire stem cross-sections was required to detect variations in ring width and distribution of CW throughout the whole cross-section. Secondly, we developed an automatic algorithm that detects tree-ring boundaries, allowing the measurement of several geometric parameters such as the distance between tree rings in scanned images of wood cross-sections. Here we present the algorithm to the dendrochronology community and explain its principal characteristics in general terms. A detailed explanation, the source code, and a demo page will be available at Image Processing On Line,\off{ \footnote{\url{https://ipolcore.ipol.im/demo/clientApp/demo.html?id=77777000390}}} a journal and public repository of image processing algorithms in accordance to the open science paradigm, allowing other teams to test the algorithm with their data. Finally, we propose the use of an equivalent disk with the same area as the corresponding traced ring for the measurement of the total area of the annual growth of the ring, rather than  few radial directions. 

\section{Methods}
\label{sec:antecedentes}
\subsection{UruDendro: A public Uruguayan data set of cross-sectioned trees with ground truth marked rings.}
 
\label{sec:ddbb}

Fourteen trees were collected in February 2020 from two tree plantations located in Buena Unión, Rivera, in northeastern Uruguay (Plywood company stands: 31°17'45"S 55°41'42" W; lumber company stands: 31°10'4'' S 55°39'42''W).  This region has a humid subtropical climate (Cfa, Koppen Climate Classification) with a mean annual rainfall of 1472 ±  373 mm/y (mean ± standard deviation), ranging from 830-2797 mm annually, and a mean annual temperature of 17.2 ± 0.4 according to publicly available climate data from the National Institute of Agricultural Research (INIA) Tacuarembo meteorological station, 1978-2022 \cite{INIA2023}.  Soils in both stands are comprised of sandstone (\textit{areniscas} in Spanish), and pertain to the national soil classes 7.2 and 7.31 according to the National Commission of Agronomic Study of Soils [\textit{Comisión Nacional de Estudio Agronómico de la Tierra} (CO.N.E.A.T.)] system.  Soil type 7.2 is considered a Tacuarembo sandstone with a slope of 10-15 degrees and deep soils with very good drainage.  Soil type 7.31 is also a Tacuarembo sandstone with a 6-10 degrees slope and good drainage. Prevailing winds are, in order of importance, from the southwest, east, south, and southeast, with an average intensity of 15 to 20 km/h.

Seven trees were harvested from tree plantations managed by a lumber company (denoted by the letter F), and seven by a plywood company (denoted by the letter L). Each company applied different silvicultural practices to the stand prior to harvest (Table \ref{tab:CultureScheme}). Cross-sections between 5 and 20 cm thick were cut from logs at 10, 165, 200, 400 and 435 cm above ground, totalling a sum of 64 cross-sections (1 to 5 cross-sections per tree). The image of each cross-section  was assigned an identification code comprising the letter of the company (F or L), a two-digit number corresponding to the individual, and a lowercase letter corresponding to the height where each cross-section was cut. Heights codes were as follows: $a = 10$ cm, $b = 165$ cm, $c = 200$ cm, $d = 400$ cm, and $e = 435$ cm above ground.

\begin{table}[ht]
\centering
\caption{Silvicultural practices within each stand for both companies. }
\begin{tabular}{llllll}
\hline
Age & Lumber company (F) & Density (trees/ha.) & Plywood company (L)  & Density (trees/ha.)       \\ \hline
0   & Plantation in 1995                & 1111                 & Plantation in 2002   & 1000   \\
3   & Thinning and pruning to 2.1 m             &                      &               & \\
4   & Thinning and pruning to 4.2 m             & 666                  & Thinning and pruning to 1.8 m & 533   \\
5   &                            &                      & Pruning to 3.4 m   &           \\
6   &                            &                      & Pruning to 4.5 m   &            \\
7   & Pruning to 6.4 m               &                      & Pruning to 6.4 m   &       \\
9   & Pruning to 7.2 m               &                      &                &        \\
10  & Thinning                        & 450                  & Thinning            & 333      \\
14  & Thinning                        & 311                  &                &     \\ \hline
\end{tabular}
\label{tab:CultureScheme}
\end{table}

\begin{figure*}[!htbp]
\begin{centering}
    \begin{subfigure}{0.3\textwidth}
    \begin{centering}
    \includegraphics[width=\textwidth]{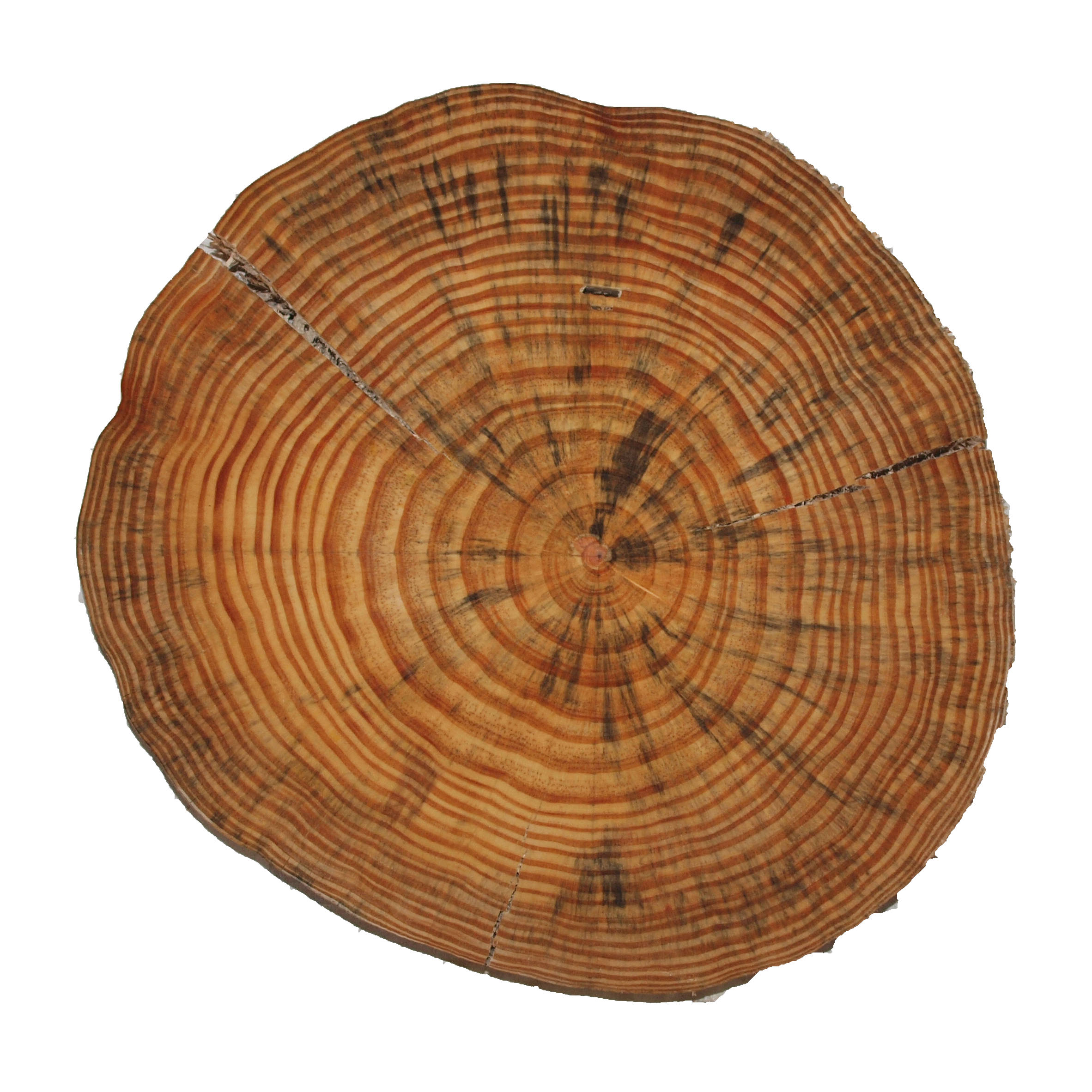}
    \label{fig:ddbb-F02a}
    \caption{F02a}
    \end{centering}
    \end{subfigure}
    \hfill
    \begin{subfigure}{0.3\textwidth}
    \begin{centering}
    \includegraphics[width=\textwidth]{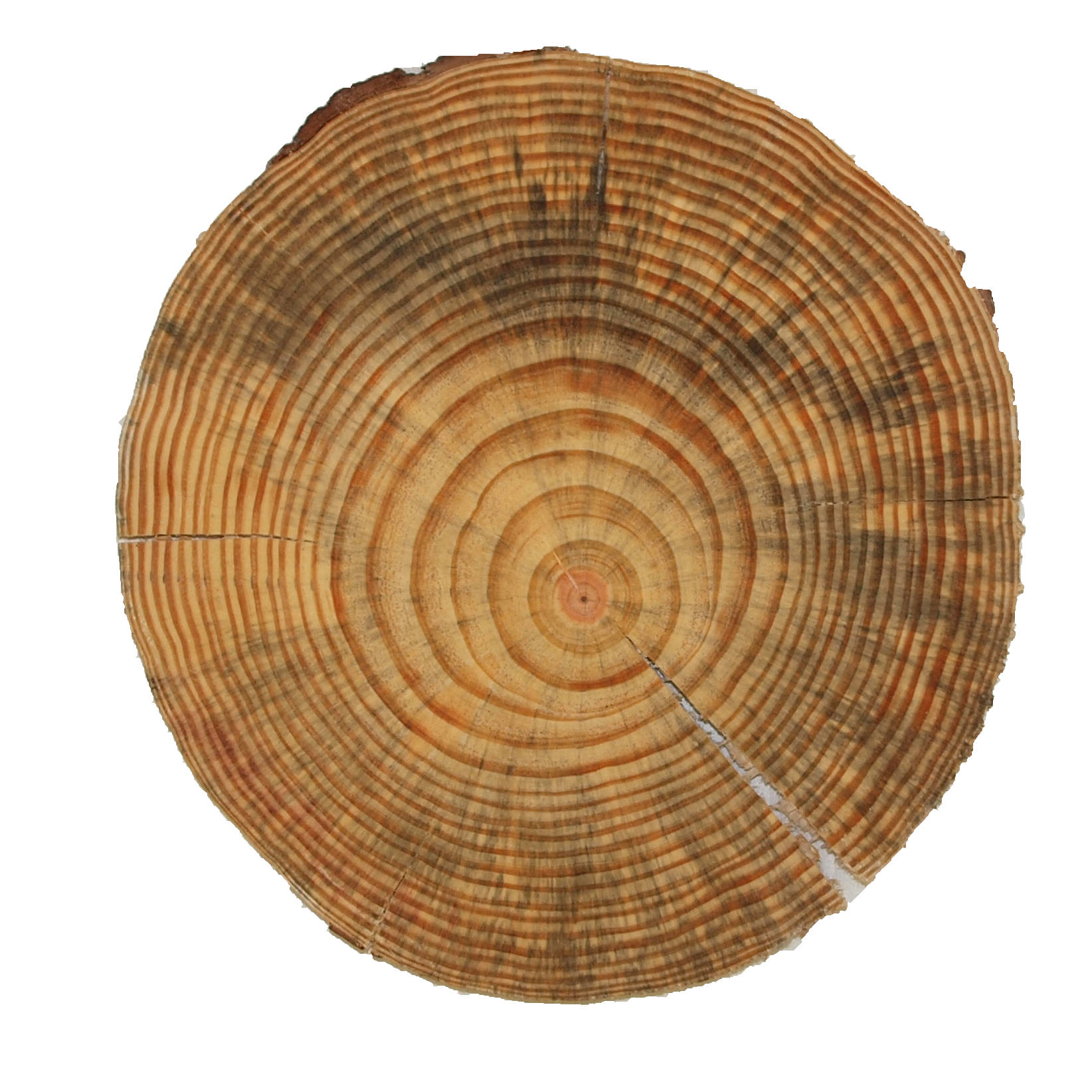}
    \label{fig:ddbb-F02b}
    \caption{F02b}
    \end{centering}
    \end{subfigure}
    \hfill
    \begin{subfigure}{0.3\textwidth}
    \begin{centering}
    \includegraphics[width=\textwidth] {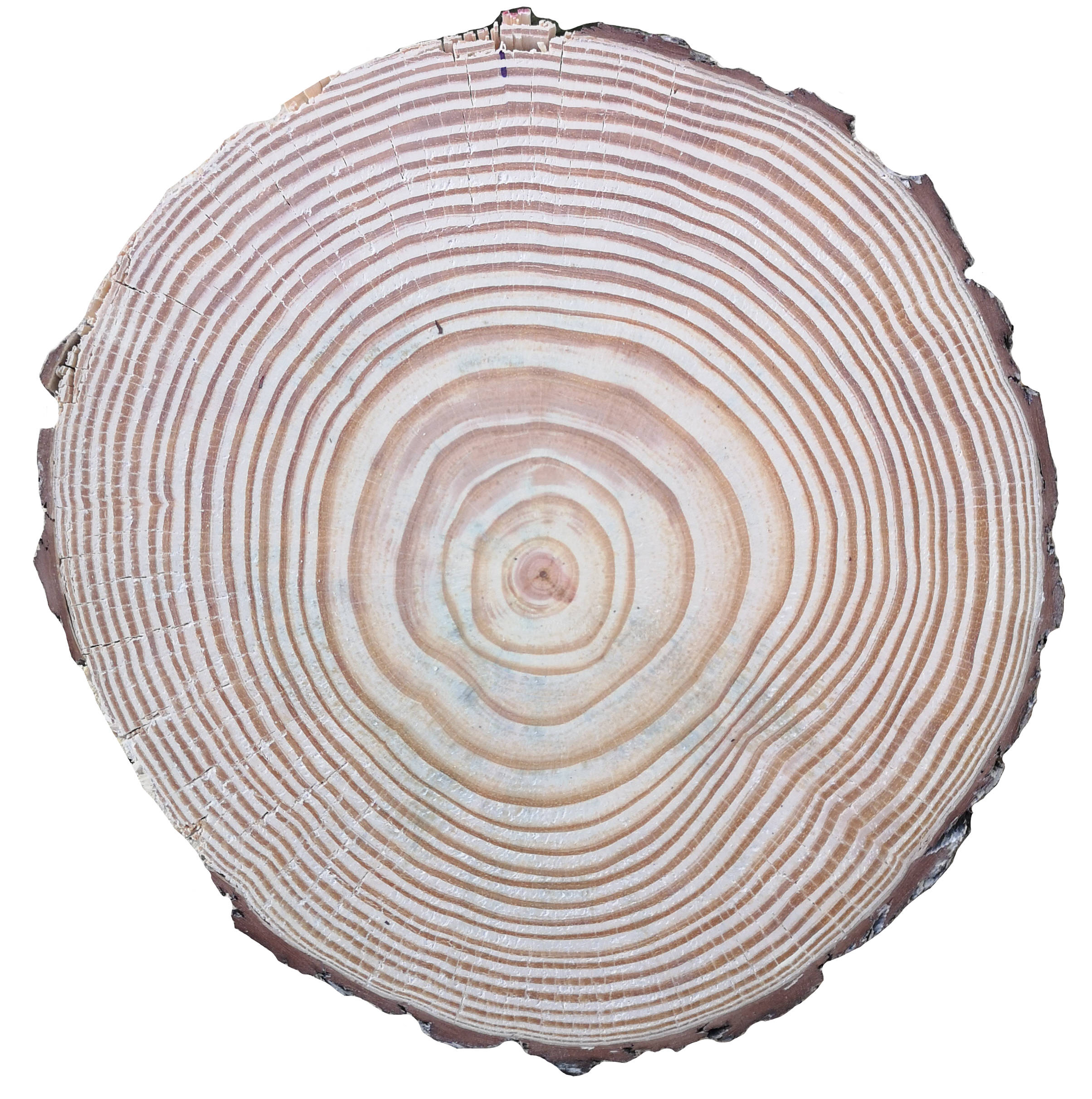}
    \label{fig:ddbb-F02c}
    \caption{F02c}
    \end{centering}
    \end{subfigure}
    \hfill

    \begin{subfigure}{0.3\textwidth}
    \begin{centering}
   \includegraphics[width=\textwidth]{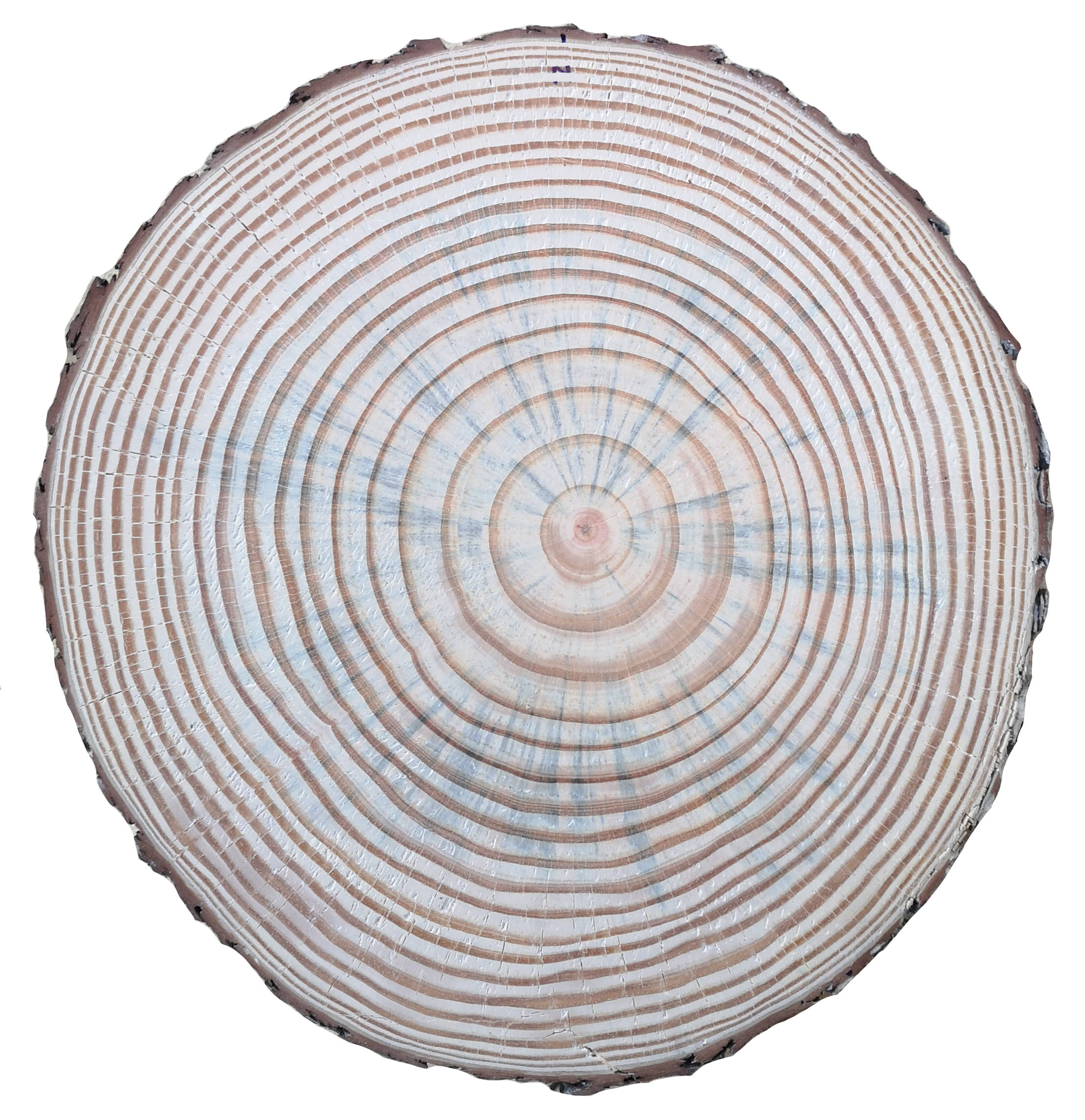}
    \label{fig:ddbb-F02d}
    \caption{F02d}
    \end{centering}
    \end{subfigure}
    \hfill
    \begin{subfigure}{0.3\textwidth}
    \begin{centering}
   \includegraphics[width=\textwidth]{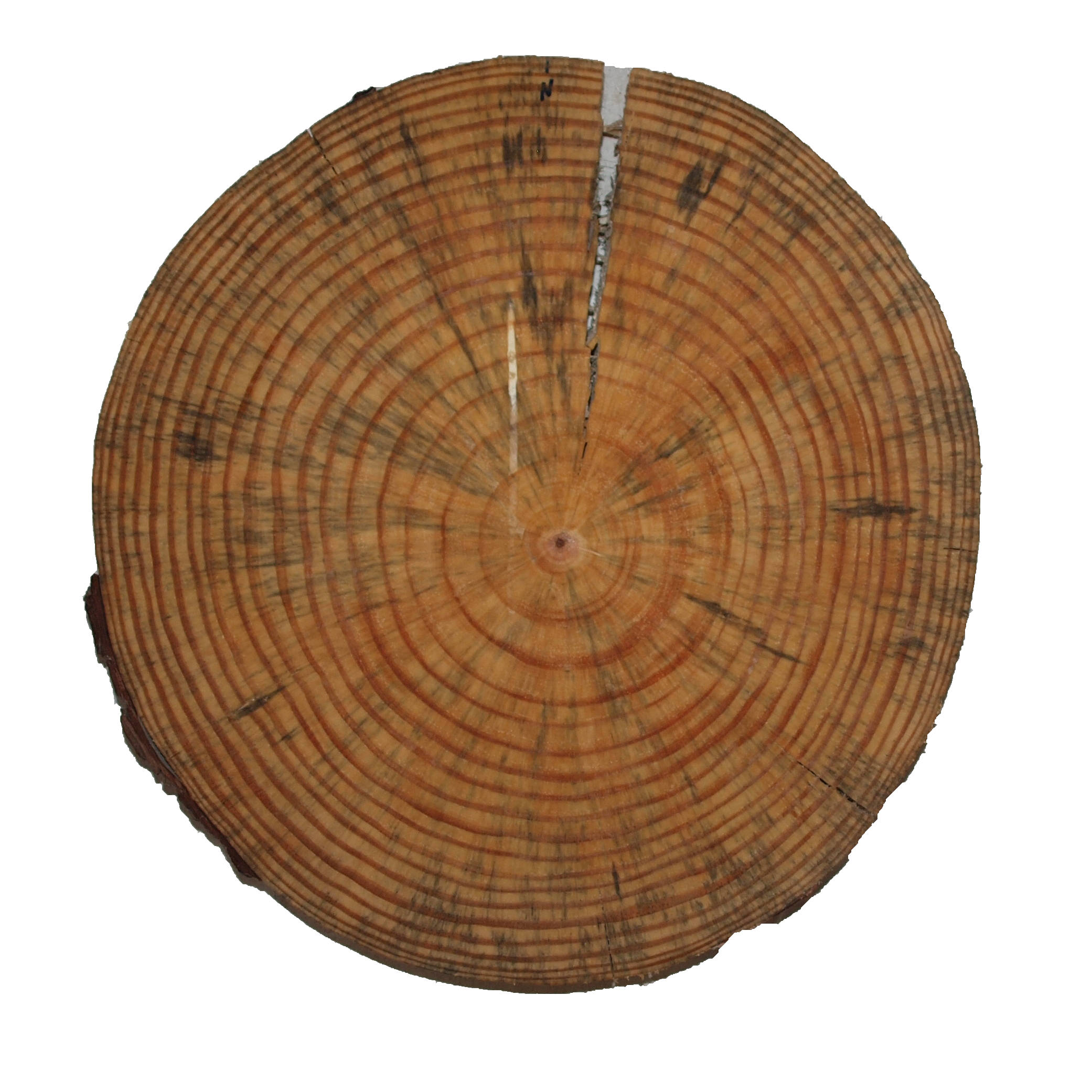}
    \label{fig:ddbb-F02e}
    \caption{F02e}
    \end{centering}
    \end{subfigure}
    \hfill
    \begin{subfigure}{0.3\textwidth}
    \begin{centering}
   \includegraphics[width=\textwidth]{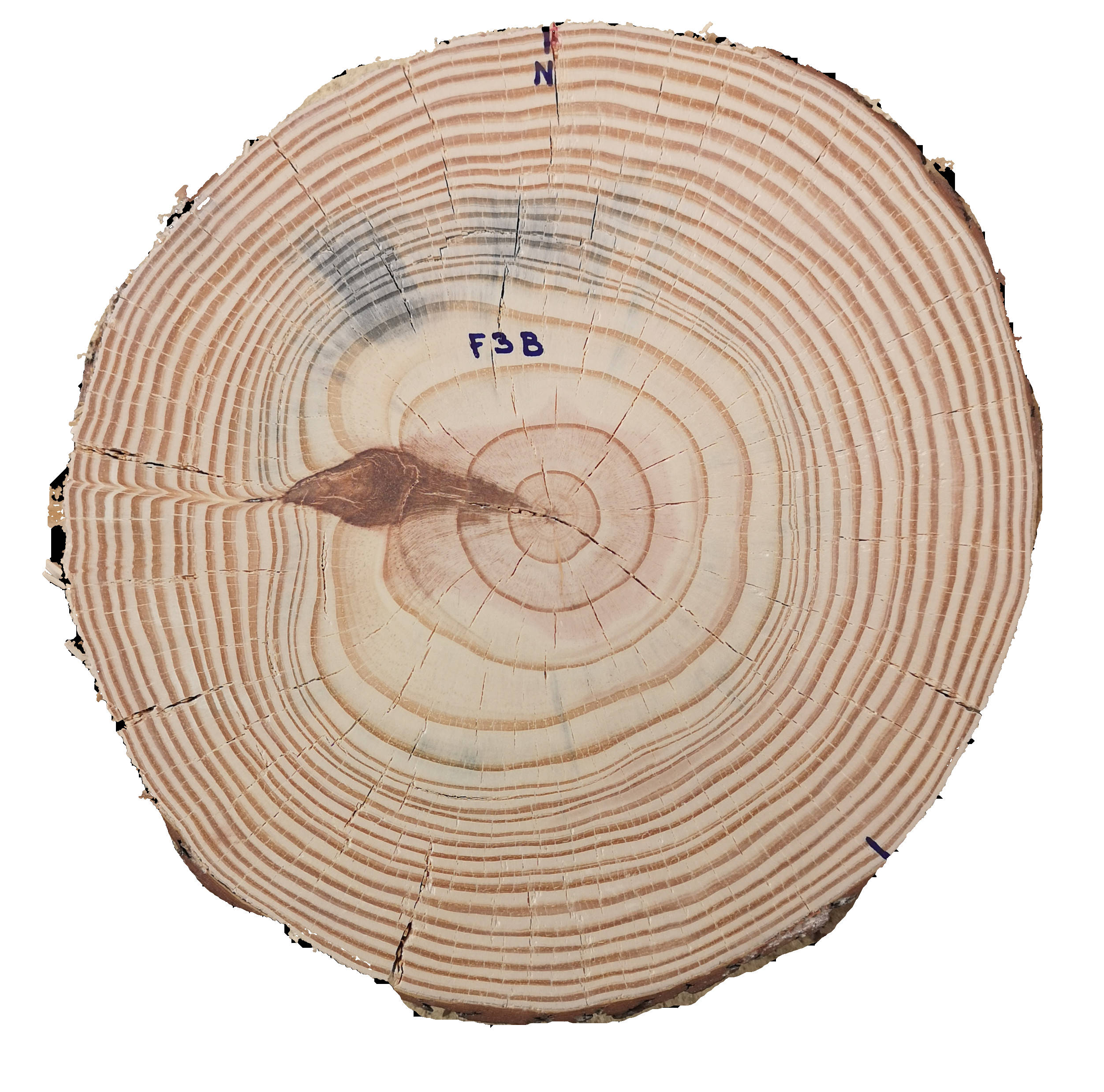}
    \label{fig:ddbb-F03c}
    \caption{F03c}
    \end{centering}
    \end{subfigure}
    \hfill
   
    \begin{subfigure}{0.3\textwidth}
    \begin{centering}
   \includegraphics[width=\textwidth]{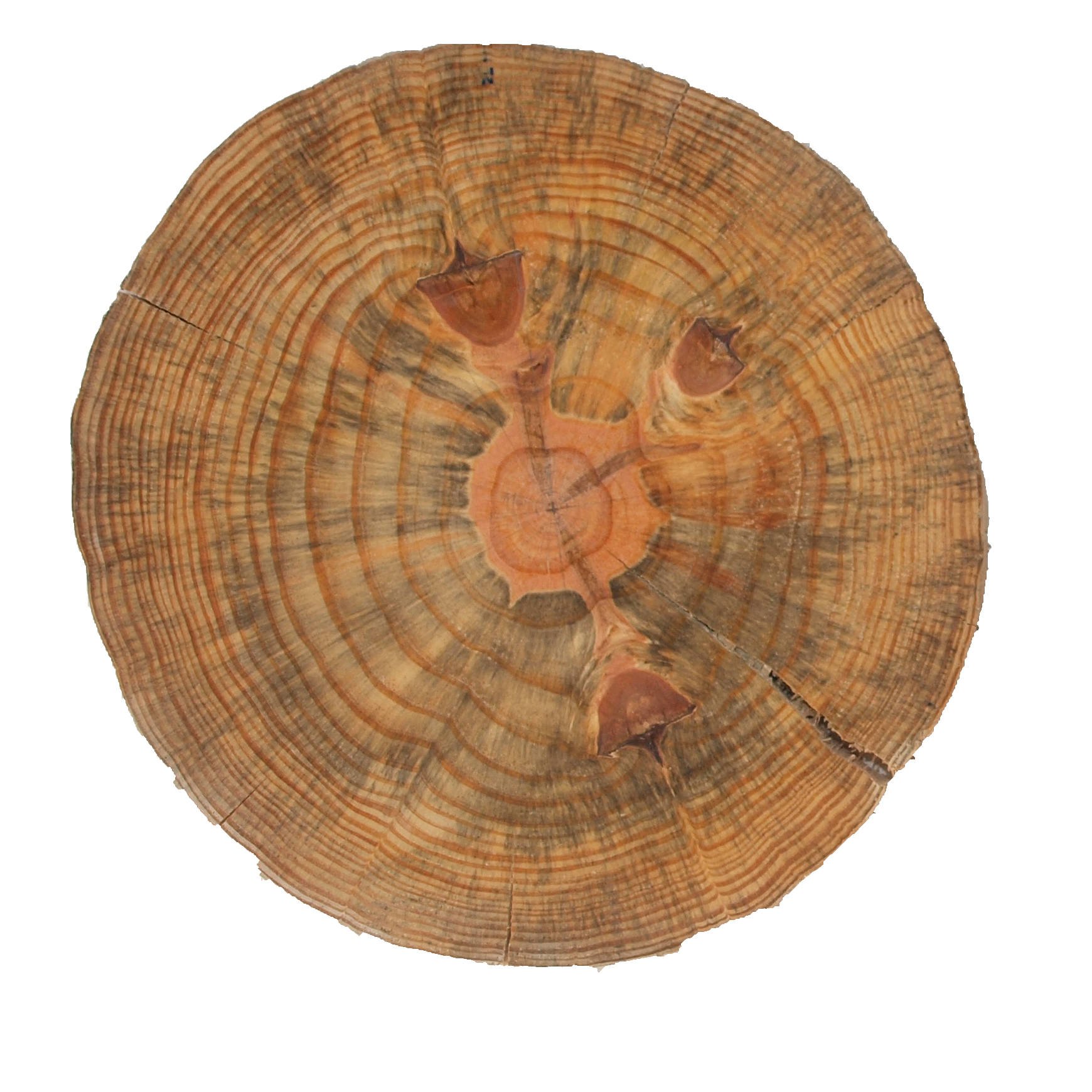}
    \label{fig:ddbb-F07b}
    \caption{F07b}
    \end{centering}
    \end{subfigure}
    \hfill
    \begin{subfigure}{0.3\textwidth}
    \begin{centering}
   \includegraphics[width=\textwidth]{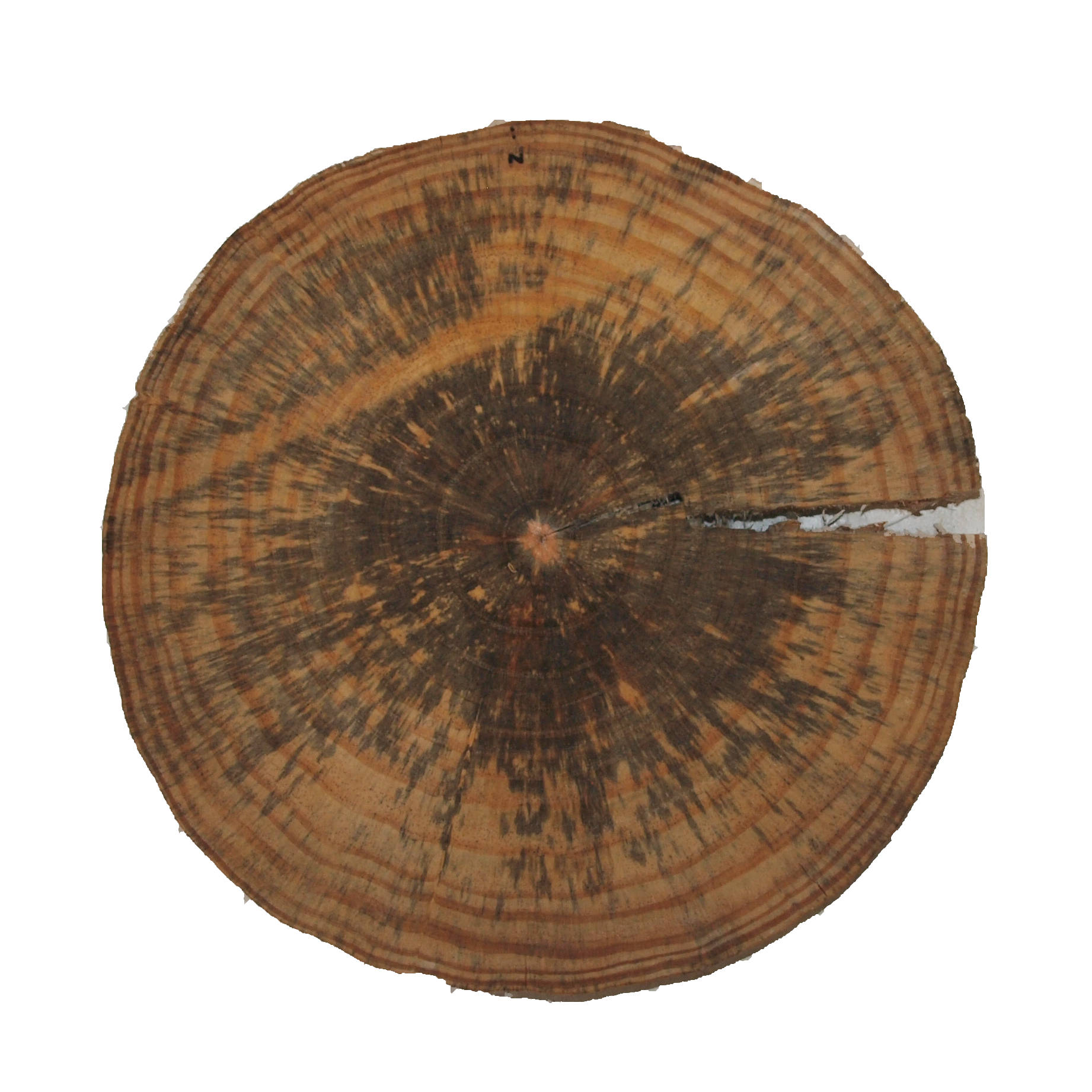}
    \label{fig:ddbb-L02b}
    \caption{L02b}
    \end{centering}
    \end{subfigure}
    \hfill
    \begin{subfigure}{0.3\textwidth}
    \begin{centering}
   \includegraphics[width=\textwidth]{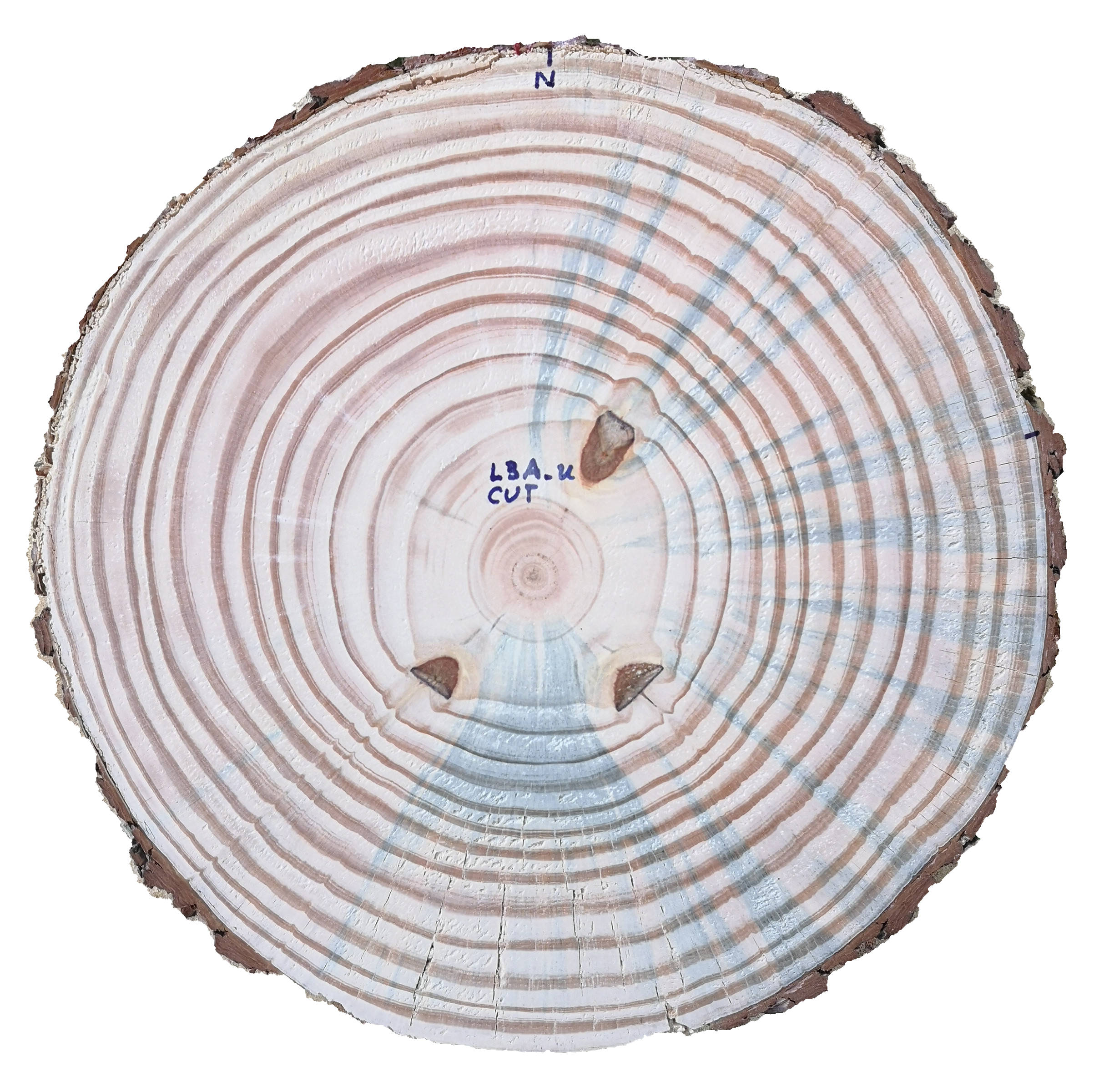}
    \label{fig:ddbb-L03c}
    \caption{L03c}
    \end{centering}
    \end{subfigure}
    \hfill
   \caption{Nine representative examples  of the 64 images in the UruDendro data set. Note the variability in color, contrast, the presence of fungus (e.g., image L02b), knots (e.g., images F07b and F03c), and cracks (e.g., images F02e and L03c). The first five images are from the same tree taken at multiple heights. }
   \label{fig:ddbb}
\end{centering}
\end{figure*}

The cross-sections were dried at room temperature without further preparation. As a consequence of the drying process, radial cracks and blue fungus stains appeared in many samples. Surfaces were smoothed with a handheld planer and a rotary sander. Photographs were taken under different lighting conditions; cross-sections a, b, and e, were photographed indoors with a cell phone, moistening the surface to maximize contrast between early and late wood, while photographs of dry cross-sections c and d were taken outdoors. For each image, the north side of the tree was marked, and a length reference for calibration was noted. \Cref{fig:ddbb} shows some images representative of the variability in samples within the database. 

\Cref{tab:data-set} summarizes the principal characteristics of the data set, composed of 64 images. The ID code of each image corresponds to the company, the specific tree, and the height of the cut. The second column indicates the total number of experts that labeled each image. The third column contains the number of rings, and the fourth is the first year of the tree which can be seen in that cross-section.  The last two columns indicate the image's height and width in pixels, respectively.

\subsection{Manual ring tracing}
Manual ring tracing was applied to generate a ground truth (GT) set of marks by experts and to correct the results generated by the automatic algorithm. In order to manually trace the rings of each cross-section, a Graphical User Interface (GUI) was developed based on Labelme\cite{Wada_Labelme_Image_Polygonal}. The software generates closed polygons formed by vertices positioned by the user over the ring borders throughout its entire circumference. The GUI imposes the structure depicted in \cref{fig:definitions}(a). Each closed polygon is a \emph{ring}. The user can place the vertices freely on top of the image.  The definition of the polygon depend on the number of vertices defined by the user. The GUI allows the magnification of the image in order to capture fine details of the ring.

The database was split among four technicians (researchers, undergraduate and graduate students) varying in their expertise in tree-ring identification, all specially trained.  Each photograph was labeled by one to four experts as seen in \Cref{tab:data-set}. To consolidate the results, a senior expert verified the number of detected rings on each photograph.

\begin{table}[htbp]
\centering
\caption{The UruDendro data set. (a) Images corresponding to the lumber company and (b) Images corresponding to the plywood company. For each one are mentioned the name, the number of expert traces (marks), the number of rings,  the first year of the tree which can be seen in the given cross-section, and the height and width of the image, in pixels.}
\label{tab:data-set}
\subfloat[Lumber company]{
\begin{tabular}{lccccc}
\hline \textbf{Name} & \textbf{Marks} & \textbf{Rings} & \textbf{Year} & \textbf{Height} & \textbf{Width} \\ \hline
\textbf{F02a} & 2 & 23 & 1996 & 2364 & 2364 \\ 
\textbf{F02b} & 2 & 22 & 1997 & 1644 & 1644 \\ 
\textbf{F02c} & 4 & 22 & 1997 & 2424 & 2408 \\ 
\textbf{F02d} & 2 & 20 & 1999 & 2288 & 2216 \\ 
\textbf{F02e} & 2 & 20 & 1999 & 2082 & 2082 \\ 
\textbf{F03a} & 2 & 24 & 1995 & 2514 & 2514 \\ 
\textbf{F03b} & 2 & 23 & 1996 & 1794 & 1794 \\ 
\textbf{F03c} & 1 & 23 & - & 2528 & 2596 \\ 
\textbf{F03d} & 1 & 21 & - & 2476 & 2504 \\ 
\textbf{F03e} & 3 & 21 & 1998 & 1961 & 1961 \\ 
\textbf{F04a} & 2 & 24 & 1995 & 2478 & 2478 \\ 
\textbf{F04b} & 1 & 23 & 1996 & 1760 & 1762 \\ 
\textbf{F04c} & 3 & 21 & - & 913 & 900 \\ 
\textbf{F04d} & 2 & 21 & - & 921 & 899 \\ 
\textbf{F04e} & 1 & 21 & 1999 & 2070 & 2072 \\ 
\textbf{F07a} & 1 & 24 & 1995 & 2400 & 2400 \\ 
\textbf{F07b} & 3 & 23 & 1996 & 1740 & 1740 \\ 
\textbf{F07c} & 1 & 23 & - & 978 & 900 \\ 
\textbf{F07d} & 1 & 22 & - & 997 & 900 \\ 
\textbf{F07e} & 1 & 22 & 1998 & 2034 & 2034 \\ 
\textbf{F08a} & 3 & 24 & 1995 & 2383 & 2383 \\ 
\textbf{F08b} & 2 & 23 & 1996 & 1776 & 1776 \\ 
\textbf{F08c} & 2 & 23 & - & 2624 & 2736 \\ 
\textbf{F08d} & 2 & 22 & - & 2388 & 2400 \\ 
\textbf{F08e} & 2 & 22 & 1998 & 1902 & 1902 \\ 
\textbf{F09a} & 1 & 24 & 1996 & 2106 & 2106 \\ 
\textbf{F09b} & 4 & 23 & 1996 & 1858 & 1858 \\ 
\textbf{F09c} & 1 & 23 & - & 2370 & 2343 \\ 
\textbf{F09d} & 1 & 23 & - & 2256 & 2288 \\ 
\textbf{F09e} & 1 & 22 & 1998 & 1610 & 1609 \\ 
\textbf{F10a} & 2 & 23 & 1996 & 2136 & 2136 \\ 
\textbf{F10b} & 2 & 22 & 1997 & 1677 & 1677 \\ 
\textbf{F10e} & 1 & 21 & 1998 & 1800 & 1800 \\ \hline
\end{tabular}
}
\subfloat[Plywood company]{
\label{}
\begin{tabular}{lccccc}
\hline \textbf{Name} & \textbf{Marks} & \textbf{Rings} & \textbf{Year} & \textbf{Height} & \textbf{Width} \\ \hline
\textbf{L02a} & 1 & 16 & 2004 & 2088 & 2088 \\ 
\textbf{L02b} & 3 & 15 & 2005 & 1842 & 1842 \\ 
\textbf{L02c} & 1 & 14 & - & 1016 & 900 \\ 
\textbf{L02d} & 2 & 14 & - & 921 & 900 \\ 
\textbf{L02e} & 2 & 14 & 2006 & 1914 & 1914 \\ 
\textbf{L03a} & 2 & 17 & 2003 & 2296 & 2296 \\ 
\textbf{L03b} & 2 & 16 & 2004 & 2088 & 2088 \\ 
\textbf{L03c} & 2 & 16 & - & 2400 & 2416 \\ 
\textbf{L03d} & 2 & 15 & - & 2503 & 2436 \\ 
\textbf{L03e} & 2 & 14 & 2005 & 1944 & 1944 \\ 
\textbf{L04a} & 4 & 17 & 2003 & 2418 & 2418 \\ 
\textbf{L04b} & 2 & 16 & 2004 & 1986 & 1986 \\ 
\textbf{L04c} & 2 & 16 & - & 2728 & 2704 \\ 
\textbf{L04d} & 2 & 15 & - & 2544 & 2512 \\ 
\textbf{L04e} & 1 & 15 & 2005 & 1992 & 1992 \\ 
\textbf{L07a} & 2 & 17 & 2003 & 2328 & 2328 \\ 
\textbf{L07b} & 2 & 16 & 2004 & 2118 & 2118 \\ 
\textbf{L07c} & 3 & 16 & - & 2492 & 2481 \\ 
\textbf{L07d} & 1 & 15 & - & 2480 & 2456 \\ 
\textbf{L07e} & 2 & 14 & 2005 & 1980 & 1980 \\ 
\textbf{L08a} & 2 & 17 & 2003 & 2268 & 2268 \\ 
\textbf{L08b} & 2 & 16 & 2004 & 1836 & 1836 \\ 
\textbf{L08c} & 1 & 16 & - & 2877 & 2736 \\ 
\textbf{L08d} & 2 & 14 & - & 2707 & 2736 \\ 
\textbf{L08e} & 1 & 15 & 2005 & 1666 & 1666 \\ 
\textbf{L09a} & 1 & 17 & 2003 & 1963 & 1964 \\ 
\textbf{L09b} & 3 & 16 & 2004 & 1802 & 1802 \\ 
\textbf{L09c} & 1 & 16 & - & 943 & 897 \\ 
\textbf{L09d} & 2 & 15 & - & 1006 & 900 \\ 
\textbf{L09e} & 1 & 15 & 2005 & 1662 & 1662 \\ 
\textbf{L11b} & 4 & 16 & 2004 & 1800 & 1800 \\ 
 &  &  & &  &  \\ 
 &  &  & &  &  \\ \hline
\end{tabular}
}
\end{table}

\Cref{tab:resultadosExpertos} shows the variability between experts, indicated as V, M, L, and C. For all images with more than one expert mark, we measured the RMS distance between the marks produced by each expert and the GT (defined as the mean value of the marks by all experts for a given cross-section). The analysis was made for cross-sections labeled by more than one expert. 

\begin{table}[htbp]
\centering
\caption{RMS Error, in pixels, between the tracing by each expert and the Ground Truth. The GT is the mean value between the experts when there is more than one tracing for a given cross-section.}

\subfloat[Lumber company]{
\label{}
\begin{tabular}{lccccc}
\hline
\textbf{Image} & \textbf{V} & \textbf{M} & \textbf{S} & \textbf{C} & \textbf{Average} \\ \hline
\textbf{F02a} & 1,343 & 1,332 &  &  & 1,338 \\ 
\textbf{F02b} & 0,866 &  & 0,860 &  & 0,863 \\ 
\textbf{F02c} & 1,012 & 1,135 & 1,224 & 1,335 & 1,177 \\ 
\textbf{F02d} & 0,641 &  &  & 0,634 & 0,638 \\ 
\textbf{F02e} &  & 1,909 &  & 1,913 & 1,911 \\ 
\textbf{F03a} &  & 2,200 & 2,220 &  & 2,210 \\ 
\textbf{F03b} & 0,799 &  &  & 0,811 & 0,805 \\ 
\textbf{F03e} &  & 1,270 & 1,076 & 1,110 & 1,152 \\ 
\textbf{F04a} & 1,162 &  & 1,166 &  & 1,164 \\ 
\textbf{F04c} &  & 1,238 & 0,986 & 1,034 & 1,086 \\ 
\textbf{F04d} & 0,616 &  &  & 0,622 & 0,619 \\ 
\textbf{F07b} &  & 3,248 & 2,148 & 2,300 & 2,565 \\ 
\textbf{F07c} &  & 0,540 &  &  & 0,540 \\ 
\textbf{F07d} &  &  &  & 0,331 & 0,331 \\ 
\textbf{F08a} & 1,554 & 1,968 & 1,398 &  & 1,640 \\ 
\textbf{F08b} & 0,813 &  &  & 0,813 & 0,813 \\ 
\textbf{F08c} & 0,943 & 0,939 &  &  & 0,941 \\ 
\textbf{F08e} & 0,944 &  & 0,942 &  & 0,943 \\ 
\textbf{F09b} & 1,180 & 1,962 & 1,047 & 1,173 & 1,341 \\ 
\textbf{F10a} & 0,945 &  & 0,942 &  & 0,944 \\ 
\textbf{F10b} & 0,815 &  &  & 0,799 & 0,807 \\ 
     &       &       &       &       & \\ \hline
\end{tabular}
}
\subfloat[Plywood company]{
\label{}
\begin{tabular}{lccccc}
\hline
\textbf{Image} & \textbf{V} & \textbf{M} & \textbf{S} & \textbf{C} & \textbf{Average} \\ \hline
\textbf{L02b} & 2,210 & 2,008 & 2,211 &  & 2,143 \\ 
\textbf{L02d} & 0,699 &  &  & 0,699 & 0,699 \\ 
\textbf{L02e} &  &  & 1,152 & 1,153 & 1,153 \\ 
\textbf{L03a} &  &  & 1,810 & 1,812 & 1,811 \\ 
\textbf{L03b} & 1,185 & 1,197 &  &  & 1,191 \\ 
\textbf{L03c} & 1,225 &  & 1,238 &  & 1,232 \\ 
\textbf{L03d} &  & 1,254 &  & 1,247 & 1,251 \\ 
\textbf{L03e} & 1,224 & 1,227 &  &  & 1,226 \\ 
\textbf{L04a} & 1,434 & 1,643 & 1,637 & 1,502 & 1,554 \\ 
\textbf{L04b} &  & 1,194 &  & 1,198 & 1,196 \\ 
\textbf{L04c} &  & 1,162 & 1,174 &  & 1,168 \\ 
\textbf{L04d} & 1,011 &  & 0,995 &  & 1,003 \\ 
\textbf{L07a} & 1,772 & 1,754 &  &  & 1,763 \\ 
\textbf{L07b} &  &  & 1,250 & 1,240 & 1,245 \\ 
\textbf{L07c} & 1,092 & 1,681 & 1,529 &  & 1,434 \\ 
\textbf{L07e} & 1,303 & 1,314 &  &  & 1,309 \\ 
\textbf{L08a} &  &  & 1,355 & 1,359 & 1,357 \\ 
\textbf{L08b} & 1,985 & 1,985 &  &  & 1,985 \\ 
\textbf{L08d} & 1,050 &  &  & 1,033 & 1,042 \\ 
\textbf{L09b} & 1,345 & 1,632 &  & 1,194 & 1,390 \\ 
\textbf{L09d} &  &  & 0,937 & 1,016 & 0,977 \\
\textbf{L11b} & 1,120 & 1,355 & 1,378 & 1,173 & 1,257 \\  \hline 
\end{tabular}
}
\label{tab:resultadosExpertos}
\end{table}

\begin{figure}[!htbp]
\centering
   \begin{subfigure}{0.3\textwidth}
   \begin{centering}
   \includegraphics[width=\textwidth]{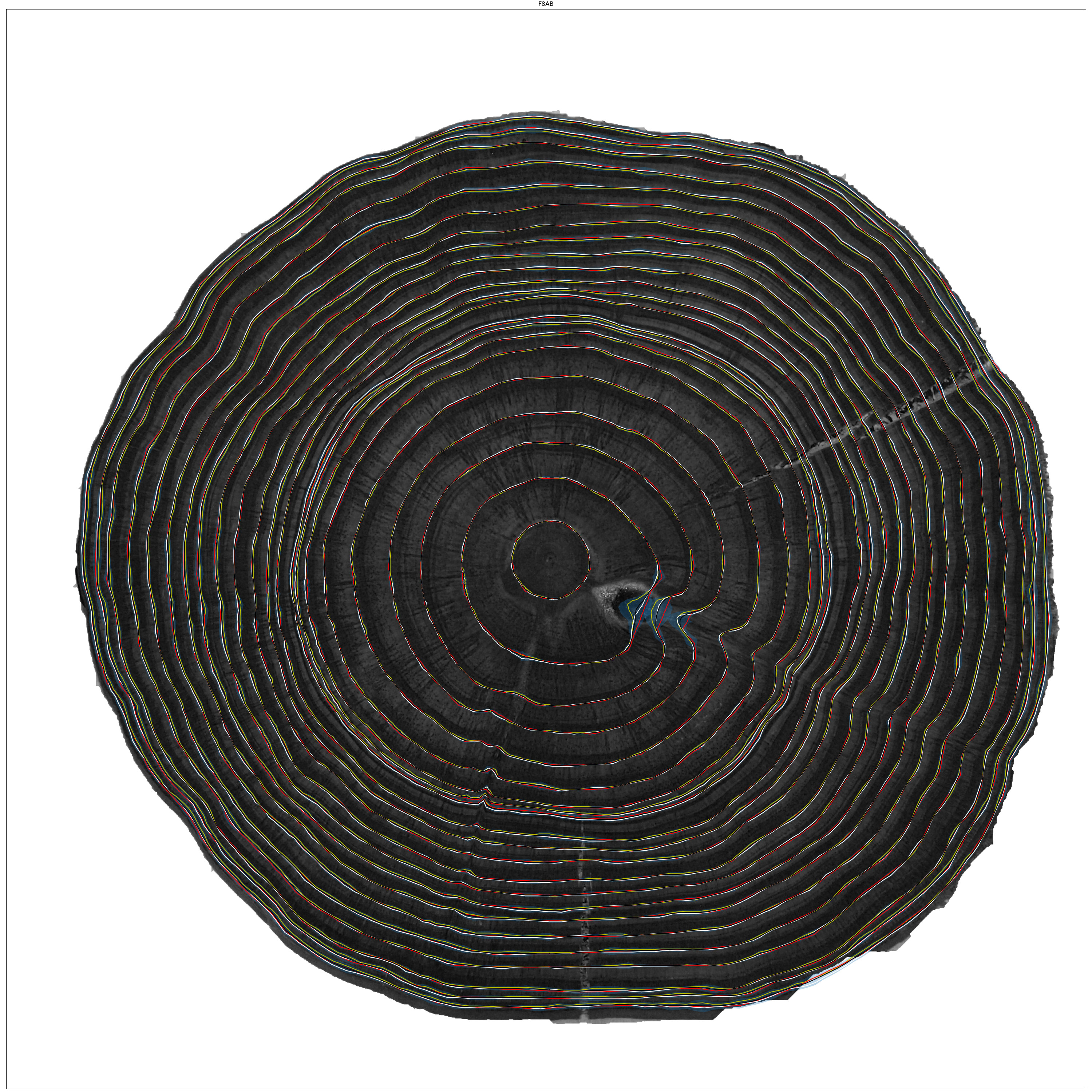}
    \caption{}
    \end{centering}
    \end{subfigure}
    \hfill
    \begin{subfigure}{0.3\textwidth}
    \begin{centering}
    \includegraphics[width=\textwidth]   {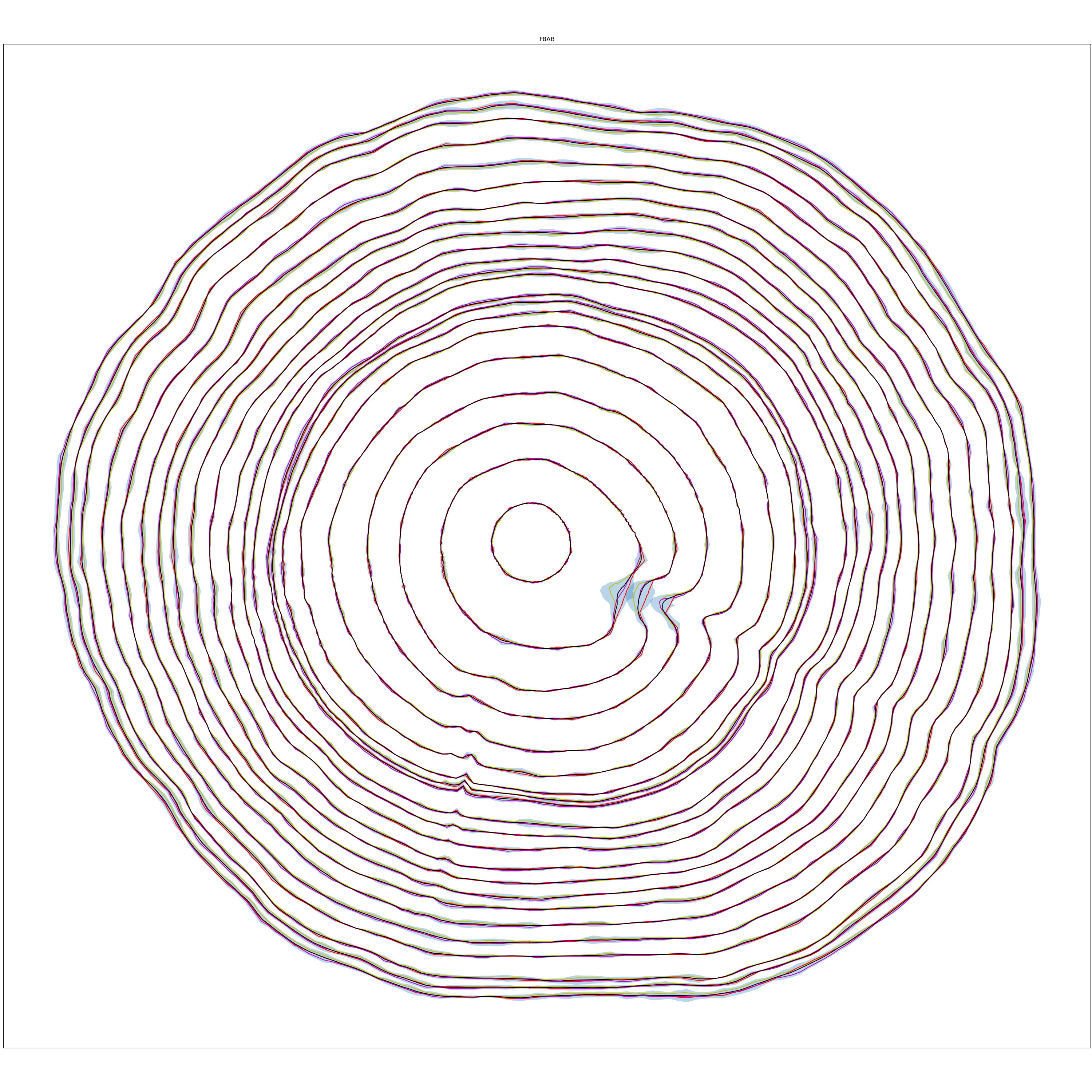}
    \caption{}
    \end{centering}
    \end{subfigure}
    \hfill
    \begin{subfigure}{0.3\textwidth}
    \begin{centering}
    \includegraphics[width=\textwidth]   {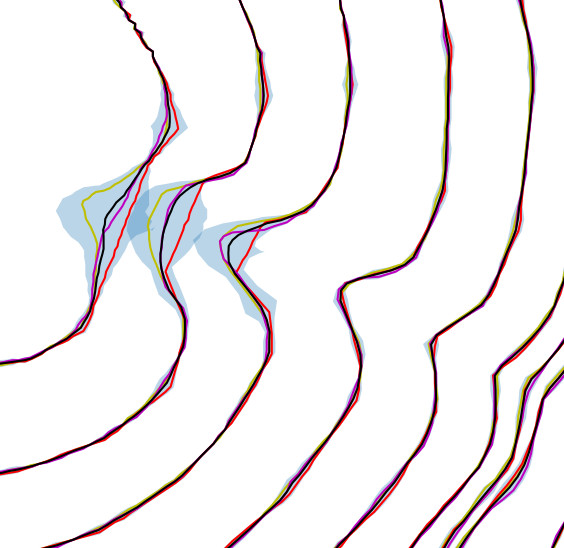}
    \caption{}
    \end{centering}
    \end{subfigure}
    \hfill
   \caption{Manual tracing of image F08a. (a) image with superimposed marks by three experts; (b) marks, (c) a detail. Marks by experts are violet, red, and yellow. In black is the mean of the three expert marks, and the band in blue is the area span by three standard deviations concerning the mean. Note the variability in the tracing by experts.}
   \label{fig:VariabilidadMarcas}
\end{figure}

\begin{figure}[!htbp]
\centering
   \begin{subfigure}{0.3\textwidth}
   \begin{centering}
   \includegraphics[width=\textwidth]{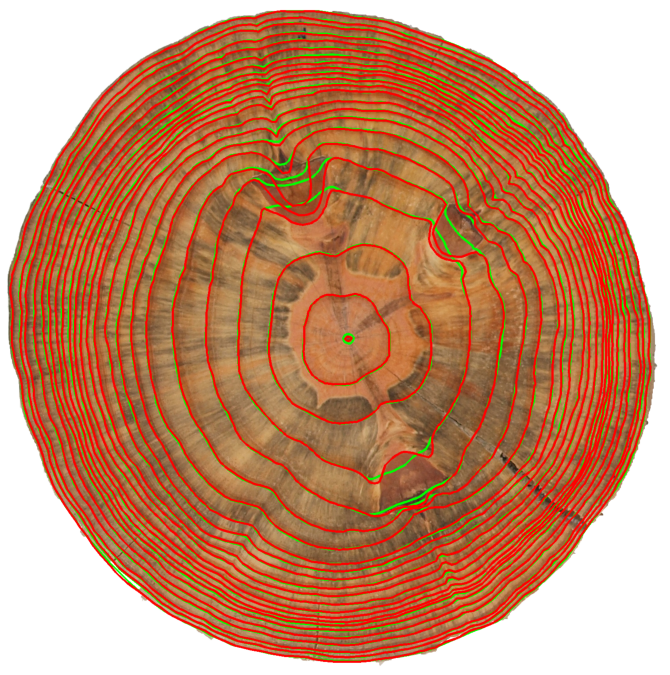}
    \caption{}
    \end{centering}
   \end{subfigure}
    \hfill
    \begin{subfigure}{0.3\textwidth}
    \begin{centering}
    \includegraphics[width=\textwidth]{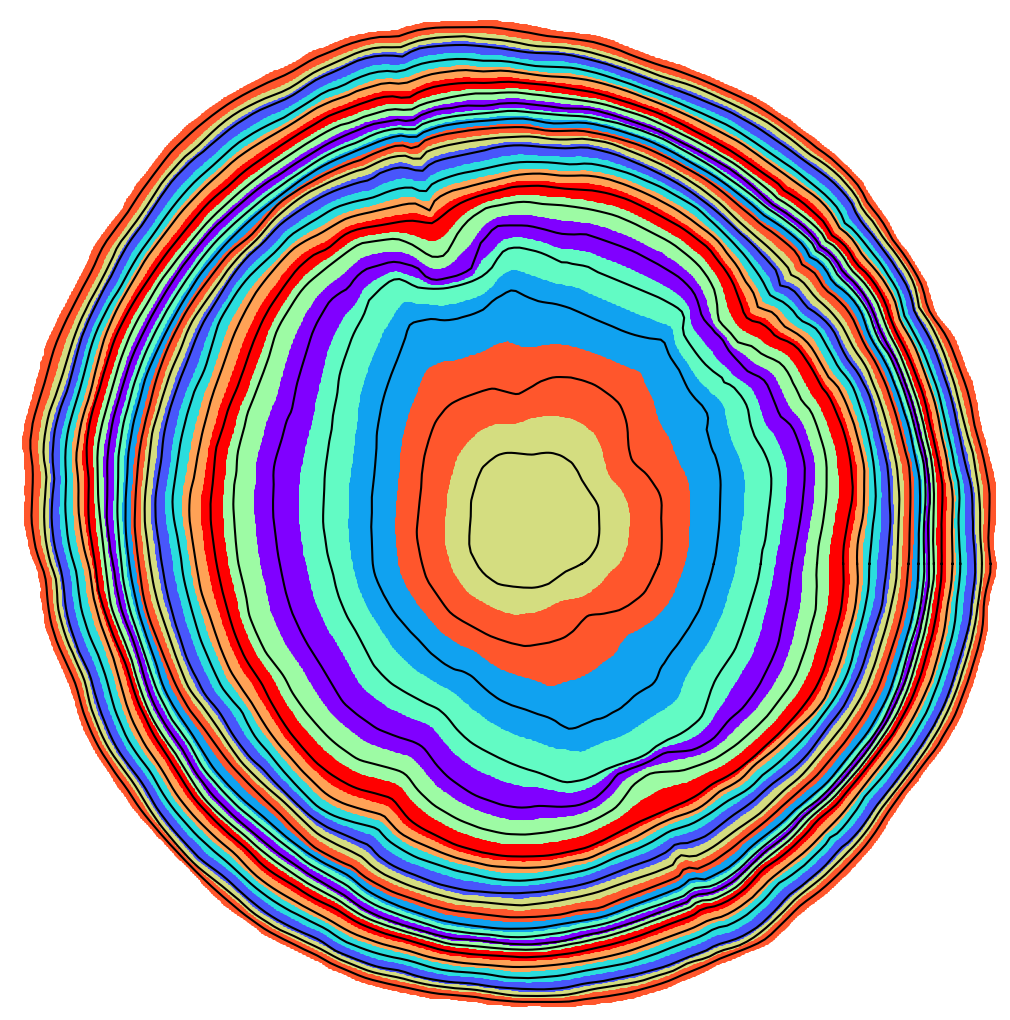}
    \caption{}
    \end{centering}
    \end{subfigure}
    \hfill
    \begin{subfigure}{0.3\textwidth}
    \begin{centering}
    \includegraphics[width=\textwidth]{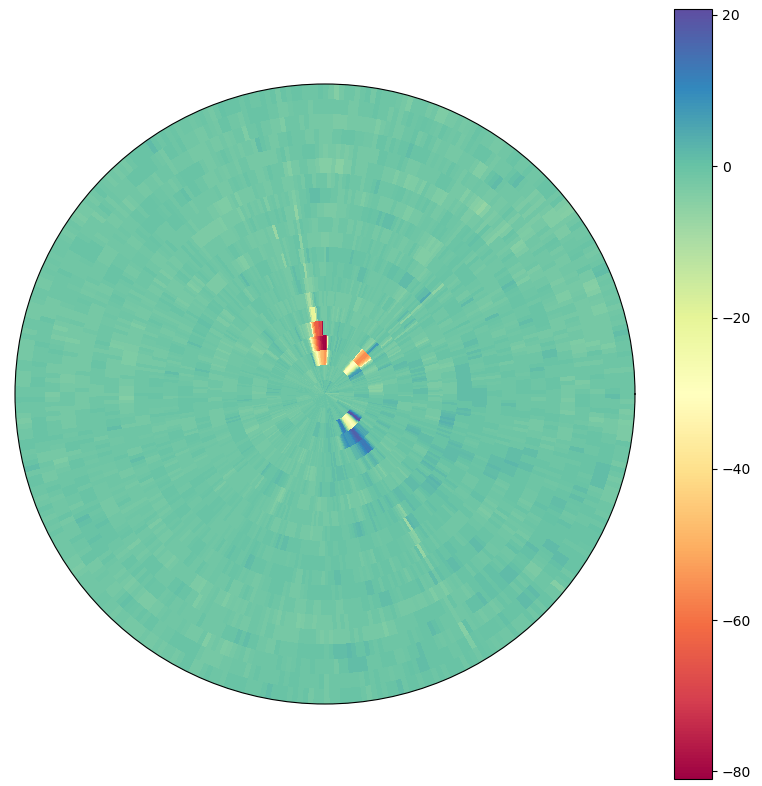}
    \caption{}
    \end{centering}
    \end{subfigure}
    \hfill
    
   \caption{Measuring the error between the expert ring marks and the ground truth for image F07b. a) in green is the ground truth, and in red are the marks produced by the expert. b) areas of influence of the ground truth rings. c) Error, in number of pixels, between the expert marks and the ground truth, see the color bar to interpret error values.}
   \label{fig:MeasuringError}
\end{figure}

\Cref{fig:VariabilidadMarcas} illustrates the ground truth produced by the experts. Most of the traces are highly coincident, even though regions with knots or discontinuities may have significant variability between the marks (see magnified detail). When several experts marked an image, we computed the mean value for each ring, and those curves were set as the ground truth. When only one expert marked an image, those marks were considered the ground truth. We expect that an accurate automatic algorithm can detect most of the tree rings, and  those rings were proximate to the GT rings. 
\Cref{fig:MeasuringError} shows the comparison between the expert marks and the GT. Differences between them are concentrated in some regions of the cross-section, shown in red in \Cref{fig:MeasuringError}c (error in pixels). The \emph{area of influence} of a ring is the region closer to a given curve (\Cref{fig:MeasuringError}b). To assign a ring generated by the algorithm to one in the ground truth we define the criterion of correct likeness between rings: more than $60 \%$ of the nodes of a generated ring lies inside the corresponding GT ring's \emph{area of influence}. When this criterion is met, that ring is included as correctly detected without considering the detection error in pixels.

The data set comprised 64 images in \emph{png} format and a set of manual ring tracings by experts for each image. The marks were saved in a \emph{json} file with a list of shapes of type polygon, each one formed by a list of points in image coordinates (pixels). The name of the \emph{json} file includes the name of the photograph and the letter identifying the expert that made the marks, as such: \emph{Company Code+two digits+height code+-+code-of-expert.json}. The \emph{json} file with the mean values for all experts that marked a given photograph is the name of the image in a \emph{json} format. For example, for the photograph corresponding to the second cross-section of the lumber company (code F) trees, obtained at 200 cm above the ground (code c), the image name is \emph{F02c.png}, the file containing the marks produced by V expert is named \emph{F02c-V.json} and the file containing the GT (mean marks made by all the experts that  marked that image) is named \emph{F02c.json}.

The UruDendro data set is organized in several files, containing the images and the marks as described in \Cref{tab:datasetNames}. The pith data is in file \textit{Pith\_locations.csv}. Indexing by cross-section name in the table gives the pixel coordinate of the pith; as a convention, the center occupies one pixel.

\begin{table}[htbp]
\centering
\caption{The UruDendro data set is grouped in six zip files, including the images and the expert ring traces.}
\begin{tabular}{lllll}\hline
\textbf{Name} & \textbf{Size} & \textbf{Format} & \textbf{Example} & \textbf{Explanation} \\ \hline
Images.zip: & 242 Mb & png & F02c.png & segmented images \\ 
Mean\_gt.zip: & 7 Mb & json & F02c.json & Ground truth annotations \\ 
Annotator\_V\_gt.zip & 892 Kb & json & F02c-V.json & Expert V annotations. \\ 
Annotator\_M\_gt.zip & 1,3 Mb & json & F02c-M.json & Expert M annotations. \\ 
Annotator\_S\_gt.zip & 830 Kb & json & F02c-S.json & Expert S annotations. \\ 
Annotator\_C\_gt.zip & 2,7 Mb & json & F02c-C.json & Expert C annotations. \\ \hline
\end{tabular}
\label{tab:datasetNames}
\end{table}

\subsection{Cross-Section Tree-ring Detection Algorithm}
\label{sec:algoritm}

We use the UruDendro image data set to test the Cross-Section Tree Ring Detection (CS-TRD). 
Here, we describe the approach in general terms, for a detailed explanation please refer to the reference. That article has been submitted to the Image Processing On Line journal and, if accepted, will be published along with the code and the demo, following the principles of the open science paradigm.

The CS-TRD algorithm is heavily based on structural characteristics of the wood, already mentioned by \cite{CerdaHM07}, among others:
\begin{itemize}
    \item The use of the entire cross-section of a tree stem instead of a core. This allows for the study of homogeneity in the whole tree structure.
    \item The following properties generally define the structure of rings on a cross-section:
    \begin{enumerate}
        \item Rings are roughly concentric, even if their shape can be irregular, i.e., two rings cannot intersect.
        \item Many radii can be traced outwards from the pith to the cortex. Those radii will cross each ring only once.
        \item For annual ring detection, the fundamental feature is the boundary determined by transitions from dark to light tones, corresponding to late to early wood transitions on gymnosperms (softwood).
        \end{enumerate}
\end{itemize}

\begin{figure}
\begin{center}
   \includegraphics[width=0.9\textwidth]{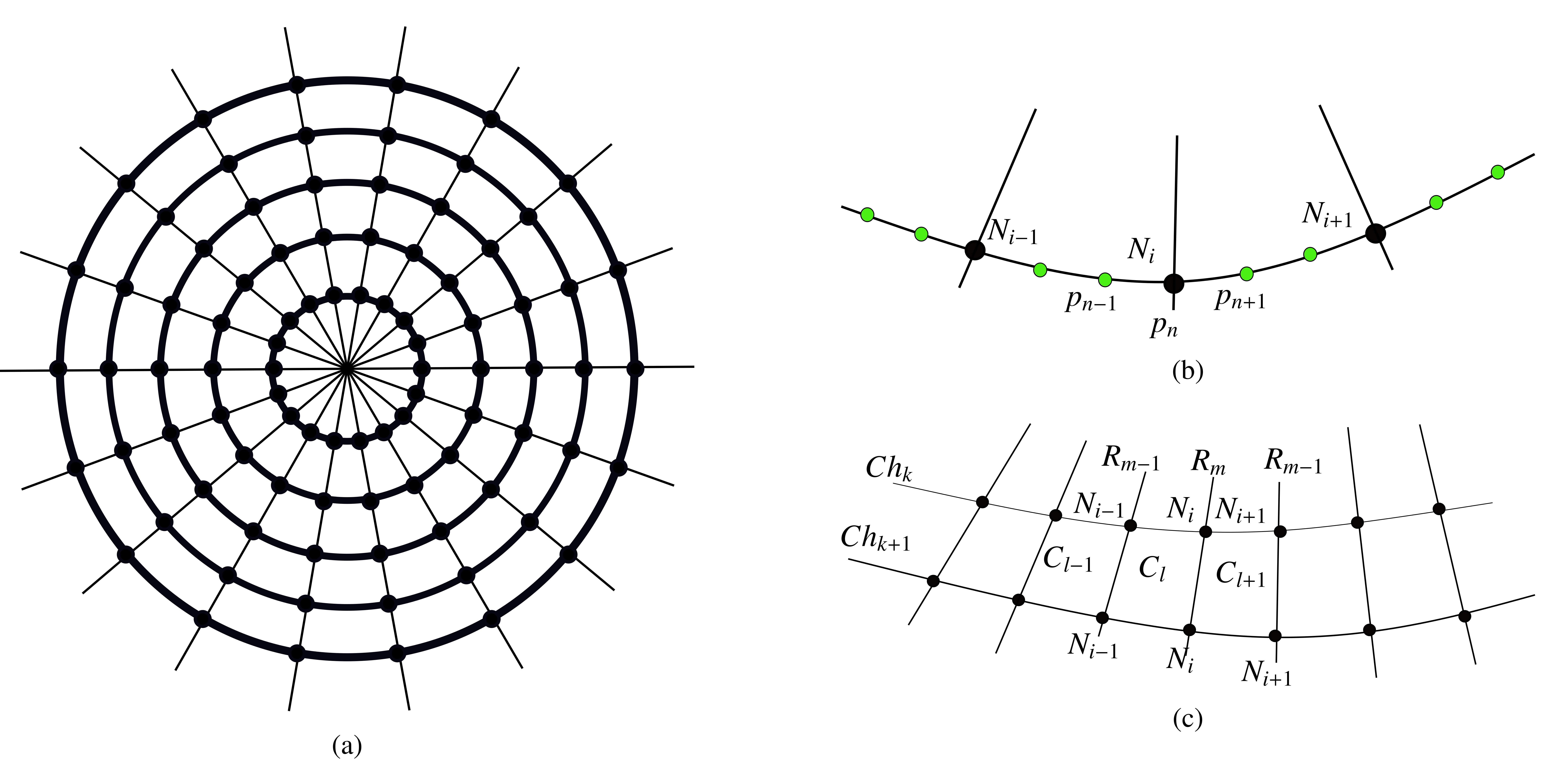}

   \caption{(a) The whole structure, called \textit{spider web}, is formed by a \textit{center} (which corresponds to the cross-section pith), $Nr$ \textit{rays} (in the drawing $Nr=18$) and the \textit{rings} (concentric curves). In the scheme, the \textit{rings} are circles, but in practice, they can be (strongly) deformed as long as they don't intersect another \textit{ring}. Each ray intersects a ring only once in a point called \textit{node}. The area limited by two consecutive \textit{rays} and two consecutive \textit{rings} is named a \textit{cell}. (b) A curve is a set of connected \textit{points} (small green dots). Some of those \textit{points} are the intersection with \textit{rays}, named \textit{nodes} (black dots). A chain is a set of connected \textit{nodes}. In this case, the \textit{node} $N_i$ is the \textit{point} $p_n$. (c) Each \textit{Chain} $Ch_k$ and $Ch_{k+1}$, intersect the  \textit{rays} $R_{m-1}$, $R_{m}$ and $R_{m+1}$ in \textit{nodes} $N_{i-1}$, $N_{i}$ and $N_{i+1}$. Those \textit{rays} and \textit{chains} (as well as the four corresponding \textit{nodes}) determines \textit{cells} $C_{l-1}$, $C_{l}$ and $C_{l+1}$.} 
   \label{fig:definitions}
\end{center}
\end{figure}

\paragraph{Nomenclature} CS-TRD algorithm use the following nomenclature, see~\cref{fig:definitions}. A global structure called \emph{spider web} includes a \emph{center}, associated with the pith, which is the origin of a certain number of \emph{rays}. The \emph{rings} are concentric and closed curves that do not intersect each other. A chain of connected pixels forms each ring. Each radius crosses a chain only once. The \emph{rings} can be viewed as a flexible chain of points with \emph{nodes} at the intersection with the \emph{rays} (larger black dots in the figure, denoted \emph{nodes}). As \cref{fig:definitions}(a) illustrates, a \emph{chain} $Ch_i$ is a set of chained pixels (small green dots in the figure, called \emph{points} and noted $P_i$). The \emph{chains} can move along a \emph{ray} in a radial direction but cannot move in a tangential direction over the \emph{chain}, i.e., they can move along a \emph{ray} as if the \emph{nodes} were hoops sliding along the \emph{rays}. The greater the number $nbR$ of \emph{rays}, the higher precision of the reconstruction of the \emph{rings}. In general $nbR = 360$. Note that this is the ideal setting. In the actual images  \emph{rings} can disappear and do not form closed curves,  \emph{cells} can have varied shapes, given that the \emph{rings} can be deformed, and so on. Figure \ref{fig:definitions}(c) illustrates the nomenclature used in the CS-TRD algorithm.

\begin{figure}
\begin{center}
   \begin{subfigure}{0.3\textwidth}
   \includegraphics[width=1\linewidth]{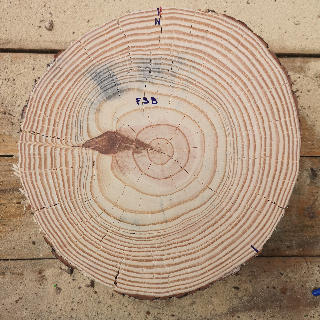}
   \caption{}
   \end{subfigure}   
   \begin{subfigure}{0.3\textwidth}
   \includegraphics[width=1\linewidth]{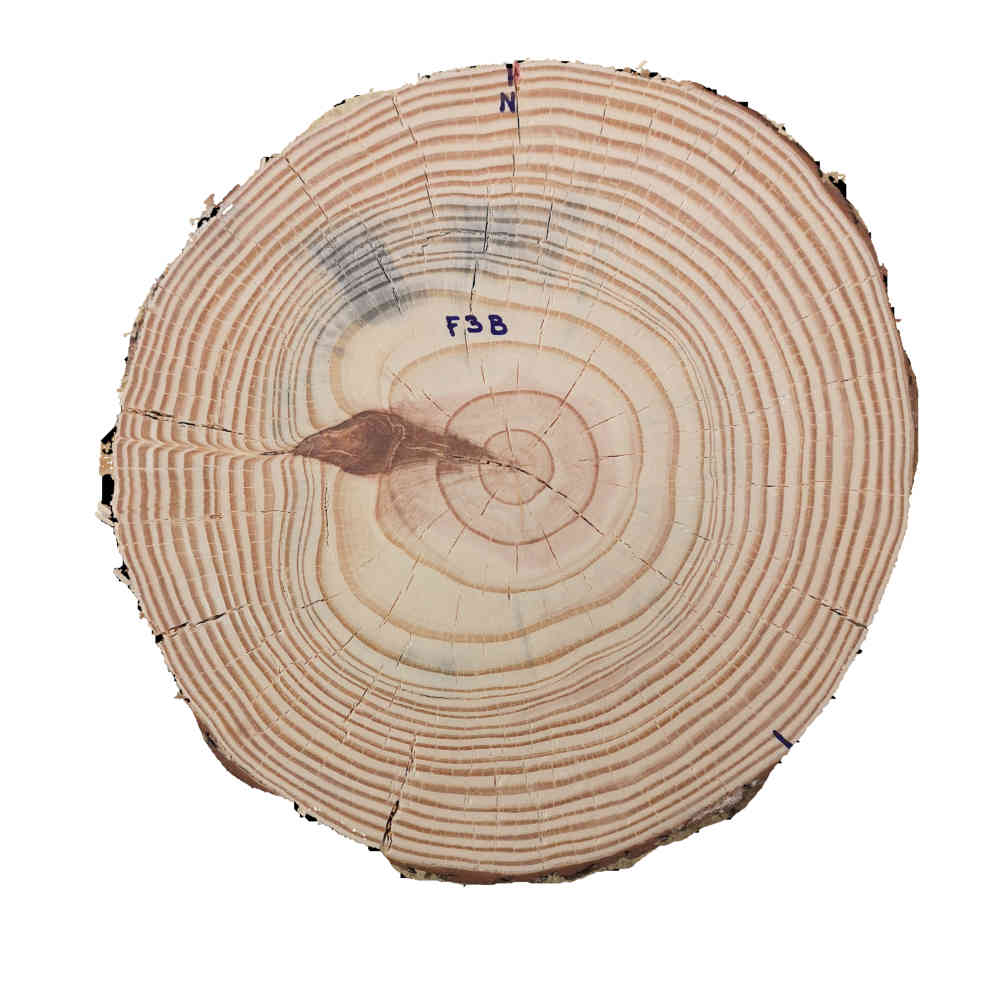}
   \caption{}
   \end{subfigure}   
   \begin{subfigure}{0.3\textwidth}
   \includegraphics[width=1\linewidth]{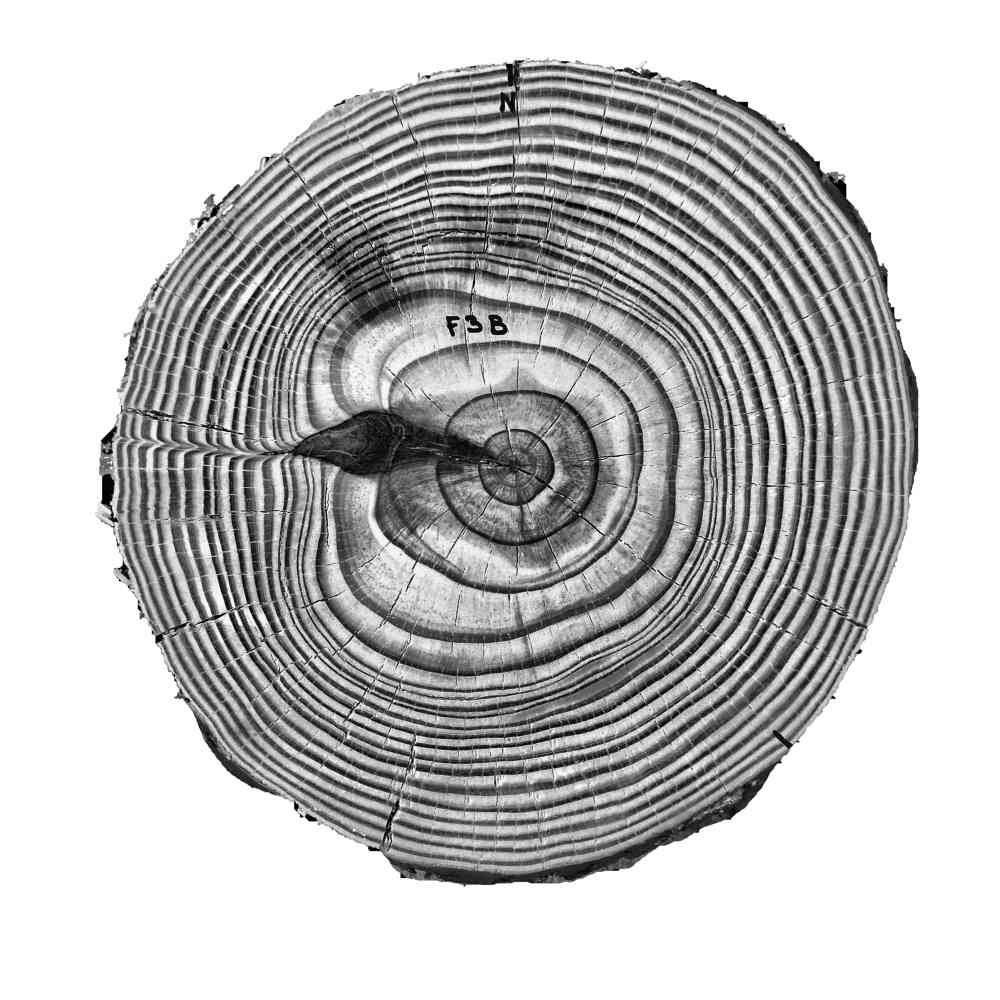}
   \caption{}
   \end{subfigure}   
   \begin{subfigure}{0.3\textwidth}
   \includegraphics[width=1\linewidth]{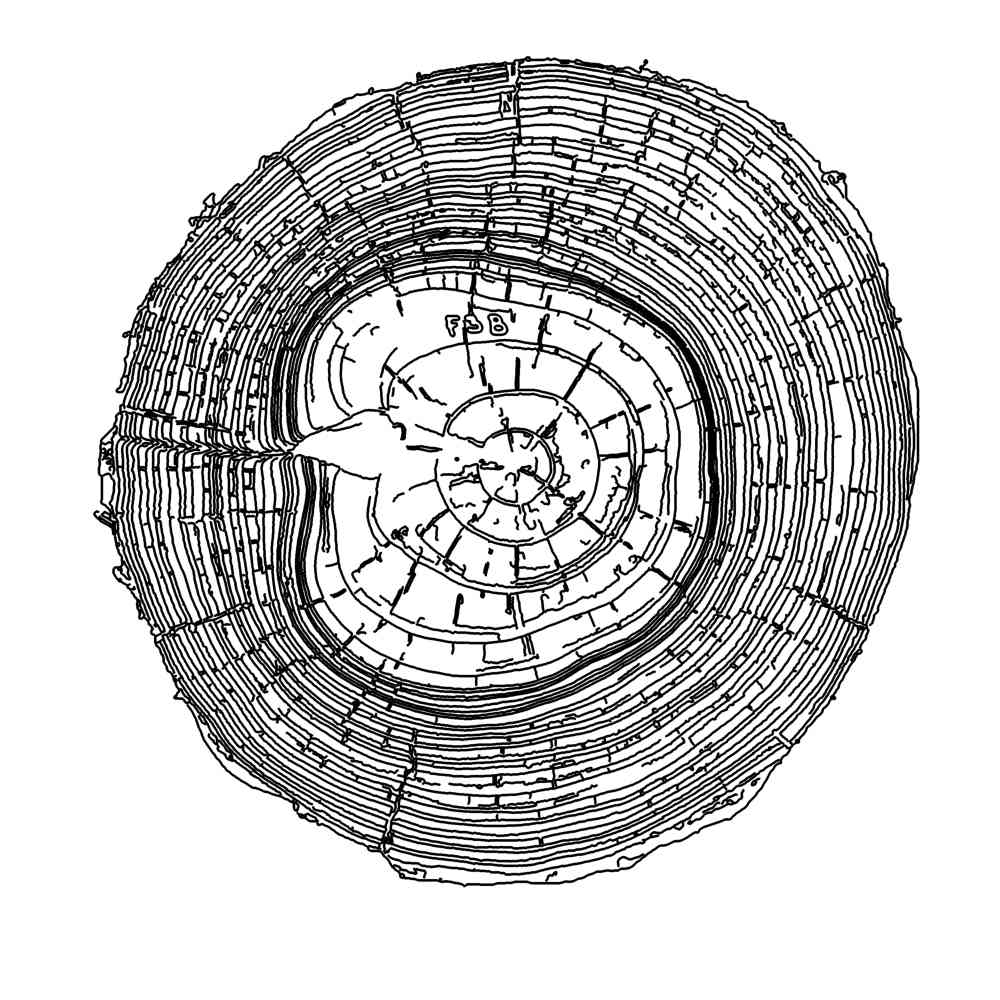}
   \caption{}
   \end{subfigure}   
   \begin{subfigure}{0.3\textwidth}
   \includegraphics[width=1\linewidth]{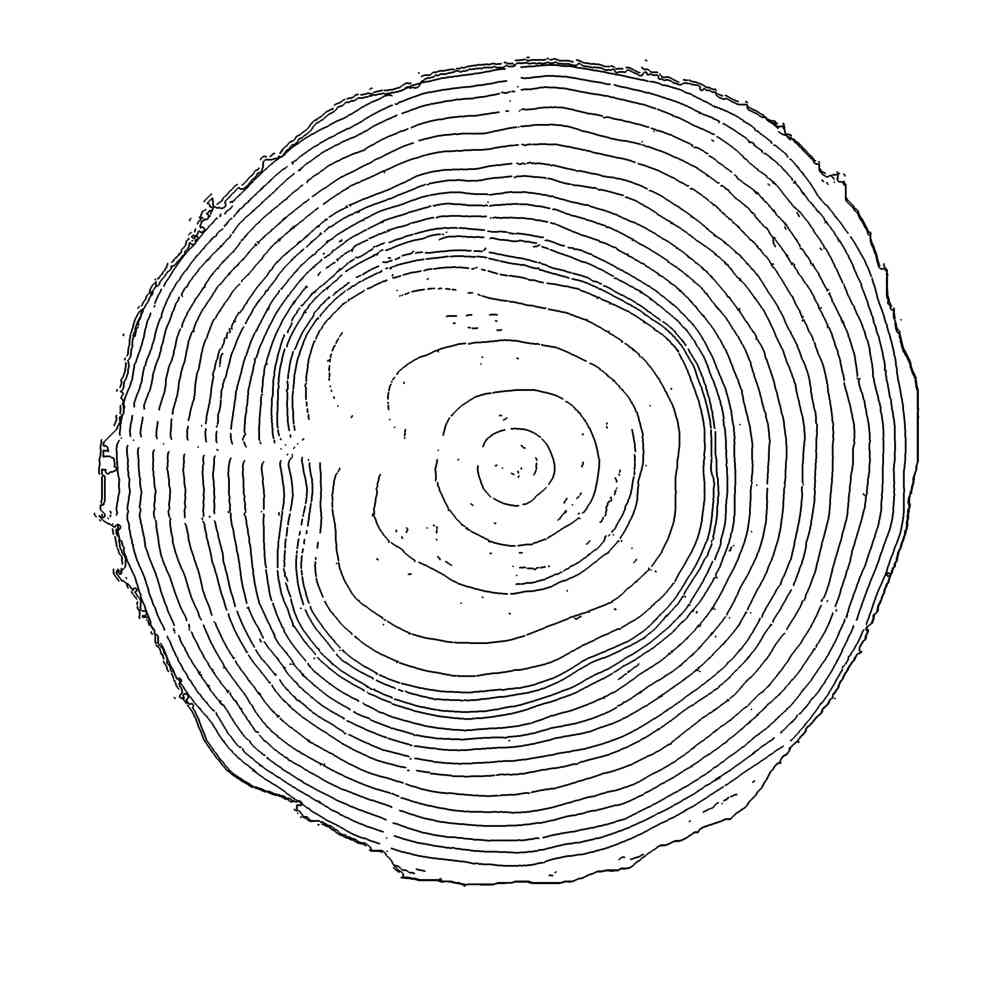}
   \caption{}
   \end{subfigure}   
   \begin{subfigure}{0.3\textwidth}
   \includegraphics[width=1\linewidth]{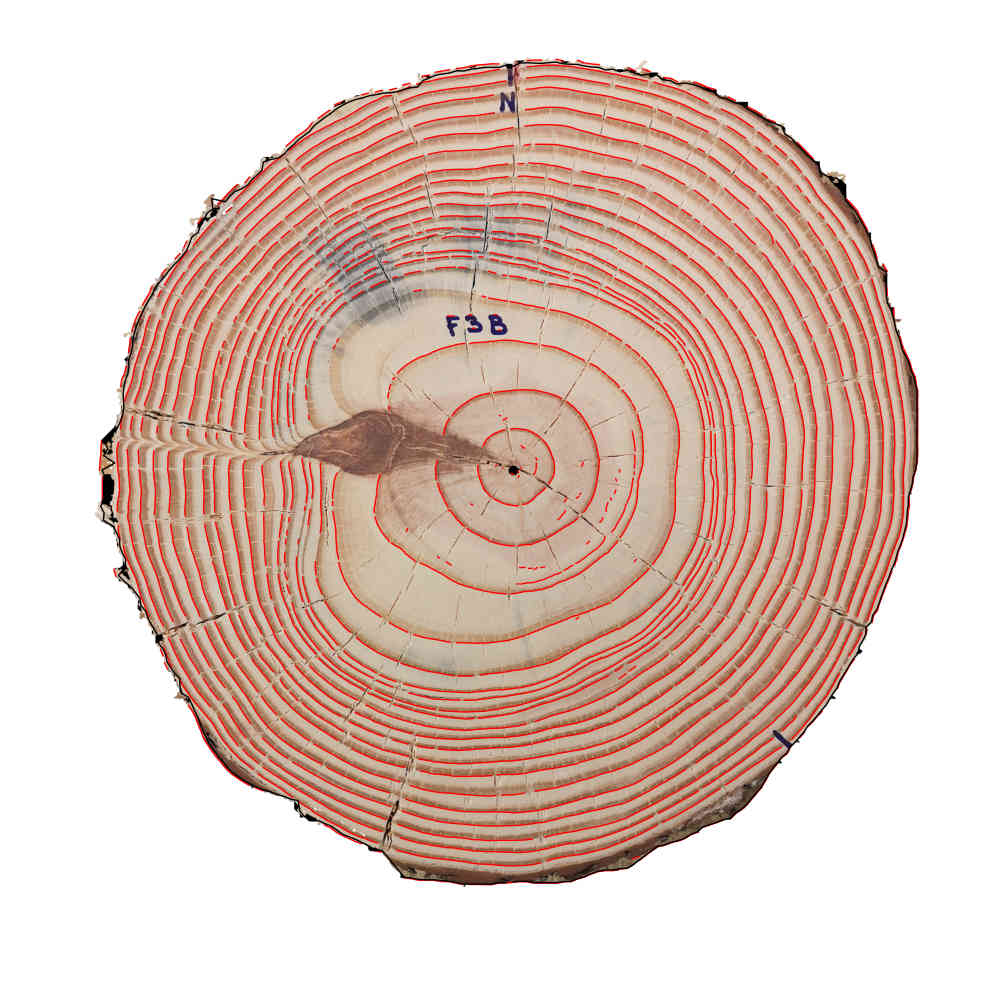}
   \caption{}
   \end{subfigure}   
   \begin{subfigure}{0.3\textwidth}
   \includegraphics[width=1\linewidth]{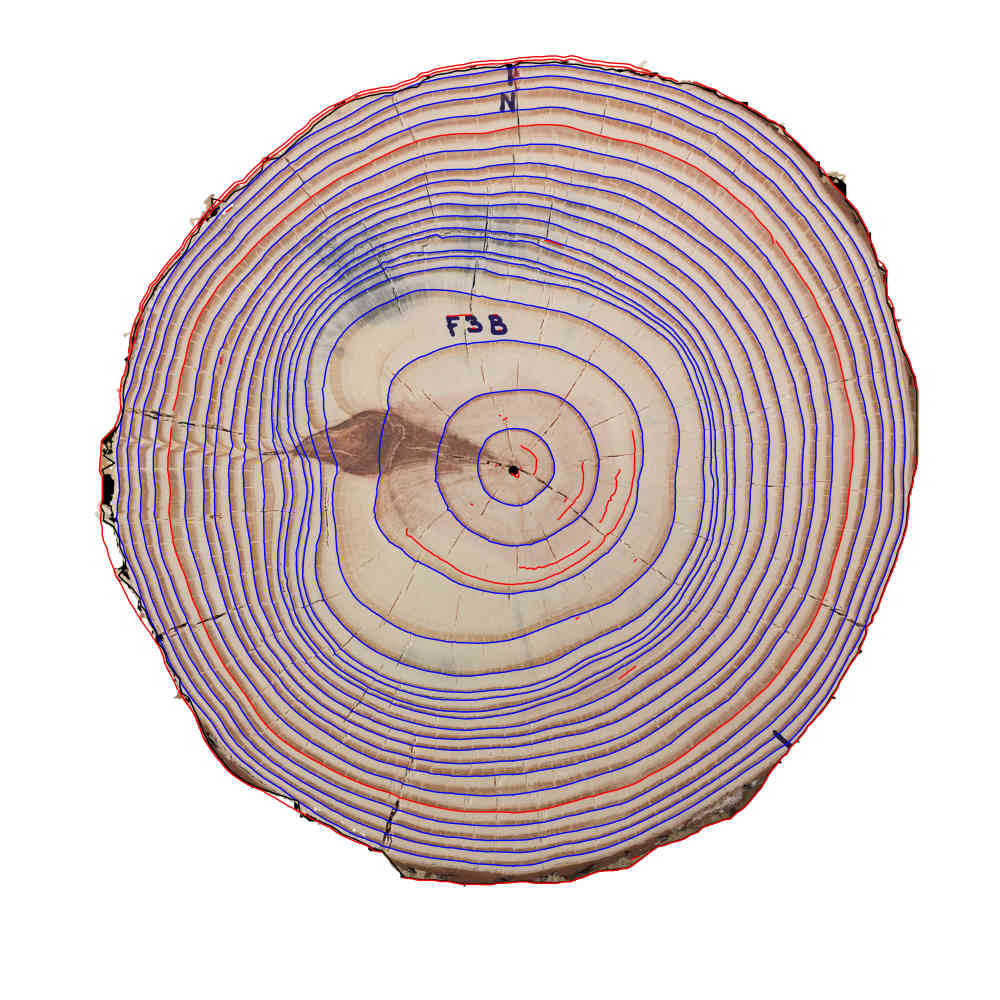}
   \caption{}
   \end{subfigure}   
   \begin{subfigure}{0.3\textwidth}
   \includegraphics[width=1\linewidth]{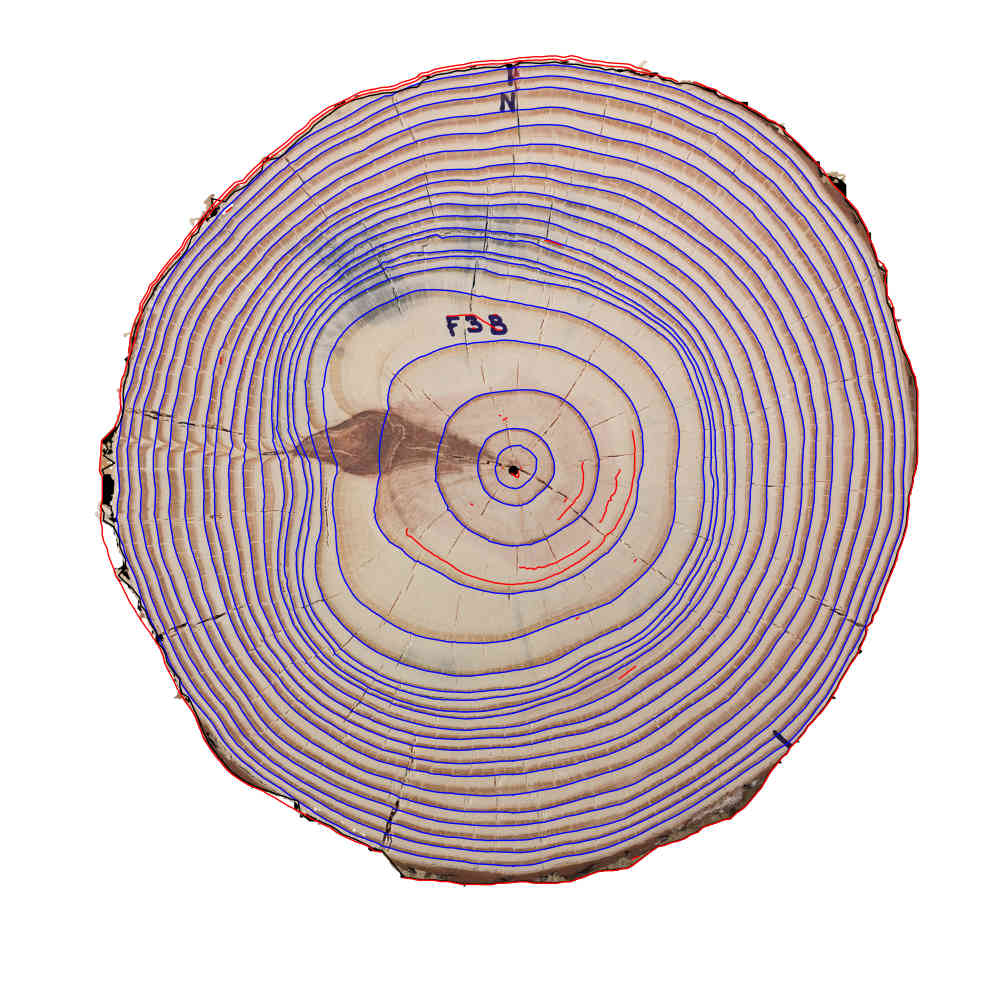}
   \caption{}
   \end{subfigure}   
   \begin{subfigure}{0.3\textwidth}
   \includegraphics[width=1\linewidth]{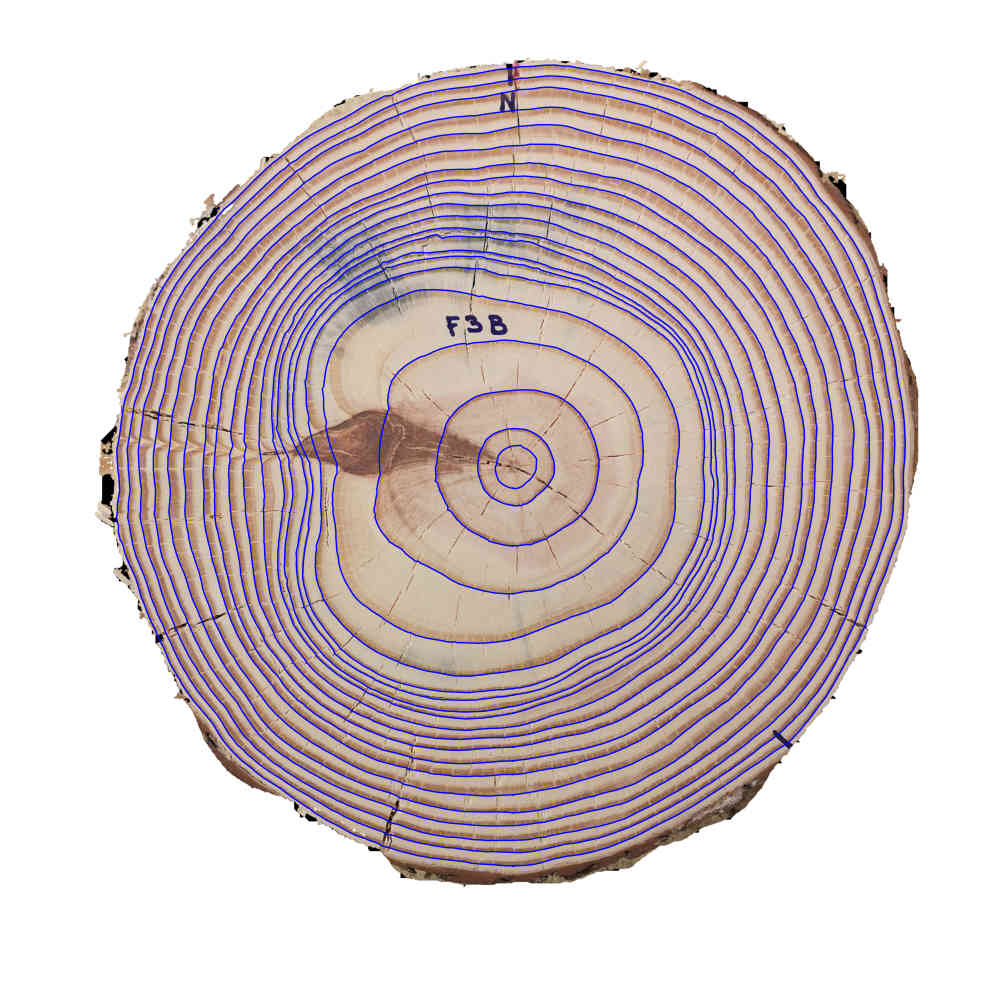}
   \caption{}
   \end{subfigure}   
      \caption{Principal steps of the CS-TRD tree-ring detection algorithm: (a) original image, (b) background extraction, (c) pre-processed image (resized,  equalized, and converted to a grayscale image), (d) the output of the Canny Devernay edge detector, (e) edges filtered by the direction of the gradient, (f) set of detected chains, (g) connected chains, (h) post-processed chains and (i) detected tree-rings.}
   \label{fig:algo}
\end{center}
\end{figure}

\Cref{fig:algo} illustrates the proposed algorithm. The center $c$ of the \textit{spider web} (the tree's pith) is required as an input. Detecting this fundamental point is a problem that can be tackled by automatic means \cite{ipolPith} or marked manually. 

\paragraph{Pre-processing.} The input image must be a cross-section without a background. CS-TRD uses a deep learning-based approach \cite{salientObject} based on two-level nested U-structures ($U^2Net$) to crop the background. \Cref{fig:algo}(b) shows an example of the output of such a procedure. The size of acquired images can vary widely, and this has an impact on the performance. On one side, the bigger the image, the slower the algorithm, as more data must be processed.
On the other hand, if the image is too small, the relevant structures will be challenging to detect. The first step is resizing the input image to a standard size of 1500x1500 pixels. Pith coordinates are resized as well. The RGB image is converted to grayscale and a histogram equalization step is applied to enhance contrast.

\paragraph{Edge detection.} The sub-pixel precision Canny Devernay edge detector  is applied \cite{DevernayIPOL} \cite{Devernay95anon-maxima}. The output of that step is a list of pixel chains corresponding to the edges present in the image. Those edges can be grouped into the following classes:

\begin{itemize}
    \item Edges\_T: produced by the growing process of the tree. It includes the edges that form the rings. Considering a direction from the pith outward, these edges are of two types: ones produced by early to late wood transitions (light to dark transitions on the image), and the late to early wood transitions (transitions from dark to light). The former ones are the annual rings.
    \item Edges\_R: mainly radial and produced by cracks, fungus, or other phenomena.
    \item Other edges produced by wood knots.
    \item Edges produced by noise.
\end{itemize}

\paragraph{Filtering the edge chains.} The edge detector operator gives a vector normal to the edge, and its direction encodes the local direction and sense of the transition. CS-TRD filter out all the points of the edge chains for which the angle between the gradient vector and the direction of the ray touching that point are greater than $30$ degrees. This process filter out all $Edges_T$ produced by the early wood transitions points inward,  and the $Edges_R$, for which the normal vector is roughly perpendicular to the \textit{rays}, see sub figure (e) of \Cref{fig:algo}. Note that this process may break an edge chain into several fragments.

\begin{figure}
\begin{center}
    \includegraphics[width=0.9\textwidth]{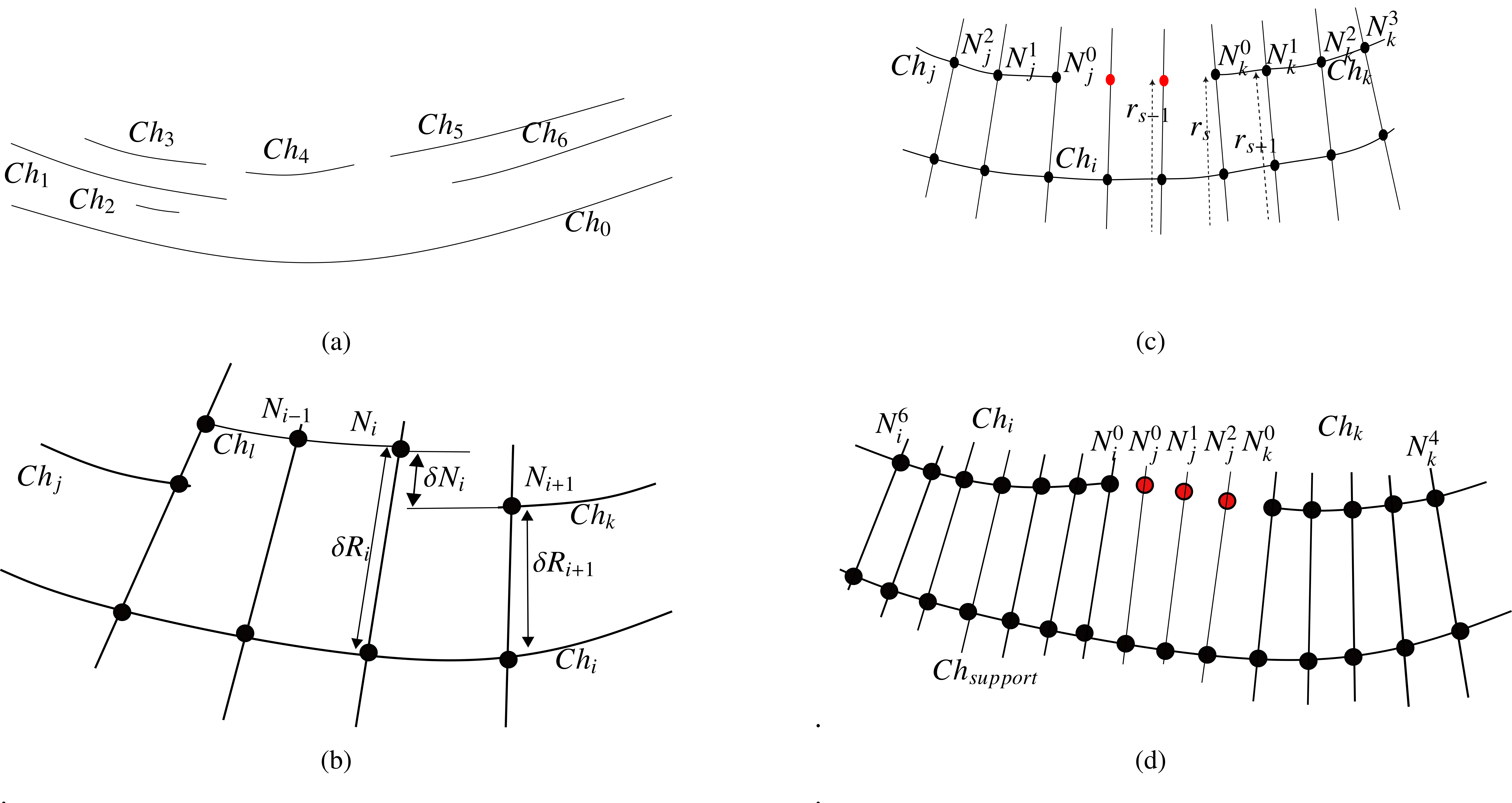}

    \caption{ An illustration of the \textit{connectivity} issue. (a) The question is if endpoint \textit{A} of $Ch_3$ must be connected to endpoint \textit{B} of $Ch_2$ (red dashed line) or to endpoint \textit{B} of $Ch_1$ (blue dashed line). (b) The same question can be posed for endpoint B of $Ch_1$ and endpoint A of $Ch_2$, but $Ch_1$ and $Ch_2$ intersect (the endpoints are crossed by the same \textit{ray}), and this connection is forbidden. Note that the connections are represented by line segments for clarity, but in fact, these are curves in the image space, as  \textit{chain} endpoints are interpolated in polar geometry. (c) Inward and outward chains: a given chain (in black) with two endpoints A and B. Its nodes (in red) appear at the intersection between the Canny Devernay output curve and the rays. The ray at endpoint A is in blue. Other chains detected by Canny Devernay are colored in white. Endpoint A's inward and outward chains are in yellow and orange, respectively.}
   \label{fig:connectivityIssue}
\end{center}
\end{figure}

In order to set an heuristic to connect the filtered chains, CS-TRD use the concepts of \textit{outward} and \textit{inward} neighbor chain. In addition, every chain has two endpoint nodes, A and B. Endpoint A is always the furthest node clockwise, while endpoint B is the most distant node counterclockwise. The \textit{outward} and \textit{inward} neighbor chains are referred to a given chain endpoint (A or B). Given a \textit{chain} endpoint and the corresponding \textit{ray}, CS-TRD finds the first \textit{chain} that intersects that \textit{ray} going from the chain to the center (named as \textit{inward}) and the first \textit{chain} that intersects that \textit{ray} going from the chain moving away from the center (named as \textit{outward}). Figure \ref{fig:connectivityIssue}(c) illustrate this. Chains are superposed over the gray-level image. The ray at endpoint A is in blue, the nodes are in red at the intersection between the rays, and the chains are in orange, black, and yellow. Orange and yellow chains are the \textit{visible} chains for the black chain at endpoint A (outward and inward, respectively).

\paragraph{Connect chains.}\label{connect} The next step of CS-TRD is to group the set of chains to form the rings. Some of these chains are spurious, produced by noise, small cracks, knots, etc., but most are part of the desired rings, as seen in Figure \ref{fig:algo}. To connect chains, CS-TRD must decide if the endpoints of two neighboring chains can be connected, as illustrated by \Cref{fig:connectivityIssue}. It uses a support chain, $Ch_0$ in the figure, to decide whether or not those chains must be connected.

In order to group chains that belong to the same ring, CS-TRD proceed as follows:
\begin{enumerate}
    \item Order all the chains by length and begin by processing the longest. The processed chain is called \textit{Chain support}, $Ch_i$. Once finish merging all the possible \textit{candidate chains} related to $Ch_i$ ($candidates_{Ch_i}$), continue with the next longest \textit{chain}.

    \item Find the chains that are visible from the \textit{Chain support} inwards (i.e., in the direction from \textit{Chain support} to the center). Visible means that a \textit{ray} that goes through the endpoint of the \textit{candidate chain} crosses the \textit{chain support} without crossing any other \textit{chains} in between. The set of \textit{candidate chains} of the \textit{Chain support} $Ch_i$ is named \textit{$candidates_{Ch_i}$}.  This is illustrated by Figure \ref{fig:candidatas}.a, in which case, the \textit{chains candidates} generated inwards by $Ch_0$ is:
    $$candidates_{Ch_0} = \{Ch_1, Ch_2, Ch_4, Ch_5, Ch_6\} $$
    \textit{Chain} $Ch_3$ is shadowed  by $Ch_1$ and $Ch_5$ is not shadowed by $Ch_6$ because at least one of its endpoints are visible from $Ch_0$. The same process is made for the chains visible from the \textit{Chain support} outwards.

\begin{figure}
\begin{center}
   \includegraphics[width=0.9\textwidth]{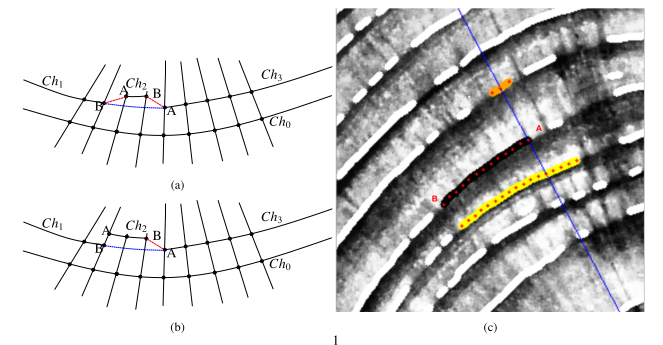}

    \caption{ (a) For the \textit{chain support} $Ch_0$, the set of  \textit{chain candidates} is formed by $Ch_1$, $Ch_2$, $Ch_4$, $Ch_5$ and $Ch_6$. \textit{Chain} $Ch_3$ is shadowed  by $Ch_1$ but $Ch_5$ is not shadowed by  $Ch_6$ because at least one endpoint of $Ch_5$ is visible from $Ch_0$. Note that a \textit{chain} becomes part of the \textit{candidate chains set} if at least one of its endpoints is visible from the \textit{chain support}. (b) Quantities used to measure the connectivity between \textit{chains}. $\delta R_i$ is the radial difference between two successive \textit{chains} along a \textit{ray} $R_i$ and $\delta N_i$ is the radial difference between two successive \textit{nodes} $N_i$ and $N_{i+1}$. Note that these nodes can be part of the same \textit{chain} or be part of two different \textit{chains} that may be merged. Support chains are represented with the name $Ch_i$. $Ch_i$ visible chains are $Ch_j$, $Ch_l$ and $Ch_k$. Chains $Ch_j$ and $Ch_k$ satisfy similarity conditions.  (c) The nomenclature used for the connect chains algorithm.   Given the support chain, $Ch_{i}$,  chains $Ch_j$ and $Ch_k$ are candidates to be connected. $N_j^n$ are the nodes of $Ch_j$, with $n=0$ for the node corresponding to the endpoint to be connected. Similarly, we note $N_k^n$ the nodes of $Ch_k$. In red are the nodes created by an interpolation process between both endpoints. We represent the radial distance to the center of $Node^s$ as $r_s$. (d) Nomenclature used for the merging chains algorithm. Given the $Ch_{support}$ chains $Ch_i$ and $Ch_k$ are candidates to merge. $N_i^n$ are the nodes of $Ch_i$, with $n=0$ for the node corresponding to the endpoint to be connected. similarly we note $N_k^n$ the nodes of $Ch_k$. In red are the nodes created by an interpolation process between both endpoints, named $N_j^n$.}  
   \label{fig:candidatas}
\end{center}
\end{figure}

    \item The set \textit{$candidates_{Ch_i}$} is explored, searching for connections between them. By construction, the \textit{chain support} is not a candidate to be merged in this step. From the endpoint of a chain, the algorithm moves forward. The next endpoint of a non-intersecting \textit{chain} in the \textit{$candidates_{Ch_i}$} set is a candidate to be connected to the first one. Two \textit{chains} intersect if there exists at least one \textit{ray} that cross both \textit{chains}. For example, in Figure \ref{fig:candidatas}.a, $Ch_6$  intersects with $Ch_5$ and non-intersects with $Ch_4$. To decide if both chains should be connected, a \textit{connectivity goodness} measure between them is calculated.
    \label{connect:item_recorrer_cadenas}
    \item The connectivity goodness measure combines three criteria:
    
    \begin{enumerate}
        \item \emph{Radial tolerance for connecting chains (RadialTol)}. The radial difference between the distance from each chain to be merged (measured at the endpoint to be connected) and the support chain should be small. For example, in figure \ref{fig:candidatas}.b, to connect node $N_i$ of $Ch_l$ and node $N_{i+1}$ of $Ch_k$, must be verified that $| \delta R_i - \delta N^{i+1}_i |  < Th_{rt}$. Where $Th_{rt}$ is a parameter. 

        \item \textit{Similar radial distances of nodes in both chains (SimilarRadialDist)}. For each chain,  a set of nodes is defined. 
        The radial distance between a node in the given chain and the corresponding node for the same ray in the support chain,  $\delta R_i$, is measured. See \Cref{fig:candidatas}.b. This defines two sets, one for each considered chain $i$ and $k$:
        $Set_j = \{ \delta R_j^0,..,\delta R_j^{n_{nodes}}\}$ and  $Set_k = \{ \delta R_k^0,..,\delta R_k^{n_{nodes}}\}$. The mean and standard deviation $Set_j(\mu_j,\sigma_j)$ and $Set_k(\mu_k,\sigma_k)$ are calculated. A parameter \textit{$Th_{ds}$} defines a range of radial distances associated with each chain: $Range_j = (\mu_j - \textit{$Th_{ds}$}*\sigma_j, \mu_j +  \textit{$Th_{ds}$}*\sigma_j )$ and $Range_k = (\mu_k -   \textit{$Th_{ds}$}*\sigma_k, \mu_k +   \textit{$Th_{ds}$}*\sigma_k )$. A non-null intersection between both distributions: $Range_j \cap Range_k \neq 0$ allows for a connection between chains $i$ and $k$.

        \item \textit{Regularity of the derivative (RegularDeriv)}. Given two chains $Ch_j$ and $Ch_k$ that can be connected and a set of interpolated nodes between their endpoints ($Ch_{jk}$), see \Cref{fig:candidatas}.d. The new virtual chain created by the connection between chains $Ch_j$ and $Ch_k$  encompassed the nodes of those two chains and the new interpolated nodes between both chains ($Ch_{jk}$, colored in red in the figure). To test the regularity of the derivative, a set of nodes for each concerned chain is defined. For chain $Ch_j$, this set is $\{ N_j^0, N_j^1,..., N_j^{n_{nodes}}\}$ where $n_{nodes}$ is the number of nodes to be considered. For each chain,  the centered derivative in each node is computed,  $\delta N^s = \frac{\|r_{s+1} - r_{s-1}\|}{2}$, where $r_s$ is the Euclidean distance between the node and the center. 
        
        The  derivatives for the nodes of the existing chains are $$Der(Ch_j, Ch_k) = \{\delta N_j^0, ..., \delta N_j^{n_{nodes}}, \delta N_k^0, ..., \delta N_k^{n_{nodes}} \} $$
        The condition is asserted if the maximum of the derivatives in the interpolated chain is less or equal to the maximum of the derivatives in the two neighboring chains times a given tolerance: $$max(Der(Ch_{jk})) \leq max ( Der(Ch_j,Ch_k) ) \times Th_{rd}$$ Where $Th_{rd}$ is a parameter.  
    \end{enumerate}

If     $\textit{RegularDeriv} \land ( \textit{SimilarRadialDist} \lor \textit{RadialTol} )$ is asserted, then chains $Ch_j$ and $Ch_k$ can be connected, where $\lor$ and $\land$ stands for the logical \textit{or} and \textit{and} symbols, respectively.

    Finally, no other chain can exist between both chains to be connected. If another chain exists in between, it must be connected to the closer one.

    This process is iterated for the whole image. In each iteration, the parameters are gradually relaxed. In the first iteration, there are a lot of small chains, but as the process progress, the concerned chains are larger and less noisy. Once the merging process advances, the relaxed parameters allow connecting more robust chains. 
    
    \item The same process is made in the outward direction.

\end{enumerate}

\subsection{Pith detection}
\label{sec:pith}

The algorithm requires a pith position as an input. It can be marked manually or identified by another algorithm using, for example, an automatic detection approach \cite{ipolPith} that is publicly available at the IPOL cite\footnote{https://www.ipol.im/pub/art/2022/338/}. 

\subsection{Metric}
\label{sec:metric}

To evaluate the performance of an automatic detection method for the data set, we developed a metric based on the one proposed by Kennel et al., \cite{KennelBS15}. Given a set of detections, \textbf{D}, we assigned each detection $d \in \textbf{D}$ to some ground-truth (GT) ring, $g$, in the set \textbf{G}. Additionally, each detection and GT ring has $Nr$ nodes. We needed the values of both the GT curves and the detected rings on the nodes corresponding to the $Nr$ rays to compute tree-ring metrics. To this aim, we sampled all the curves at those positions.

\begin{figure}[!htbp]
\begin{centering}
    \begin{subfigure}{0.3\textwidth}
    \begin{centering}
   \includegraphics[width=\textwidth]{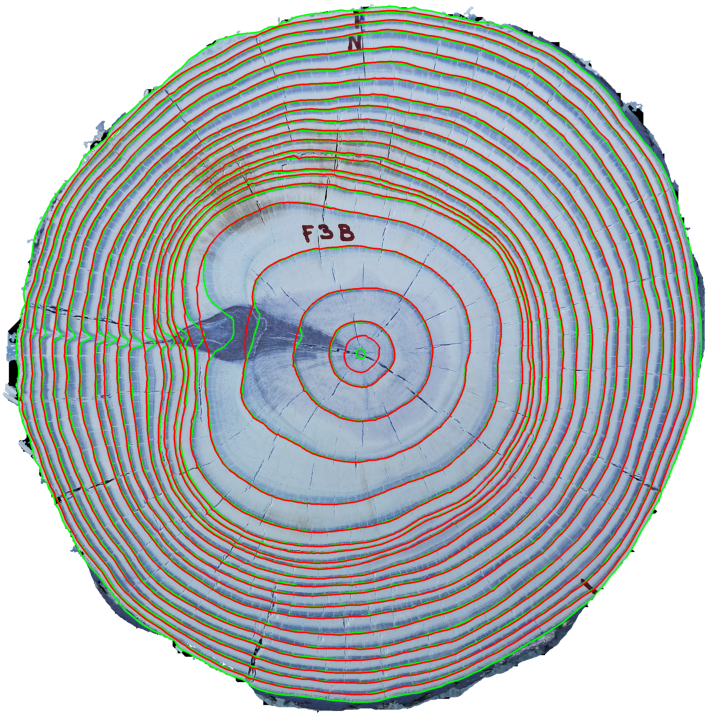}
    \label{fig:dtvsgt}
    \caption{}
    \end{centering}
    \end{subfigure}
    \hfill
    \begin{subfigure}{0.3\textwidth}
    \begin{centering}
   \includegraphics[width=\textwidth]{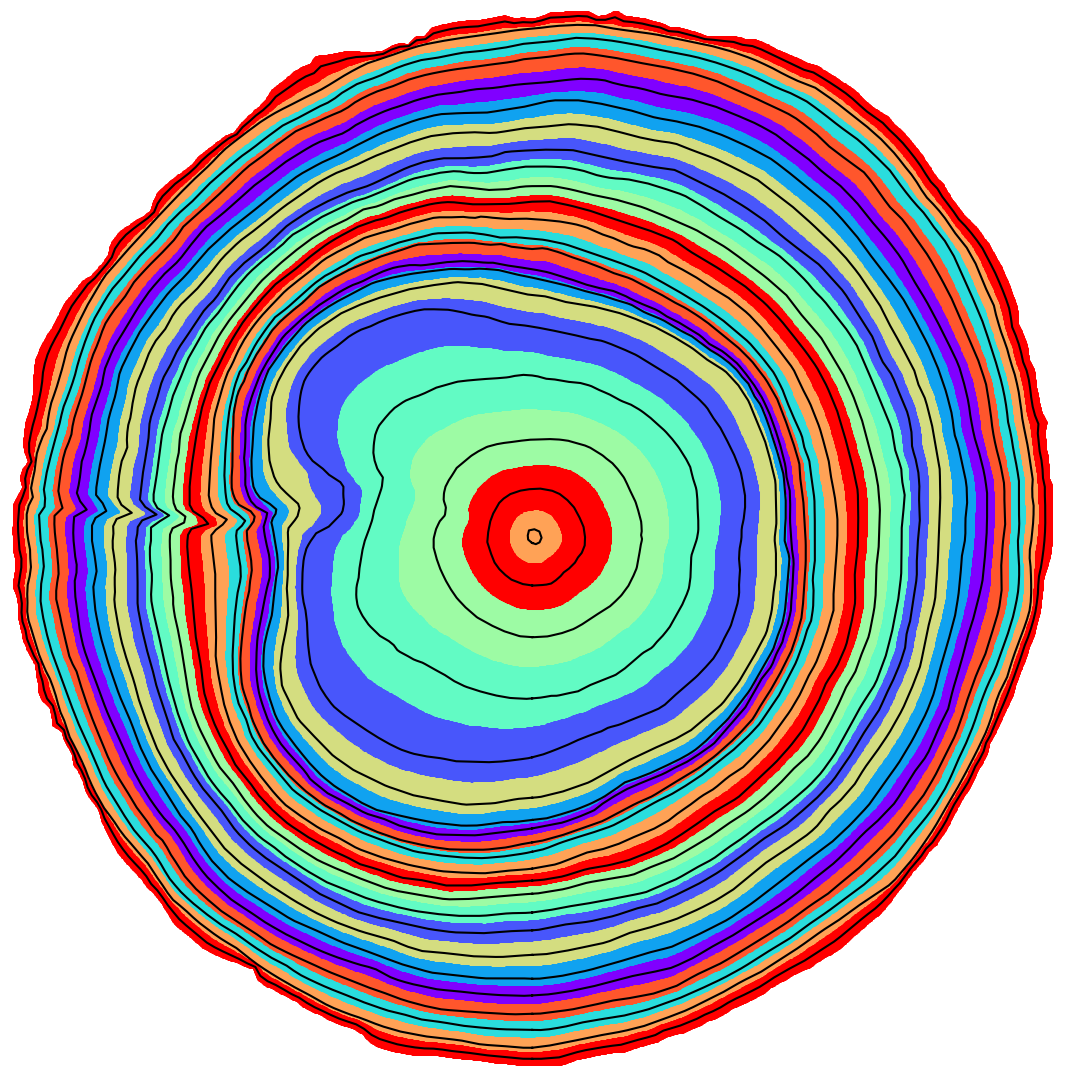}
    \label{fig:influence}
    \caption{}
    \end{centering}
    \end{subfigure}
    \hfill
       \begin{subfigure}{0.3\textwidth}
    \begin{centering}
   \includegraphics[width=\textwidth]{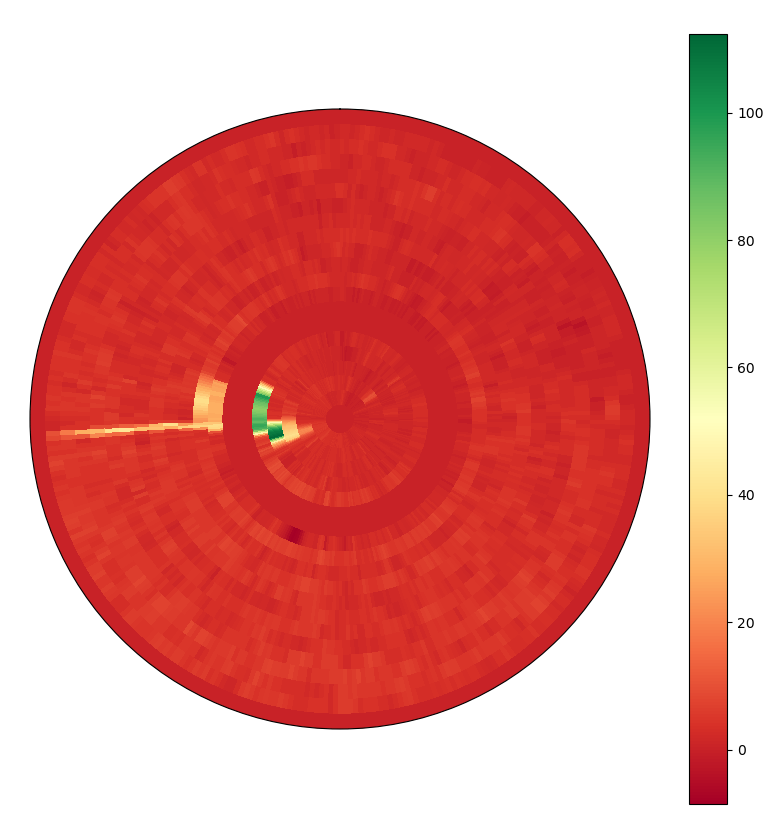}
    \label{fig:error}
    \caption{}
    \end{centering}
    \end{subfigure}
    \hfill
   \caption{Measuring the error between automatic detection and the GT for image F03d. (a) the GT in green;  the rings produced by CS-TRD in red. (b) Areas of influence of the GT rings. (c) Error, in pixels, between the detection and the GT.}
   \label{fig:influece_area}
\end{centering}
\end{figure}

To assign a ring as detected, we defined an influence area for each GT ring as the set of pixels that have the mentioned ring as the most proximate.  This criterion generates a band that surrounds the ring from its inner and outer sides. The influence areas cover the whole cross-section (see \Cref{fig:influece_area}(b)). For each ray, the frontier between influence areas was the middle point between the nodes of consecutive GT rings. \Cref{fig:influece_area} illustrates the performance over the F03d image. \Cref{fig:influece_area}(a) shows the red detections and the green GT traces. \Cref{fig:influece_area}(b) shows the influence area. Each GT ring was colored black and was the center of its influence area. To assign a detected curve to a GT ring, we had to guarantee that the given detection was the closest one to the ring and that it was close enough. The distance (in pixels) between a detection and a GT ring was defined by \Cref{equ:rmse}. 
\begin{equation}
    Dist = \sqrt{\frac{1}{Nr}\sum_{i=0}^{Nr-1}\left(dt_i-gt_i\right)^2}\hspace{0.5cm} 
    \label{equ:rmse}
\end{equation}

Where $dt_i$ is the radial distance of the detection and $gt_i$ the radial distance of the corresponding GT ring, for the same node $i$.

Figure \ref{fig:influece_area}(c) shows the error in pixels between the GT rings and the detected curves assigned to them. The red color represents a low error, while the yellow-green color represents a high error. The error is concentrated around the knot, which compromises the precise detection of some rings.  These type of heat-maps are constructed with an homogeneous radial distribution of rings, highlighting the error at a given ring, not a coordinate of the image.

\begin{figure*}[!htbp]
\begin{centering}
    \begin{subfigure}{0.3\textwidth}
    \begin{centering}
   \includegraphics[width=\textwidth]{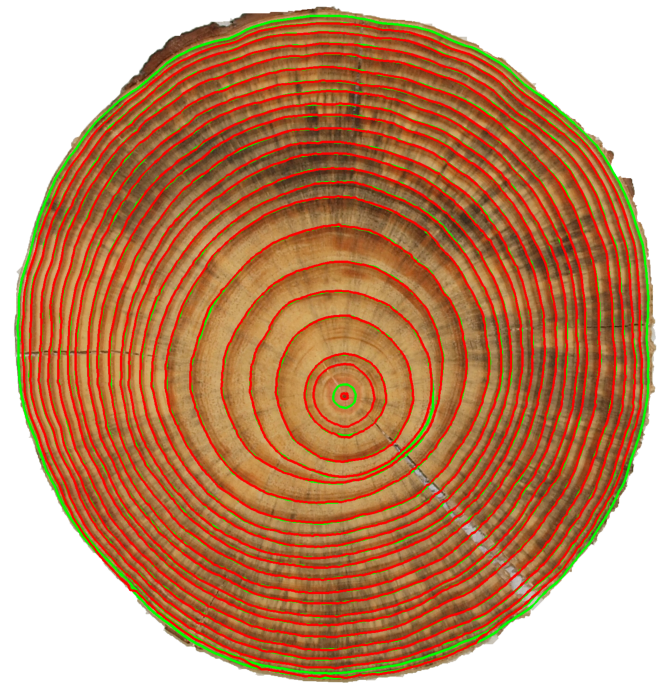}
    \label{fig:f02b_dt_and_gt}
    \caption{}
    \end{centering}
    \end{subfigure}
    \hfill
    \begin{subfigure}{0.3\textwidth}
    \begin{centering}
   \includegraphics[width=\textwidth]{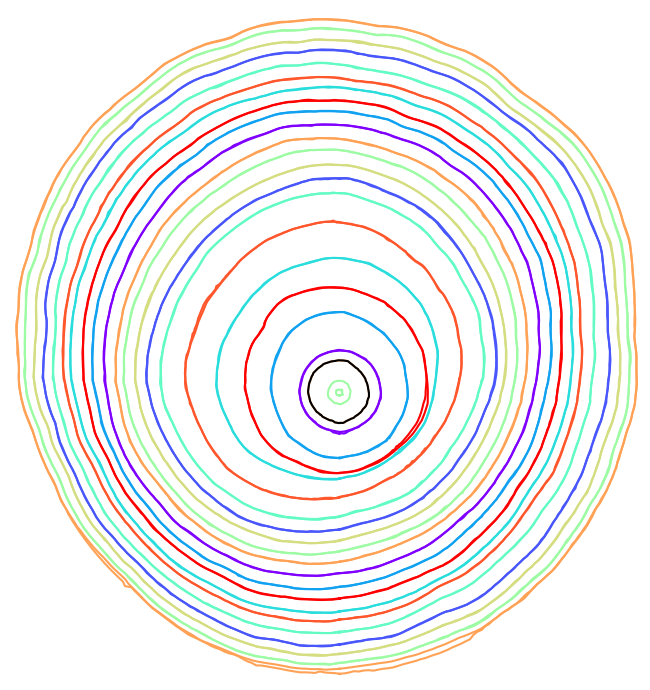}
   \label{fig:f02b_asignacion}
    \caption{}
   \end{centering}
    \end{subfigure}
    \begin{subfigure}{0.3\textwidth}
    \begin{centering}
   \includegraphics[width=\textwidth]{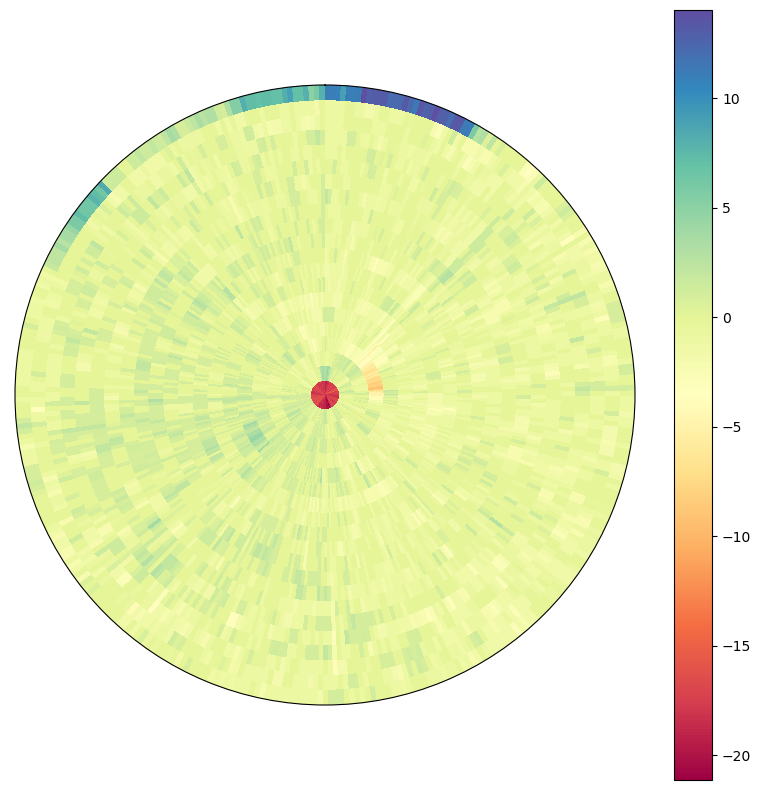}
    \label{fig:f02b_mapa_error}
    \caption{}
    \end{centering}
    \end{subfigure}
    \hfill
   \caption{Measuring the error between automatic detection and the GT for image F02b. (a) The GT rings are in green; the rings detected by the CS-TRD method are in red. (b) Assignations between detected and GT rings. (c) Error, in pixels, between the detection and the GT.}
   \label{fig:F02b_metric}
\end{centering}
\end{figure*}

\Cref{fig:F02b_metric} illustrates the results for the F02b image. \Cref{fig:F02b_metric}(a) show the CS-TRD detection in red and the GT in green.  \Cref{fig:F02b_metric}(b) shows the assignation between GT and detections. Detected rings were assigned the same color as their corresponding GT ring. Detection with no GT assignation was colored in black. In this case,  the second detection counting from the pith outward was not assigned to any GT. Additionally, the CS-TRD method did not detect the outermost ring. \Cref{fig:F02b_metric}(c) shows the radial difference between detected and GT rings. The Euclidean distance between the detected and GT ring nodes belonging to the same ray was computed. The radial error is negative if the detected node is in the inward direction concerning the GT node and positive in the opposite direction.

Once all detected chains were assigned to the GT rings, we calculated the following indicators:
\begin{enumerate}
    \item True Positive (TP): if the detected closed chain was assigned to a GT ring.
    \item False Positive (FP): if the detected closed chain was not assigned to a GT ring.
    \item False Negative (FN): if a GT ring was not assigned to any detected closed chain.
\end{enumerate}

Finally, Precision is
given by $P=\frac{TP}{TP+FP}$, Recall  by
$R=\frac{TP}{TP+FN}$ and the F-Score by $F=\frac{2PR}{P+R}$. 

\subsection{Script for performance algorithm evaluation}

On the UruDendro site, an evaluation code that permits the comparison of the results of any algorithm with the ground truth is available. A  Python script named \emph{metric\_influence\_area.py} is proposed. It requires the following inputs:
\begin{itemize}
\item \emph{dt\_filename.json}: name of the file with the results of the evaluated algorithm. An \emph{json} file with all the detected rings as a list of shapes of type polygon, each one formed by a list of points in image coordinates (pixels).  This \emph{json} file used the structure defined by Labelme Tool \cite{Wada_Labelme_Image_Polygonal}, defined as follows, not used keys are indicated:
\begin{itemize}
    \item imagePath (string) Optional. Local Path to the image file. 
    \item imageHeight (integer) Optional. Image height in pixels.
    \item imageWidth (integer) Optional. Image width in pixels. 
    \item version (string). Optional. Software version (not used).
    \item flags (dictionary). Optional. Flags used by the detection with comments (not used). 
    \item imageData (string). Optional. String with image matrix stored in hexadecimal (not used).
    \item shapes (list of dictionaries). Required. List of rings. Each element is a dictionary with the following keys. 
    \begin{itemize}
        \item label (string). Optional. Ring identifier.
        \item points (list of pixels coordinates). Required. Each pixel coordinate is a floating point number. Pixel (x,y) where x refers to the horizontal image axis and y to the vertical image axis.
    \end{itemize}
\end{itemize}

\item \emph{gt\_filename.json}: cross-section ground truth filename, from the UruDendro dataset, with the same structure described above.
\end{itemize}

The script requires the following parameters:
\begin{itemize}
\item \emph{img\_filename}: image filename from the UruDendro dataset, in \emph{png} format.
\item \emph{cx}: x pith coordinate in pixels.
\item \emph{cy}: y pith coordinate in pixels.
\item \emph{output\_dir}: output directory for the results.
\item \emph{Th}: threshold to consider a detection valid. A floating point number between 0 and 1.
\end{itemize}

The script  prints the \emph{F1-score}, \emph{Precision}, \emph{Recall}, and \emph{RMSE}. In addition, it generates in the \emph{output\_dir}, the following files:
\begin{itemize}
\item \emph{dots\_curve\_and\_rays.png} illustrates rays, ground truth, and detected rings.
\item \emph{influence\_area.png} illustrates the influence area of each ground truth. To consider a detection as True Positive, it has to have more than $th \times Nr$ nodes within the area of the assigned ground truth, where $Nr$ is the number of nodes of the detection.
\item \emph{assigned\_dt\_gt.png} illustrates the assignation between detection and ground truth. The detection and the assigned ground truth are colored with the same color. False Positive is colored in white.
\item \emph{rmse.png} illustrates, using a bar plot, the root mean square error per ring. The ring location is left empty if a ground truth ring has no assigned detection (False Negative).
\item \emph{heat\_map\_Spectral.png} illustrates the detection of a radial error in polar coordinates using a Spectral color map.
\end{itemize}

\section{Results}
\label{sec:results}

\begin{table}[htbp]
\footnotesize
\centering
\caption{Results over the UruDendro data set. \textbf{TP}: True Positives, \textbf{FP}: False Positives, \textbf{FN}: False Negatives. True Negatives are zero for all images. \textbf{P}: Precision, \textbf{R}: Recall, \textbf{F}: F-Score. \textbf{RMSE}: mean error, in pixels, between GT and detected rings. Execution time in seconds.}
\subfloat[Lumber company]{
\begin{tabular}{lccccccrr} 
\hline
\textbf{Name} & \textbf{TP} & \textbf{FP} &  \textbf{FN} & \textbf{P} & \textbf{R} & \textbf{F} & \textbf{RMSE} & \textbf{Time} \\ 
\hline
\textbf{F02a}    & 20 & 0   & 3  & 1.00 & 0.87 & 0.93 & 1.50          & 15.29       \\ 
\textbf{F02b}    & 21 & 1   & 1  & 0.96 & 0.96 & 0.96 & 4.02          & 13.26       \\ 
\textbf{F02c}    & 21 & 0   & 1  & 1.00 & 0.96 & 0.98 & 3.76          & 12.26       \\ 
\textbf{F02d}    & 20 & 1   & 0  & 0.95 & 1.00 & 0.98 & 2.11          & 8.34        \\ 
\textbf{F02e}    & 20 & 1    & 0  & 0.95 & 1.00 & 0.98 & 7.62          & 23.31       \\ 
\textbf{F03a}    & 22 & 2    & 2  & 0.92 & 0.92 & 0.92 & 8.11          & 19.37       \\ 
\textbf{F03b}    & 20 & 0    & 3  & 1.00 & 0.87 & 0.93 & 2.15          & 13.95       \\ 
\textbf{F03c}    & 23 & 0   & 1  & 1.00 & 0.96 & 0.98 & 10.69         & 7.34        \\ 
\textbf{F03d}    & 19 & 1    & 2  & 0.95 & 0.91 & 0.93 & 7.81          & 11.26       \\ 
\textbf{F03e}    & 20 & 2   & 1  & 0.91 & 0.95 & 0.93 & 1.66          & 15.70       \\ 
\textbf{F04a}    & 21 & 1    & 3  & 0.96 & 0.88 & 0.91 & 7.71          & 28.90       \\ 
\textbf{F04b}    & 19 & 0    & 4  & 1.00 & 0.83 & 0.91 & 4.60          & 40.24       \\ 
\textbf{F04c}    & 18 & 1   & 3  & 0.95 & 0.86 & 0.90 & 5.60          & 26.88       \\ 
\textbf{F04d}    & 17 & 3   & 4  & 0.85 & 0.81 & 0.83 & 2.90          & 55.38       \\ 
\textbf{F04e}    & 19 & 2    & 2  & 0.91 & 0.91 & 0.91 & 9.94          & 24.33       \\ 
\textbf{F07a}    & 18 & 1   & 6  & 0.95 & 0.75 & 0.84 & 11.68         & 18.27       \\ 
\textbf{F07b}    & 17 & 3    & 6  & 0.85 & 0.74 & 0.79 & 7.99          & 45.74       \\ 
\textbf{F07c}    & 20 & 2    & 3  & 0.91 & 0.87 & 0.89 & 4.85          & 35.81       \\ 
\textbf{F07d}    & 20 & 0   & 2  & 1.00 & 0.91 & 0.95 & 1.04          & 19.95       \\ 
\textbf{F07e}    & 8  & 4    & 14 & 0.67 & 0.36 & 0.47 & 8.17          & 40.29       \\ 
\textbf{F08a}    & 21 & 1    & 3  & 0.96 & 0.88 & 0.91 & 5.28          & 17.13       \\ 
\textbf{F08b}    & 21 & 1   & 2  & 0.96 & 0.91 & 0.93 & 1.72          & 23.57       \\ 
\textbf{F08c}    & 21 & 1   & 2  & 0.96 & 0.91 & 0.93 & 2.05          & 13.25       \\ 
\textbf{F08d}    & 20 & 0   & 2  & 1.00 & 0.91 & 0.95 & 2.10          & 9.54        \\ 
\textbf{F08e}    & 22 & 0   & 0  & 1.00 & 1.00 & 1.00 & 6.70          & 17.65       \\ 
\textbf{F09a}    & 21 & 0   & 3  & 1.00 & 0.88 & 0.93 & 2.20          & 16.13       \\ 
\textbf{F09b}    & 22 & 1    & 1  & 0.96 & 0.96 & 0.96 & 2.91          & 11.98       \\ 
\textbf{F09c}    & 21 & 0    & 3  & 1.00 & 0.88 & 0.93 & 2.83          & 9.11        \\ 
\textbf{F09d}    & 21 & 0   & 2  & 1.00 & 0.91 & 0.96 & 3.44          & 8.71        \\ 
\textbf{F09e}    & 14 & 5   & 8  & 0.74 & 0.64 & 0.68 & 7.08          & 38.73       \\ 
\textbf{F10a}    & 17 & 1   & 5  & 0.94 & 0.77 & 0.85 & 4.24          & 15.23       \\ 
\textbf{F10b}    & 19 & 2  & 3  & 0.91 & 0.86 & 0.88 & 4.75          & 20.39       \\ 
\textbf{F10e}    & 18 & 0   & 2  & 1.00 & 0.90 & 0.95 & 2.13          & 12.57       \\ 
\hline
\end{tabular}
}
\subfloat[Plywood company]{
\begin{tabular}{lccccccrr} 
\hline
\textbf{Name} & \textbf{TP} & \textbf{FP} &  \textbf{FN} & \textbf{P} & \textbf{R} & \textbf{F} & \textbf{RMSE} & \textbf{Time} \\ 
\hline
\textbf{L02a}    & 14 & 1   & 2  & 0.93 & 0.88 & 0.90 & 16.95         & 26.21       \\ 
\textbf{L02b}    & 4  & 2   & 11 & 0.67 & 0.27 & 0.38 & 9.41          & 21.90       \\ 
\textbf{L02c}    & 11 & 0   & 2  & 1.00 & 0.85 & 0.92 & 5.33          & 21.33       \\ 
\textbf{L02d}    & 7  & 3   & 7  & 0.70 & 0.50 & 0.58 & 5.66          & 28.80       \\ 
\textbf{L02e}    & 11 & 0   & 3  & 1.00 & 0.79 & 0.88 & 4.92          & 18.66       \\ 
\textbf{L03a}    & 14 & 0   & 3  & 1.00 & 0.82 & 0.90 & 3.45          & 16.20       \\ 
\textbf{L03b}    & 15 & 1   & 1  & 0.94 & 0.94 & 0.94 & 2.22          & 10.17       \\ 
\textbf{L03c}    & 15 & 1   & 1  & 0.94 & 0.94 & 0.94 & 9.01          & 8.28        \\ 
\textbf{L03d}    & 14 & 0   & 1  & 1.00 & 0.93 & 0.97 & 10.63         & 8.26        \\ 
\textbf{L03e}    & 13 & 0   & 1  & 1.00 & 0.93 & 0.96 & 3.97          & 14.96       \\ 
\textbf{L04a}    & 15 & 0    & 2  & 1.00 & 0.88 & 0.94 & 6.21          & 7.98        \\ 
\textbf{L04b}    & 15 & 0   & 1  & 1.00 & 0.94 & 0.97 & 6.35          & 10.66       \\ 
\textbf{L04c}    & 14 & 0  & 2  & 1.00 & 0.88 & 0.93 & 3.30          & 8.26        \\ 
\textbf{L04d}    & 14 & 1   & 2  & 0.93 & 0.88 & 0.90 & 7.88          & 6.19        \\ 
\textbf{L04e}    & 10 & 1   & 5  & 0.91 & 0.67 & 0.77 & 4.09          & 10.14       \\ 
\textbf{L07a}    & 13 & 1   & 4  & 0.93 & 0.77 & 0.84 & 1.89          & 13.96       \\ 
\textbf{L07b}    & 13 & 0   & 3  & 1.00 & 0.81 & 0.90 & 6.56          & 9.01        \\ 
\textbf{L07c}    & 14 & 1   & 3  & 0.93 & 0.82 & 0.88 & 2.41          & 5.56        \\ 
\textbf{L07d}    & 14 & 0   & 2  & 1.00 & 0.88 & 0.93 & 1.73          & 5.22        \\ 
\textbf{L07e}    & 11  & 0   & 3  & 1.00 & 0.79 & 0.88 & 13.26         & 17.80       \\ 
\textbf{L08a}    & 15 & 0   & 2  & 1.00 & 0.88 & 0.94 & 2.38          & 8.94        \\ 
\textbf{L08b}    & 14 & 1   & 2  & 0.93 & 0.88 & 0.90 & 11.99         & 24.48       \\ 
\textbf{L08c}    & 15 & 0   & 1  & 1.00 & 0.94 & 0.97 & 2.52          & 8.74        \\ 
\textbf{L08d}    & 13 & 0   & 1  & 1.00 & 0.93 & 0.96 & 9.57          & 5.45        \\ 
\textbf{L08e}    & 14 & 1   & 1  & 0.93 & 0.93 & 0.93 & 8.14          & 17.82       \\ 
\textbf{L09a}    & 14 & 0   & 3  & 1.00 & 0.82 & 0.90 & 3.29          & 10.03       \\ 
\textbf{L09b}    & 15 & 1   & 1  & 0.94 & 0.94 & 0.94 & 2.26          & 13.54       \\ 
\textbf{L09c}    & 15 & 2   & 1  & 0.88 & 0.94 & 0.91 & 2.90          & 12.02       \\ 
\textbf{L09d}    & 13 & 1   & 2  & 0.93 & 0.87 & 0.90 & 2.12          & 12.03       \\ 
\textbf{L09e}    & 13 & 0   & 2  & 1.00 & 0.87 & 0.93 & 4.14          & 14.66       \\ 
\textbf{L11b}    & 15 & 1   & 1  & 0.94 & 0.94 & 0.94 & 1.54          & 10.96       \\ 
        &    &     &    &      &     &       &  & \\
        &    &     &    &      &     &       &  & \\
\hline
\end{tabular}
}
\label{Tab:results_nuestra-60}
\end{table}

\subsection{Automatic ring detection}

 \Cref{Tab:results_nuestra-60} illustrates the CS-TRD performance over the UruDendro data set. All images were resized to 1500x1500. The CS-TRD edge detector parameter $\sigma$ was  set to 3. The Average F1-Score is 89\%, the average Precision is 95\%, and the  average Recall is 86\%. For example, in the image \textit{F03d}, the method fails to detect two GT rings, so $FN=2$. The other rings are correctly detected. The table also shows the execution time for each image and the RMSE error (\Cref{equ:rmse}) between the detected and GT rings. The mean RMSE error between GT and detected rings over the whole data set is 5.27 pixels, and the mean execution time for each image is 17.27 seconds.

\begin{figure}
\begin{centering}
    \begin{subfigure}{0.3\textwidth}
    \begin{centering}
    \includegraphics[width=\textwidth]{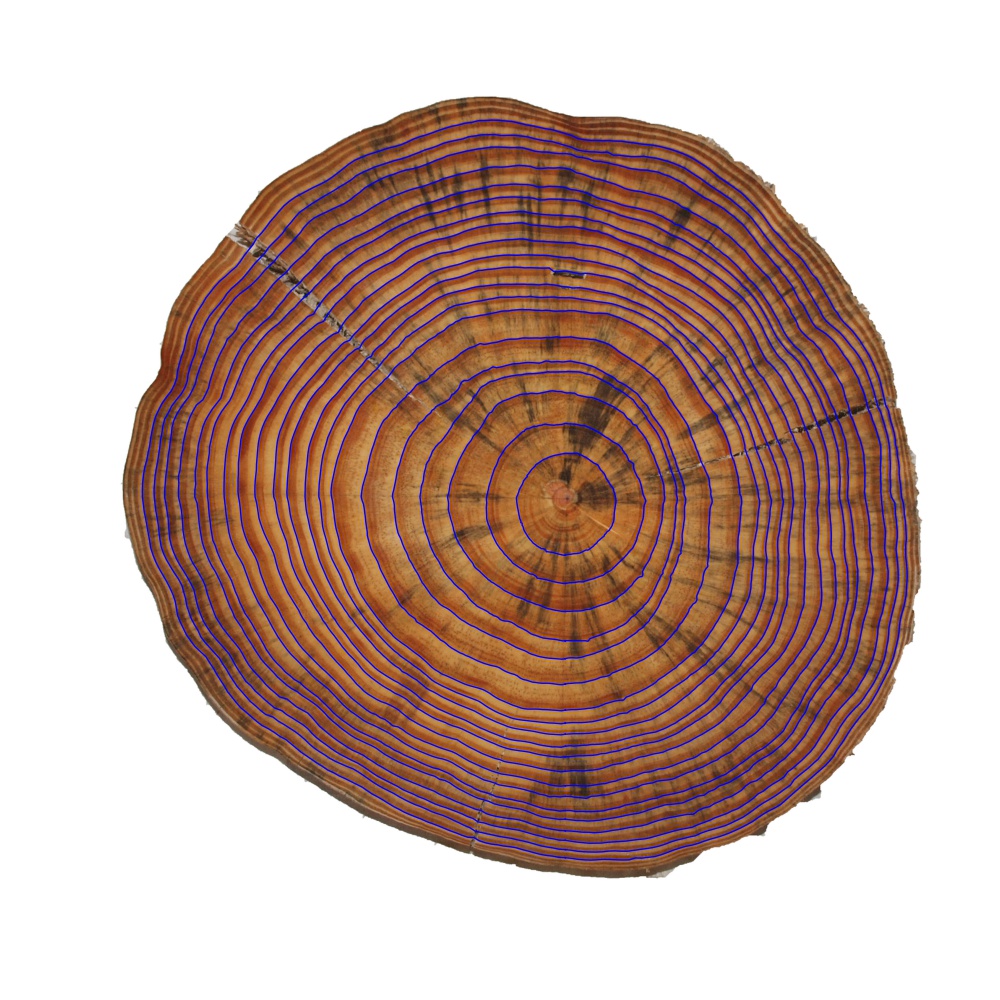}
    \label{fig:ddbb-F02a_output}
    \caption{F02a}
    \end{centering}
    \end{subfigure}
    \hfill
    \begin{subfigure}{0.3\textwidth}
    \begin{centering}
    \includegraphics[width=\textwidth]{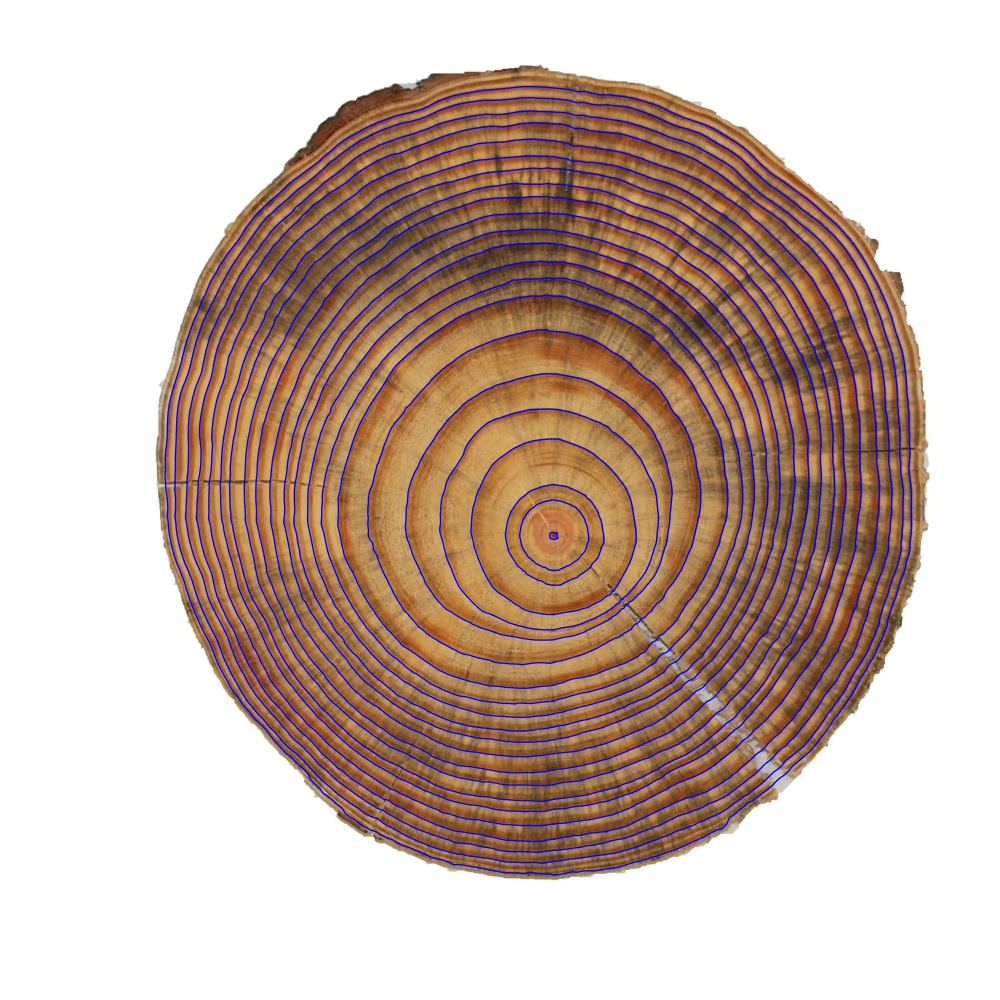}
    \label{fig:ddbb-F02b_output}
    \caption{F02b}
    \end{centering}
    \end{subfigure}
    \hfill
    \begin{subfigure}{0.3\textwidth}
    \begin{centering}
    \includegraphics[width=\textwidth]{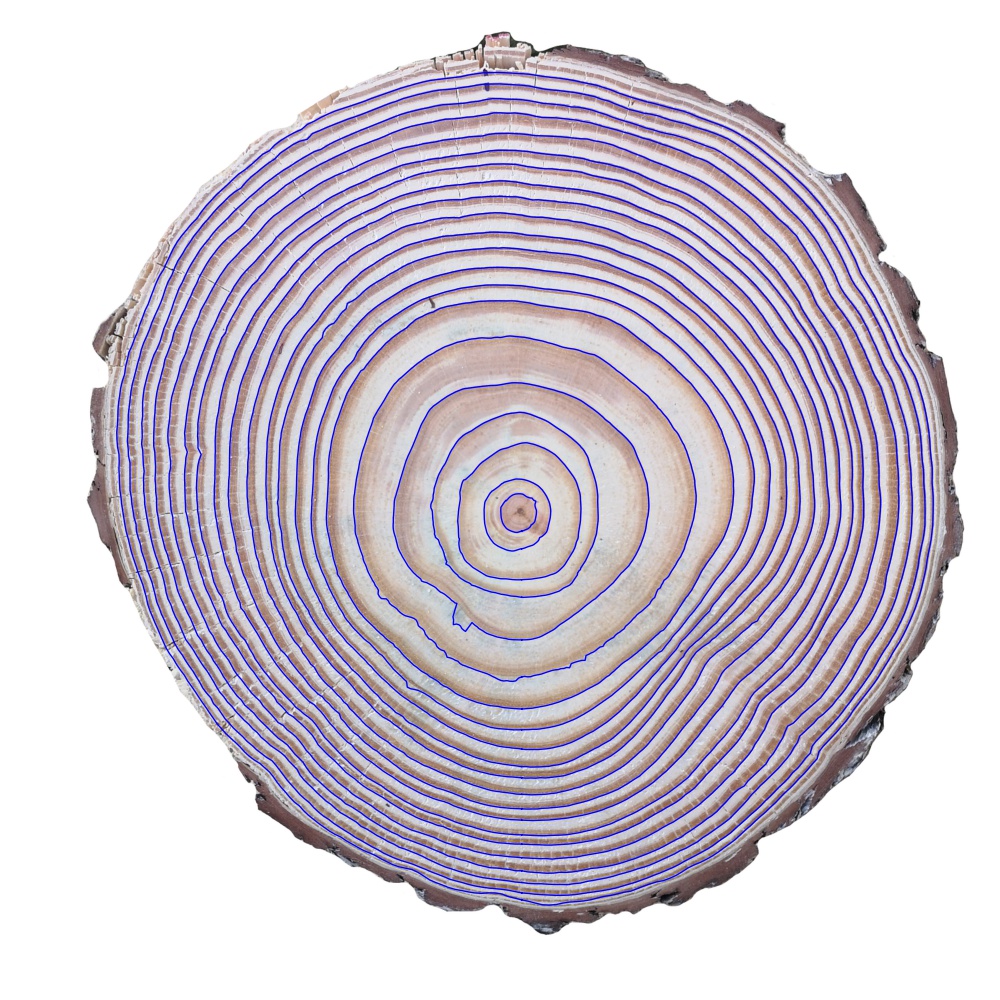}
    \label{fig:ddbb-F02c_output_output}
    \caption{F02c}
    \end{centering}
    \end{subfigure}
    \hfill
    \begin{subfigure}{0.3\textwidth}
    \begin{centering}
   \includegraphics[width=\textwidth]{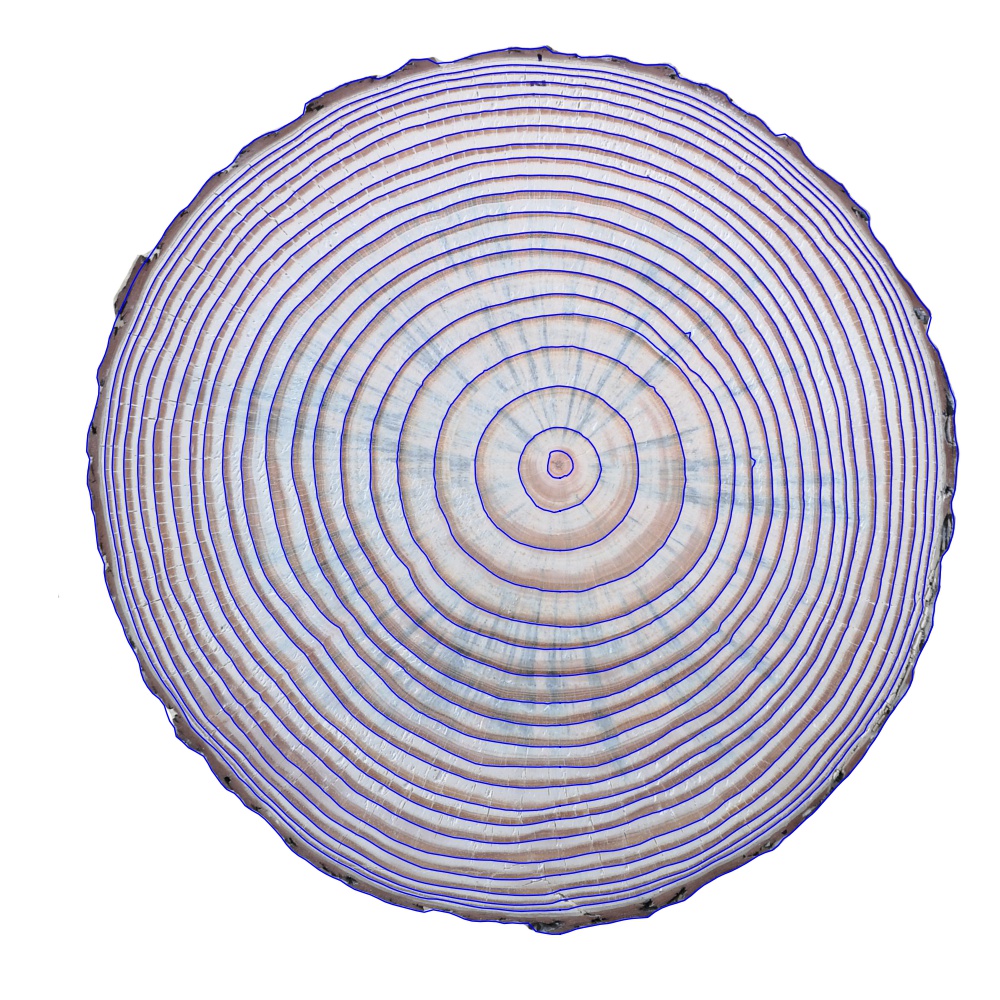}
    \label{fig:ddbb-F02d_output}
    \caption{F02d}
    \end{centering}
    \end{subfigure}
    \hfill
    \begin{subfigure}{0.3\textwidth}
    \begin{centering}
   \includegraphics[width=\textwidth]{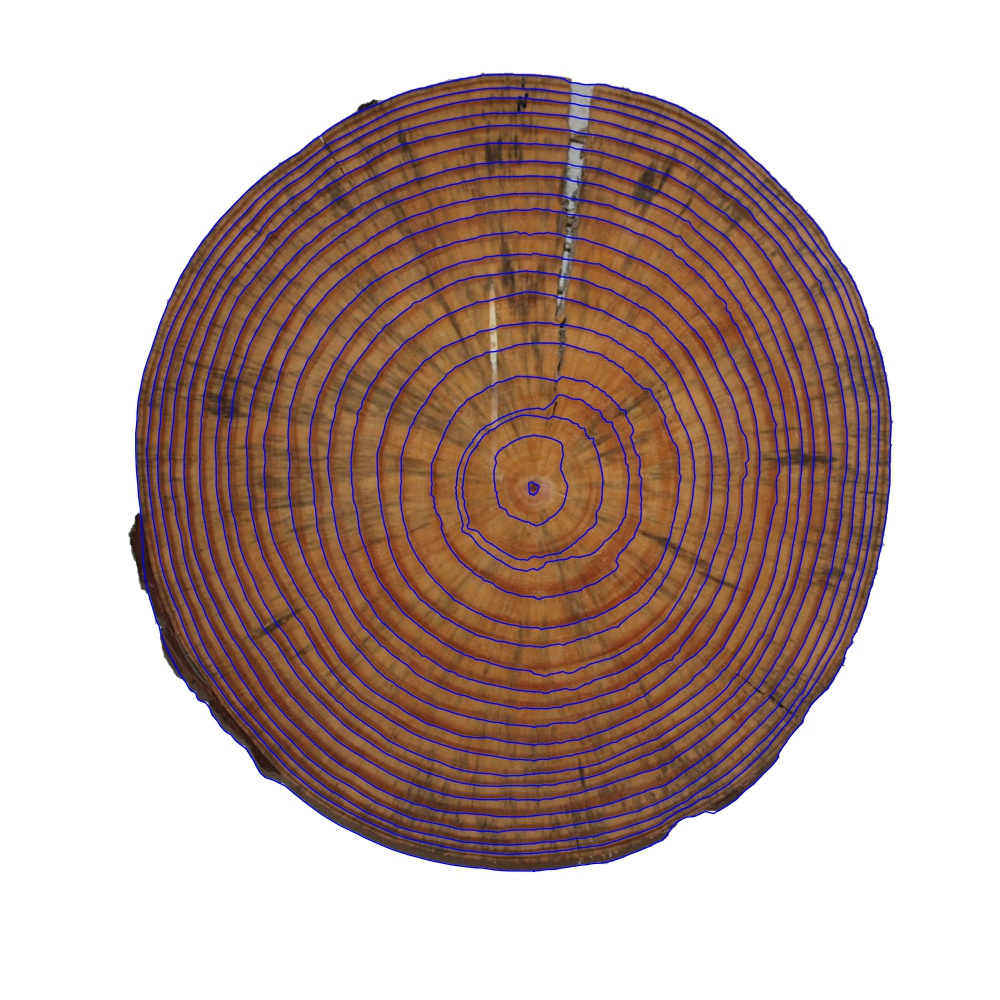}
    \label{fig:ddbb-F02e_output}
    \caption{F02e}
    \end{centering}
    \end{subfigure}
    \hfill
    \begin{subfigure}{0.3\textwidth}
    \begin{centering}
   \includegraphics[width=\textwidth]{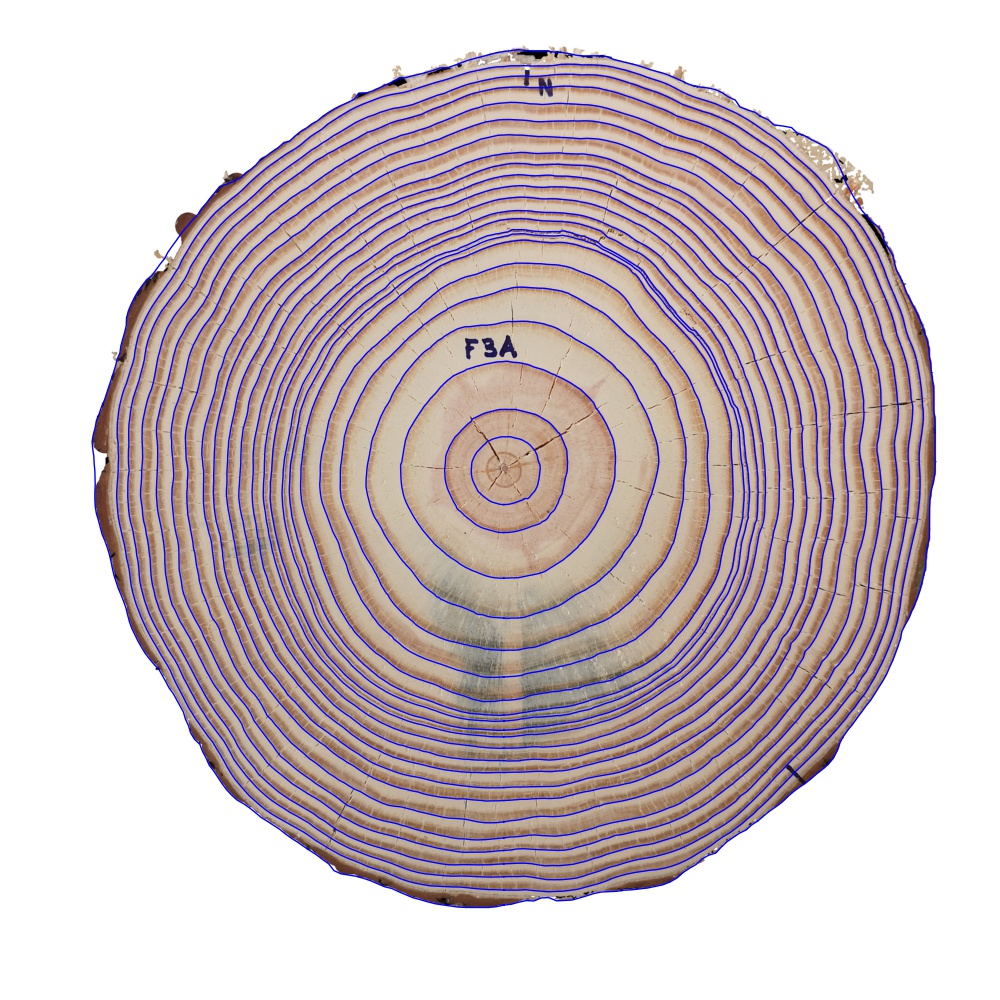}
    \label{fig:ddbb-F03c_output}
    \caption{F03c}
    \end{centering}
    \end{subfigure}
    \hfill
    \begin{subfigure}{0.3\textwidth}
    \begin{centering}
   \includegraphics[width=\textwidth]{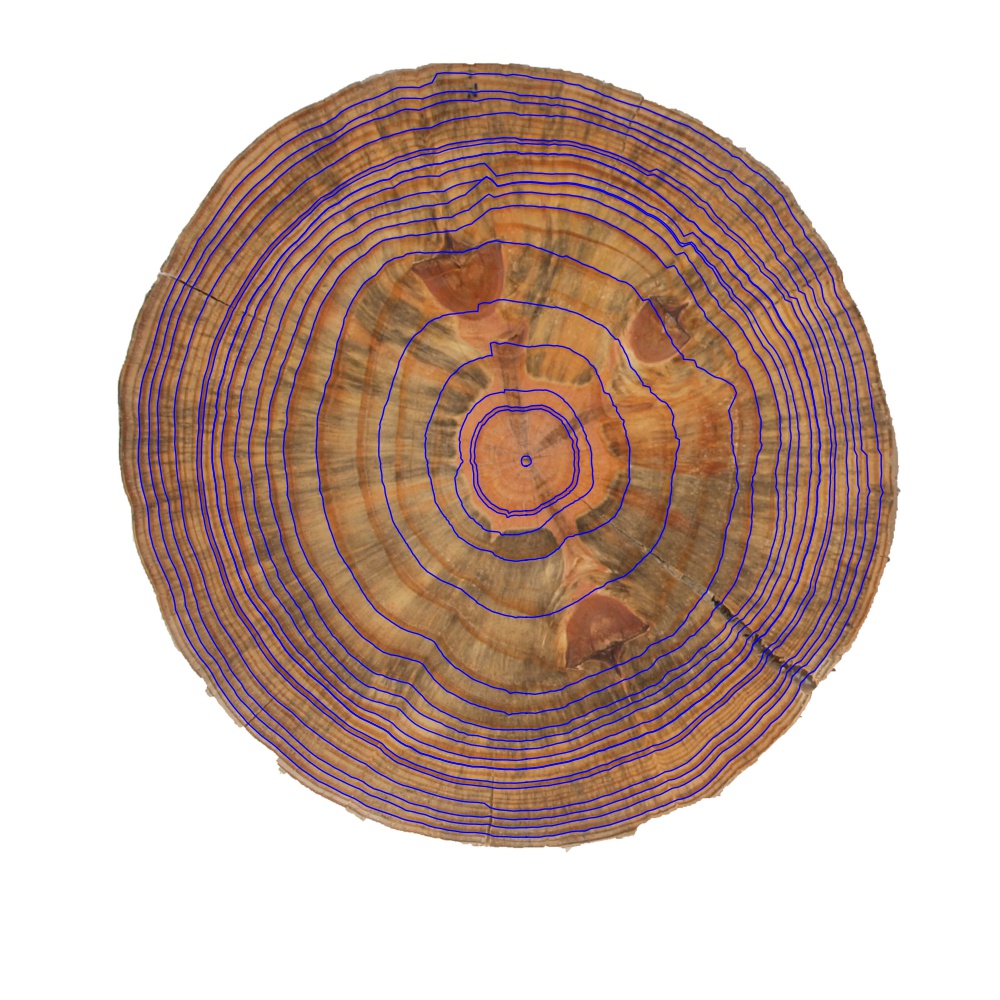}
    \label{fig:ddbb-F07b_output}
    \caption{F07b}
    \end{centering}
    \end{subfigure}
    \hfill
    \begin{subfigure}{0.3\textwidth}
    \begin{centering}
   \includegraphics[width=\textwidth]{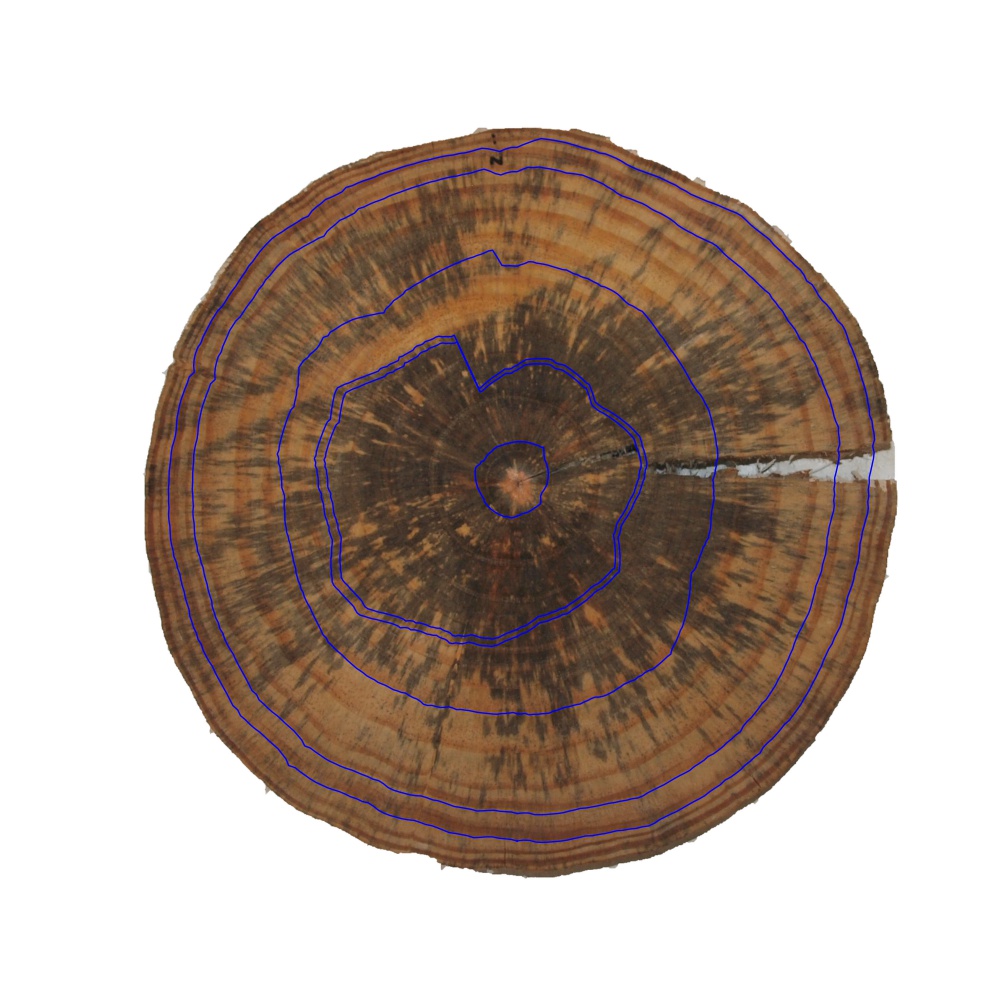}
    \label{fig:ddbb-L02b_output}
    \caption{L02b}
    \end{centering}
    \end{subfigure}
    \hfill
    \begin{subfigure}{0.3\textwidth}
    \begin{centering}
   \includegraphics[width=\textwidth]{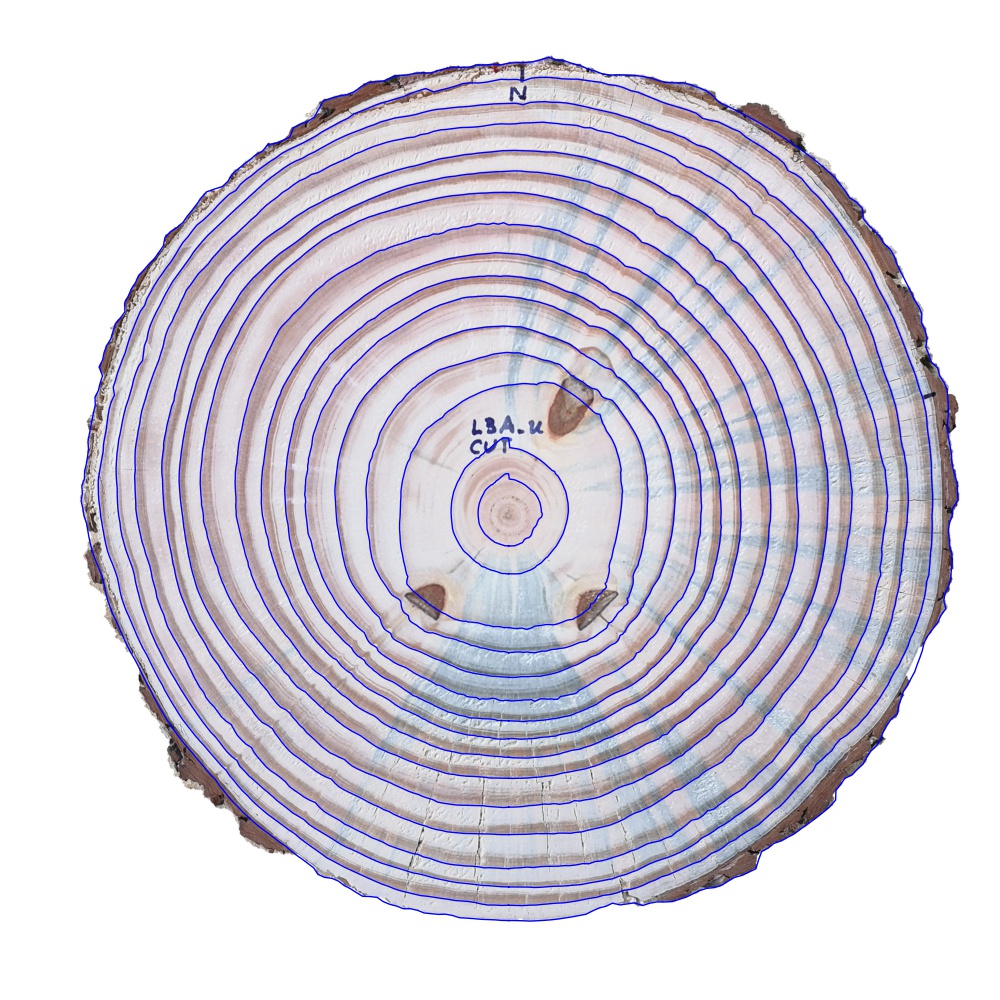}
    \label{fig:ddbb-L03c_output}
    \caption{L03c}
    \end{centering}
    \end{subfigure}
    \hfill
   \caption{Some results for the UruDendro data set. In general, the results are very good. In the case of image L02b, it seems clear that there is not enough edge information to see the rings due to the fungus stain.}
   \label{fig:ddbb_output}
\end{centering}
\end{figure}

 \Cref{fig:ddbb_output} illustrates the CS-TRD output results for some images in the UruDendro data set: Images F02a, F02b, F02c, F02d, F02e, F03c, and L03c which have an F-Score above 93\%. 

\subsection{Tree-ring measurement}

\subsubsection{Manual measurements}

\begin{figure}
\begin{centering}
\begin{subfigure}{0.45\textwidth}
\begin{centering}
\includegraphics[width=\linewidth]{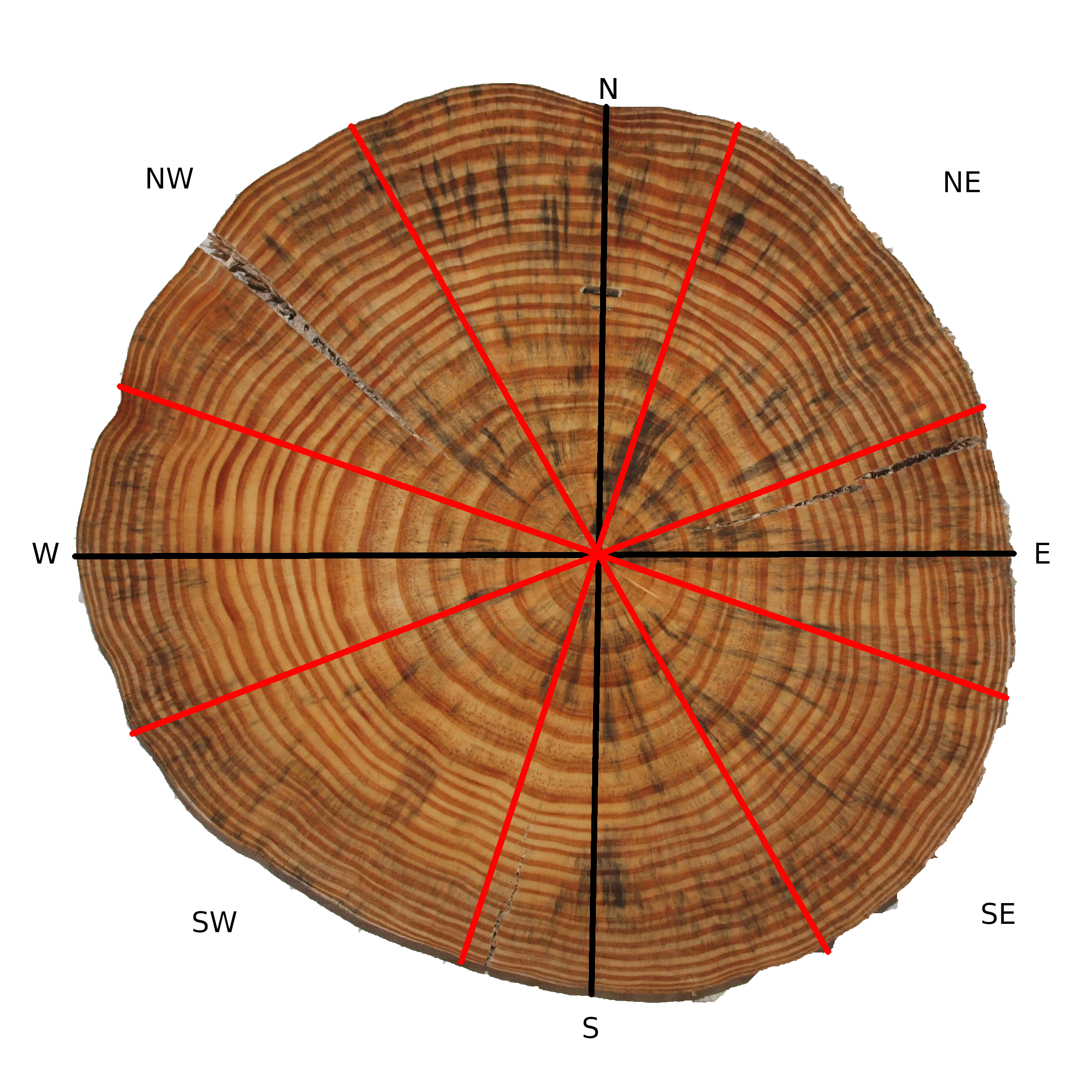}
\caption{}
\end{centering}
\end{subfigure}
\begin{subfigure}{0.45\textwidth}
\begin{centering}
\includegraphics[width=\linewidth]{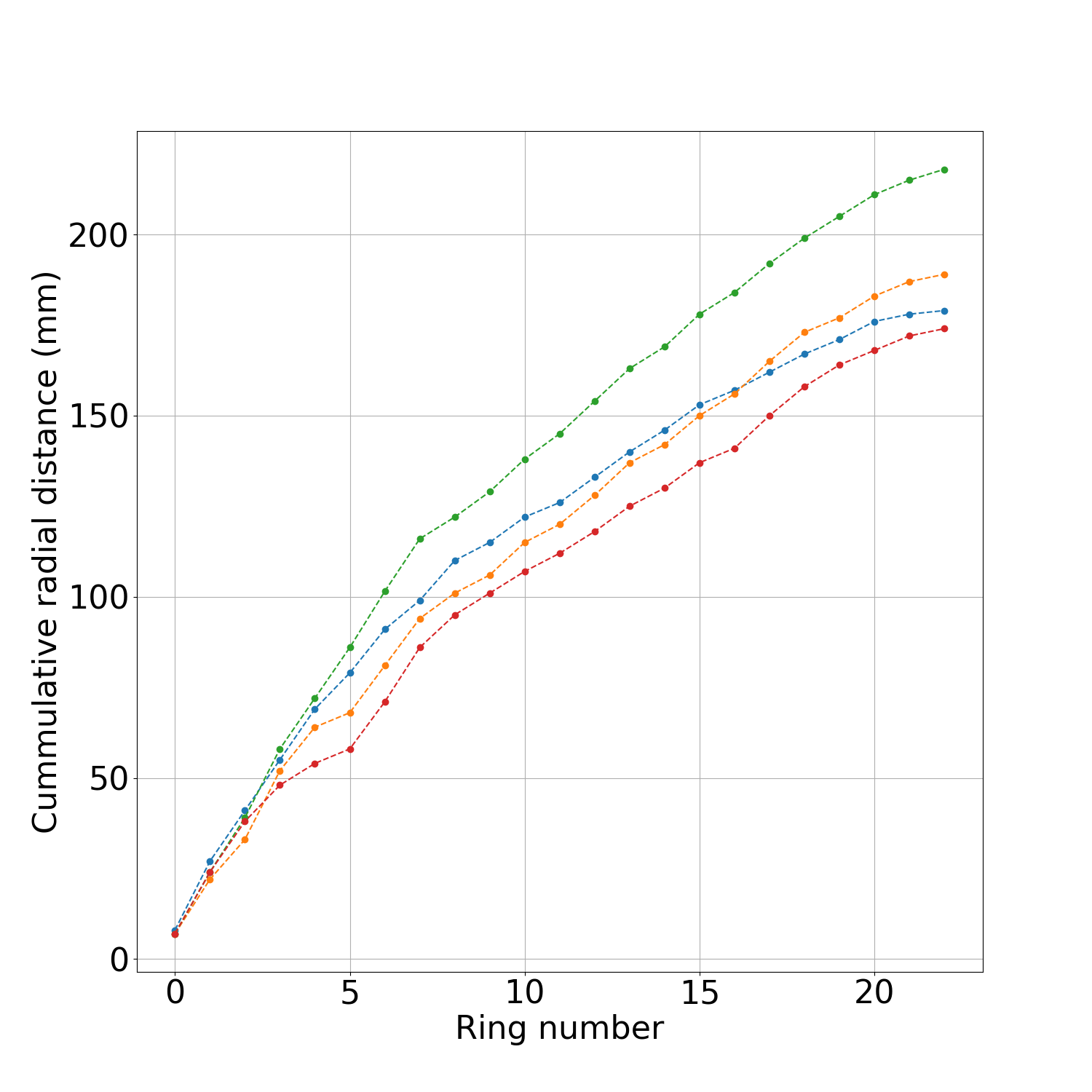}
\caption{}
\end{centering}
\end{subfigure}
\caption{(a) Cardinal directions over a tree cross-section (in this case, image F02a). (b) Ring width was measured by an expert over the four black lines corresponding to each cardinal direction. The South direction is blue, the North is orange, the West is green, and the East is red. Measurements are in millimeters from the center to the intersection of the rings in the given direction.}
\label{fig:L02b_cumulative_radial}
\end{centering}
\end{figure}

For 35 images of the UruDendro data set, an expert measured the ring width (in millimeters) along the north, south, east, and west axes as illustrated by the black lines in \Cref{fig:L02b_cumulative_radial}.a. For the rings traced by the expert over the L02b image, the distance from the center to the intersection between each ring and the line is shown in \Cref{fig:L02b_cumulative_radial}.b. Each color corresponds to a different cardinal direction. Measuring the distance to the pith from the GT rings along the cardinal lines using the CS-TRD software provided the same measurements as those made by the expert, in pixels. Assuming a linear model for the relationship between radial distances in millimeters and pixels, we computed the pixel dimensions in millimeters for a given image with expert manual annotations. \Cref{equ:calibration} shows the relation between pixel and millimeter measurements.

\begin{equation}
    R_{(mm)} = m * R_{(pixels)}
    \label{equ:calibration}
\end{equation}

Where $R_{(mm)}$ is the measurement in millimeters over a (horizontal or vertical) direction, $R_{(pixels)}$ is the same measurement but in pixels, and  \textit{m} is a constant that needs to be found.  We estimated the value of the constant \textit{m} using all the measurements along a given cardinal direction on a given cross-section or combining the measurements along several cardinal directions. 

Given a data subset (\textit{e.g.}, the measurements made along the North and South path), for a given cross-section, we have the measures made manually by the expert, \textit{$X_h$}, in millimeters, and the measurement computed by the CS-TRD software, \textit{$X_{sw}$}, in pixels. The value of \textit{m} was computed by fitting the data by a least squares method.  This method allows the calibration of each image to know the dimensions in millimeters. We found $\Delta x \approx \Delta y$,  indicating that the pixels were almost square-shaped. The values for each image are in the supplemental material.

\subsubsection{Automatic tree-ring measurements}
\label{sec:automaticMeasument}

\begin{figure*}
\begin{centering}
\begin{subfigure}{0.45\textwidth}
\begin{centering}
\includegraphics[width=\linewidth]{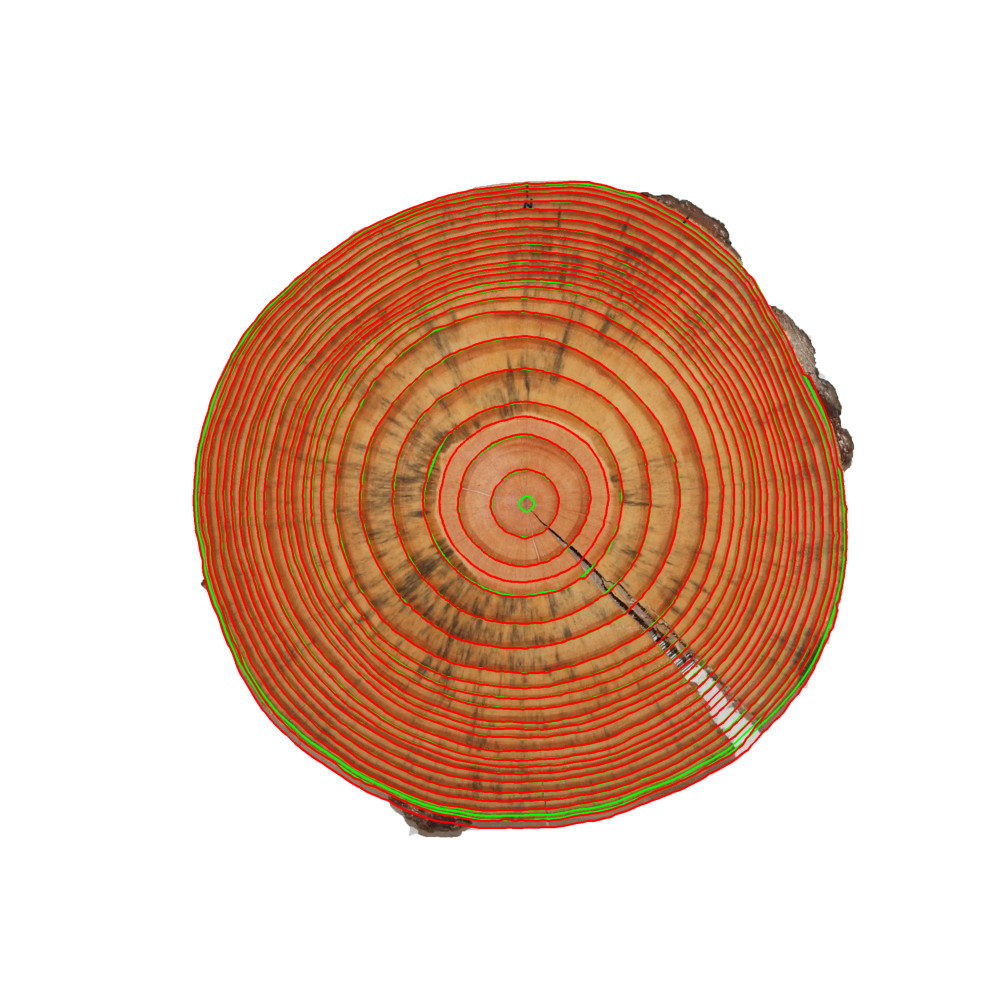}
\caption{}
\end{centering}
\end{subfigure}
\begin{subfigure}{0.4\textwidth}
\begin{centering}
\includegraphics[width=\linewidth]{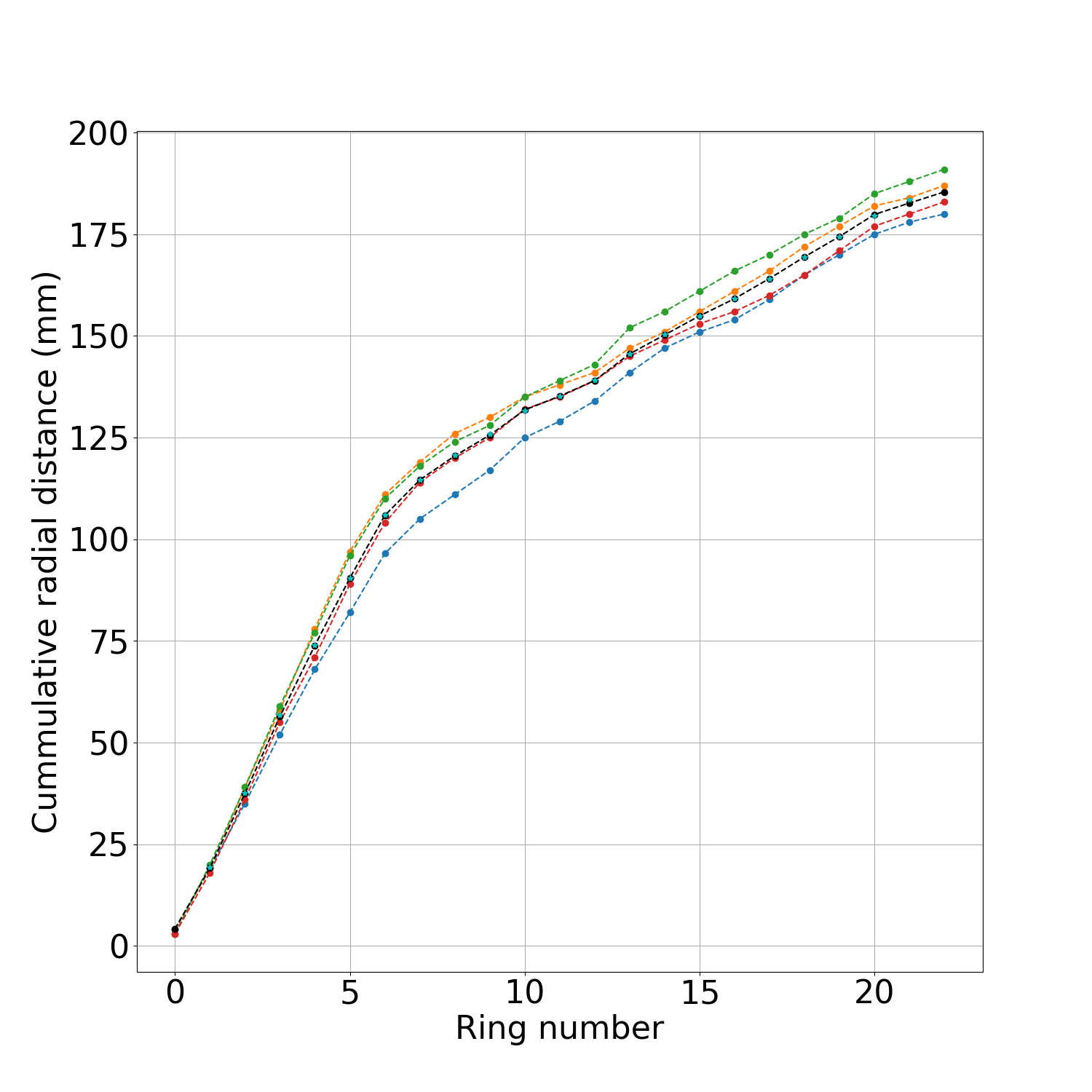}
\caption{}
\end{centering}
\end{subfigure}
\begin{subfigure}{0.4\textwidth}
\begin{centering}
\includegraphics[width=\linewidth]{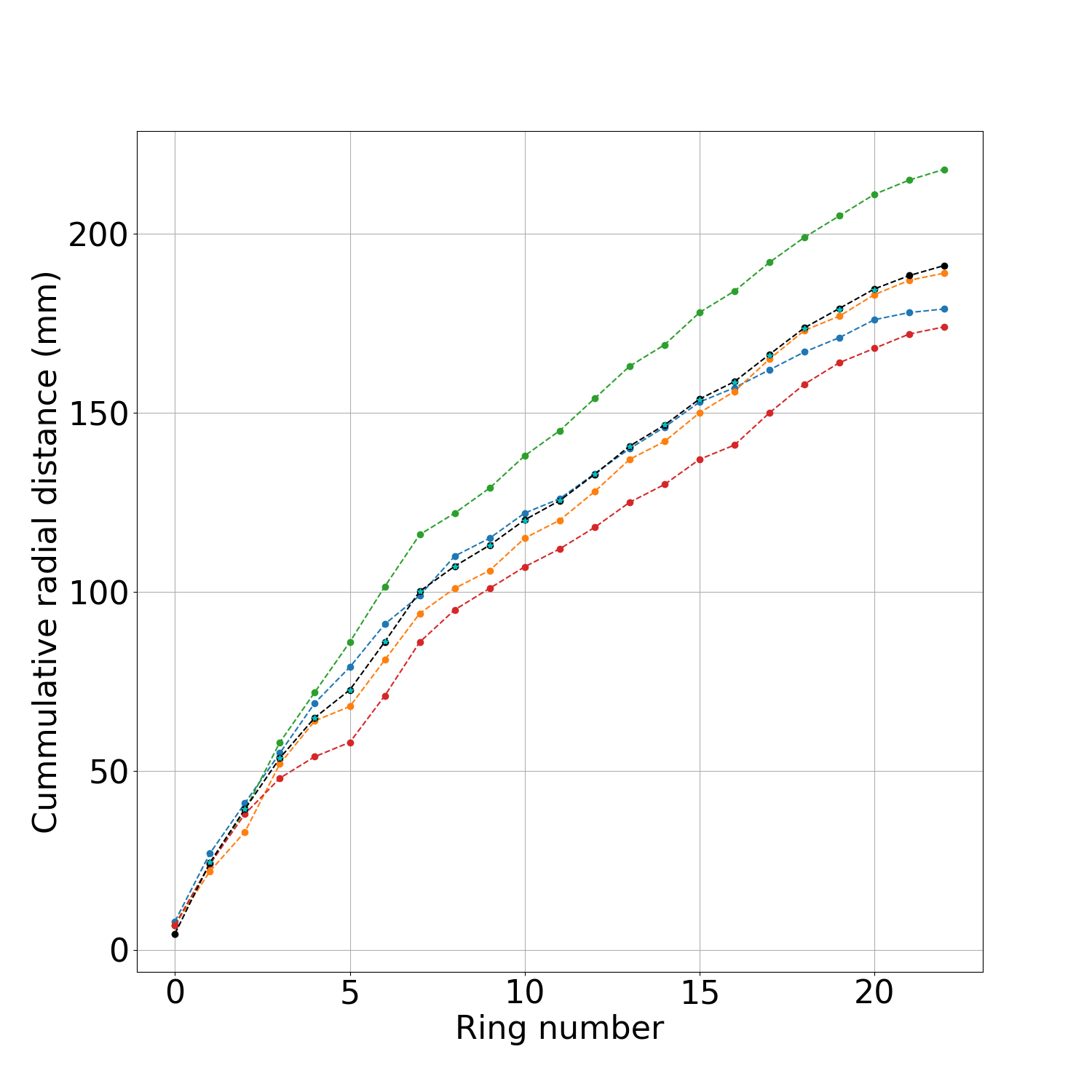}
\caption{}
\end{centering}
\end{subfigure}
\begin{subfigure}{0.4\textwidth}
\begin{centering}
\includegraphics[width=\linewidth]{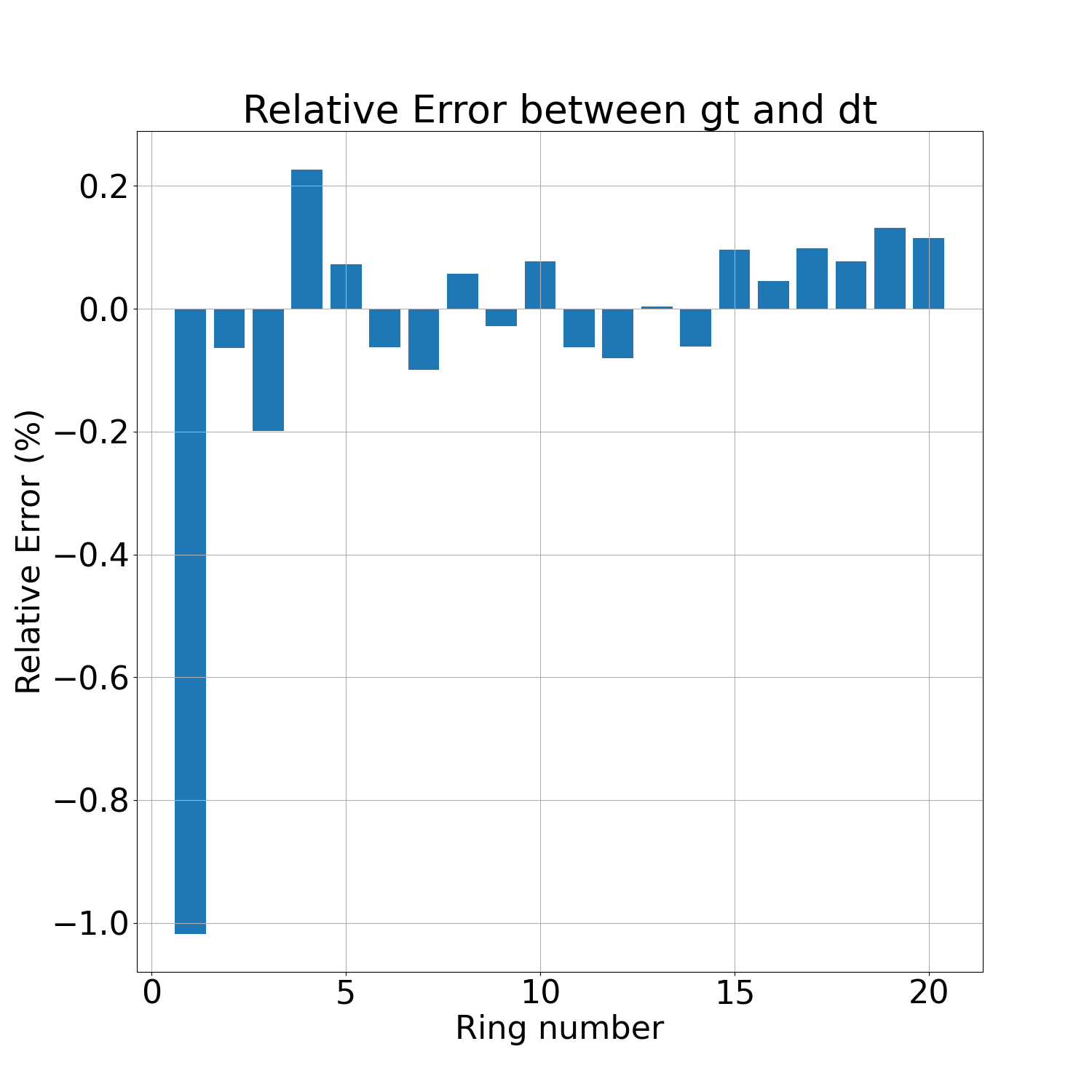}
\caption{}
\end{centering}
\end{subfigure}

\caption{Comparison between cross-section measurements made by one expert and the CS-TRD software. (a) Image F09b. CS-TRD detections are in red, and GT rings are in green. Note that CS-TRD made two false positives, two false negatives, and twenty-one true positives. (b) Image F09b. Ring width was measured by an expert over the four black cardinal lines. The South direction is in blue, the North in orange, the West in green, and the East in red. The equivalent radius is the black curve. Note that for that black (equivalent) curve, the measurements made using the GT rings are in black, and the measurements made using the CS-TRD output are in light blue and are almost superposed, indicating that CS-TRD gives a fairly accurate measure. Measurements are in millimeters from the center to the intersection of the rings in a given direction. (c) The same measurements but for image F02a, showing that the equivalent area is accurate for both symmetric and asymmetric cross-sections (d) Image F02a. Difference between GT and CS-TRD output equivalent ring width, in percentage. }
\label{fig:equivalent_width}
\end{centering}
\end{figure*}

We compared the manual measurements made by an expert with those made by the CS-TRD software. Radial growth rates were not uniform in all ray directions. For example, in \Cref{fig:ddbb_output} image L03c was roughly symmetric, and image F02b presented substantial radial asymmetries, i.e. the rings exhibited variable growth in different linear directions. Therefore, a metric of ring width along all cardinal directions was considered more representative of the overall tree diameter growth. If we approximate a ring by an equivalent circle, and knowing that the area of a disk $i$, with radius $r_i$ is $Area_{i} = \pi r_i^2$, we can use the (manually marked or automatically detected) rings to estimate an equivalent ring radius.  For a given ring, the equivalent radius $r^{eq}_i = \sqrt{Area_{i}/\pi}$, where $Area_{i}$ is the area inside the (manually marked or automatically detected) ring $i$ and $r^{eq}_i$ is the radius of a disk with the same area as the corresponding traced ring.  The difference between this equivalent radius for successive rings gives an idea of the global growth of that ring, named $\Delta r^{eq}_i$, which can be calculated by \Cref{equ:equi_width}.

\begin{equation}
    \Delta r^{eq}_i =  r^{eq}_i -  r^{eq}_{i-1} = \sqrt{Area_{i}/\pi} - \sqrt{Area_{i-1}/\pi} 
    \label{equ:equi_width}
\end{equation}

 \Cref{fig:equivalent_width} illustrates the cumulative ring width measurements made by hand as well as the cumulative equivalent ring width estimated with \Cref{equ:equi_width} using the GT rings as well as the output of the CS-TRD software. As can be seen, this is a more representative value for the global area growth for a given ring, independent of cross-section asymmetries.

Experts required between 1 and 8 hours per cross-section to trace the GT delineation for the UruDendro data set. The procedure was accelerated using the CS-TRD method, as the software processed a cross-section in 17 seconds, on average. As shown in  \Cref{fig:equivalent_width}.d, automatic ring delineation was very precise: the maximum ring width relative error between CS-TRD detections and GT was 1\%, and almost all rings are detected. Deleting false positive rings as well as delineating a few missing rings is an easy task that can be done using an open source tool such as Labelme \cite{Wada_Labelme_Image_Polygonal}.  

\section{Discussion and conclusions}
\label{sec:conclusions}

UruDendro, a new public data set of cross-sections of \textit{Pinus taeda} from Uruguay, is presented along with the ground truth traces of the pith and annual rings. The publicly available dataset can be used to test different algorithms for pith and tree ring automatic detection.

The CS-TRD software for automatically detecting the tree rings is presented in its summarized version to the dendrochronology community. At the same time, an in-depth presentation of the method will be made available to the image-processing community through an in-depth publication. 
The results on the overall performance of the CS-TRD method applied to the UruDendro data set of \textit{Pinus taeda} cross-sections show the potential of this algorithm to automatically solve some dendro-metric tasks and aid practitioners in the phase of ring identification and delimitation. We conclude that tree-ring measurements can be automated using the CS-TRD method, at least for some angiosperm species. The manual correction of missing or incorrect rings can be conducted using a simple graphical interface.  Overall, this technique allows for precise ring-width measurements obtained in minutes. 

The automatic detection of the entire annual ring and the use of the \emph{equivalent radius} introduced in section \ref{sec:automaticMeasument} (an equivalent cross-section with the same area as the corresponding traced ring) enables the measurement of the total area of the annual growth of the ring, rather than simply radial growth in one dimension. This minimizes the influence of directional inhomogeneities in the structure and growth of the tree. The ability to obtain multiple ring width measurements and an average increment representing the entire cross-section's growth provides the potential for future research on  stem transverse  area increment, volume increment, and carbon sequestration when combined with volumetric data.  To the best of our knowledge, the use of the \emph{equivalent radius} is new for this purpose.

One of the future applications of this work is the automatic detection of variability in the early to latewood transition, including the formation of compression wood.  Compression wood is darker than regular wood as it absorbs light to a greater degree due to its high lignin content and scatters light to a lesser degree than typical early wood due to the thick walls of CW fibers.\cite{ForestResearch2005}. Its darker coloring is the optical property that distinguishes CW from the naked eye.  Spectral analysis of compression wood has been developed for other conifers in Norway  \cite{nystrom1999real} and for \textit{Pinus radiata} D.Don in New Zealand  \cite{pont2007disc}.  The automatic detection of rings in South America in \textit{Pinus taeda } is a step toward tackling the problem of CW detection and understanding the potential factors that influence its formation.

\textbf{Declaration of Competing Interest}
The authors declare that they have no known competing financial interests or personal relationships that could have appeared to influence the work reported in this paper.

\textbf{Data Availability}
The UruDendro dataset is openly and freely available at the website (occluded for double-anonymized evaluation purposes) \off{\url{https://iie.fing.edu.uy/proyectos/madera/}}. It can be downloaded upon accepting an open science inspired license agreement.

A detailed explanation, the source code, and a demo page will be available at the Image Processing On Line website (occluded for double-anonymized evaluation purposes)\off{\url{https://ipolcore.ipol.im/demo/clientApp/demo.html?id=77777000390}}.

\textbf{Acknowledgements}
The study is an interdisciplinary collaboration among researchers at Uruguayan institutions: Facultad de Ingenieria, Facultad de Agronomia, CENUR Noreste, and CENUR Litoral Norte, all from Universidad de la República. A.S. was supported by the Fellowship ANII FCE\_1\_2019\_1\_155963 and the Program for Development in Basic Sciences (PEDECIBA).  ANII FCE$\_$1$\_$2019$\_$155963 provided logistical support for coauthor C.L. The companies Grupo Lumin, FYMNSA, Arboreal, Agroempresa Forestal, Global Forest Partnership and Cambium, are acknowledged for logistical support during tree harvesting. Fundación Latitud, Forestry Department of Facultad de Agronomía, and the Forestry System of the Instituto Nacional de Investigación Agropecuaria (INIA) are acknowledged for being part of the experimental design that led to the collection of cross-sections analyzed in this work.


{\small
\bibliographystyle{apalike}
\bibliography{TreeRingDetection}
}

\end{document}